\RequirePackage{fix-cm}
\documentclass[10pt,twocolumn,letterpaper]{article}
\usepackage[pagenumbers]{cvpr}
\usepackage[accsupp]{axessibility}
\InputIfFileExists{latexml.sty}{}{}
\providecommand{\iflatexml}{\iffalse}
\usepackage[dvipsnames]{xcolor}
\usepackage{array,multirow,graphicx,booktabs,adjustbox}
\iflatexml
\usepackage{placeins}
\else
\usepackage{tikz,pgfplots,placeins}
\usetikzlibrary{arrows.meta,decorations.pathreplacing}
\pgfplotsset{compat=1.18}
\definecolor{irbRed}{rgb}{0.72,0.08,0.08}
\definecolor{irbGreen}{rgb}{0.0,0.49,0.0}
\definecolor{irbBlue}{rgb}{0.12,0.29,0.60}
\definecolor{irbGray}{gray}{0.45}
\pgfplotsset{
  irb/.style={
    font=\scriptsize,
    axis line style={black!65, line width=0.5pt},
    every tick/.style={black!65, line width=0.5pt},
    tick align=outside, tick pos=left,
    major tick length=2pt,
    xlabel style={font=\scriptsize, yshift=2pt},
    ylabel style={font=\scriptsize, yshift=-3pt},
    title style={font=\footnotesize\bfseries, yshift=-4pt},
    legend style={font=\scriptsize, draw=none, fill=none, inner sep=1pt, row sep=-1pt},
    legend cell align=left,
    axis x line*=bottom, axis y line*=left,
    clip=false,
  },
  irb line/.style={line width=0.8pt, mark size=1.4pt},
}

\fi

\definecolor{cvprblue}{rgb}{0.21,0.49,0.74}
\usepackage[pagebackref,breaklinks,colorlinks,allcolors=cvprblue]{hyperref}
\hypersetup{
  pdftitle={Rethinking 3DGS for Degraded Inputs in the Era of Foundation Models},
  pdfauthor={Xingyu Meng, Fei Yu, Minyi Liu, Haisen Zhao, Qiong Zeng, and Wenzheng Chen}
}
\usepackage{colortbl}
\definecolor{best1}{RGB}{255,152,152}
\definecolor{best2}{RGB}{255,203,152}
\definecolor{best3}{RGB}{255,247,173}

\definecolor{titleBlue}{HTML}{2675A6}
\definecolor{titleAmber}{HTML}{AA6518}

\iflatexml
  \newcommand{\papertitlebreak}{\space}
\else
  \newcommand{\papertitlebreak}{\\}
\fi

\title{Rethinking 3DGS for \textcolor{titleBlue}{Degraded Inputs}\papertitlebreak in the Era of \textcolor{titleAmber}{Foundation Models}}
\iflatexml\else
\author{Xingyu Meng\textsuperscript{1,*}, Fei Yu\textsuperscript{1,2,*},
  Minyi Liu\textsuperscript{1}, Haisen Zhao\textsuperscript{1},
  Qiong Zeng\textsuperscript{1}, Wenzheng Chen\textsuperscript{2}}
\newcommand{\paperaffiliations}{%
  \textsuperscript{1}Shandong University\qquad
  \textsuperscript{2}Peking University}

\makeatletter
\renewcommand{\@maketitle}{%
  \newpage\null
  \vbox to 135.302pt{%
    \vspace{15pt}%
    \centering
    {\fontsize{17}{19}\selectfont\bfseries \@title\par}%
    \vspace{22.5pt}%
    {\fontsize{12}{14}\selectfont \@author\par}%
    \vspace{3pt}%
    {\fontsize{11.5}{13}\selectfont \paperaffiliations\par}%
    \vfil
  }%
}
\newcommand{\ps@paperfirst}{%
  \ps@plain
  \def\@oddfoot{%
    \rlap{\raisebox{1.5pt}[0pt][0pt]{%
      \shortstack[l]{\rule{0.85in}{0.35pt}\\[1pt]%
        \footnotesize\textsuperscript{*}Equal contribution.}}}%
    \hfil\thepage\hfil}%
  \let\@evenfoot\@oddfoot
}
\makeatother
\AtBeginDocument{%
  \crefname{appendix}{Sec.}{Secs.}%
  \Crefname{appendix}{Section}{Sections}%
  \crefname{subappendix}{Sec.}{Secs.}%
  \Crefname{subappendix}{Section}{Sections}%
}
\fi

\begin{document}
\newcommand{\teaserpanels}{%
  \includegraphics[width=\linewidth,alt={Two diagonal rendering comparisons, a chart of the four task-specific PSNR gains before and after input recovery, and a curve showing when camera updates help.}]{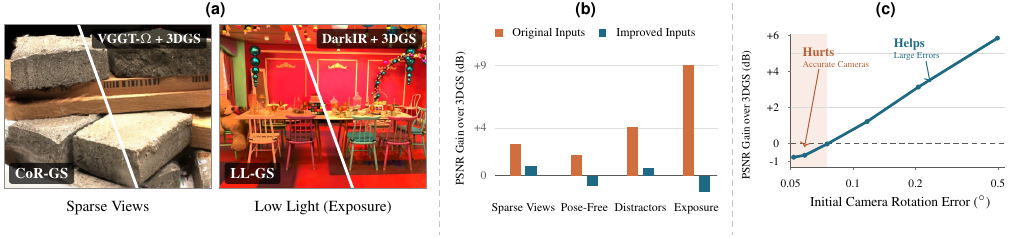}}
\newcommand{\teasercaption}{%
  \textbf{Recovering degraded inputs changes the value of task-specific 3DGS training components.}
      (a) Standard 3DGS with recovered inputs is compared with CoR-GS on sparse views and LL-GS on low-light images, both using their native inputs.
      (b) With matched inputs, the mean PSNR gain of task-specific methods shrinks in all four tasks and becomes negative for unknown poses and poor exposure.
      (c) In a controlled pose experiment, updating cameras improves test PSNR at large initial rotation errors and lowers it when the initial poses are already accurate.}

\iflatexml
  \begin{titlepage}
    \maketitle
    \begin{center}
      \mbox{Xingyu~Meng\textsuperscript{1,*}}\quad
      \mbox{Fei~Yu\textsuperscript{1,2,*}}\quad
      \mbox{Minyi~Liu\textsuperscript{1}}\\
      \mbox{Haisen~Zhao\textsuperscript{1}}\quad
      \mbox{Qiong~Zeng\textsuperscript{1}}\quad
      \mbox{Wenzheng~Chen\textsuperscript{2}}\\
      \mbox{\textsuperscript{1}Shandong University}\qquad
      \mbox{\textsuperscript{2}Peking University}\\
      \textsuperscript{*}Equal contribution.
    \end{center}
    \begin{abstract}
\iflatexml\lxAddAnnotation{:name=Abstract}\fi

\vspace{-10pt}

3D Gaussian Splatting (3DGS) enables high-quality 3D reconstruction, but often struggles with degraded inputs, such as sparse views, unknown camera poses, or corrupted images.
To handle these problems, many methods add depth guidance, camera refinement, or mechanisms for handling unreliable image regions during 3DGS training.
However, recent vision foundation models can now recover much of the missing or unreliable information before 3DGS training, raising the question of whether these additional components are still needed.

To investigate this question, we use pretrained vision models to improve the inputs, then compare standard 3DGS with task-specific methods on both the original and improved inputs.
The PSNR gain of task-specific methods over standard 3DGS shrinks in 338 of 394 matched comparisons (86\%).
On improved inputs, this gain turns negative on average for unknown camera poses and poor exposure.
Why do some components still help while others hurt?

Controlled experiments show that, as the inputs improve, errors that arise during 3DGS training can limit further gains.
Early rendering errors can drive pose or exposure updates that disturb already accurate cameras or restored images.
Some gains that persist after input recovery can be traced to common 3DGS fitting problems, including floating Gaussians and inconsistent colors across views.
These findings motivate training methods that preserve accurate inputs and address the remaining errors in Gaussian fitting.

\end{abstract}

  \end{titlepage}
  \begin{figure*}
    \centering
    \teaserpanels
    \caption{\teasercaption}\label{fig:teaser}
  \end{figure*}
\else
  \definecolor{irbTeaserBlue}{RGB}{30,105,131}
\definecolor{irbTeaserOrange}{RGB}{208,110,61}

\newcommand{\irbTeaserDiagonal}[4]{%
  \begin{tikzpicture}[x=1cm,y=1cm]
    \path[use as bounding box] (0,0) rectangle (3.50,2.78);
    \begin{scope}
      \clip (0,0)--(0,2.78)--(1.25,2.78)--(2.25,0)--cycle;
      \node[inner sep=0pt,anchor=center] at (1.75,1.39)
        {#1};
    \end{scope}
    \begin{scope}
      \clip (1.25,2.78)--(3.50,2.78)--(3.50,0)--(2.25,0)--cycle;
      \node[inner sep=0pt,anchor=center] at (1.75,1.39)
        {#2};
    \end{scope}
    \draw[white,line width=1.35pt] (1.25,2.78)--(2.25,0);
    \node[anchor=south west,rounded corners=1.2pt,fill=black,fill opacity=0.70,
          text=white,text opacity=1,inner xsep=2.8pt,inner ysep=2.5pt,
          font=\fontsize{7.1}{8}\selectfont\bfseries]
          at (0.075,0.075) {#3};
    \node[anchor=north east,rounded corners=1.2pt,fill=black,fill opacity=0.70,
          text=white,text opacity=1,inner xsep=2.7pt,inner ysep=2.4pt,
          font=\fontsize{6.5}{7.4}\selectfont\bfseries]
          at (3.425,2.705) {#4};
    \draw[black!45,line width=0.35pt] (0,0) rectangle (3.50,2.78);
  \end{tikzpicture}%
}
\newcommand{\irbTeaserSectionTitle}[1]{%
  \noindent\makebox[\linewidth][c]{%
    \adjustbox{max width=\linewidth}{%
      \sffamily\fontsize{7.8}{9}\selectfont\bfseries #1}}%
  \par\vspace{2pt}%
}
\newcommand{\irbTeaserDivider}{%
  \begin{minipage}[t]{0.008\textwidth}%
    \vspace{0pt}\centering
    \begin{tikzpicture}[baseline={(0,0)}]
      \draw[gray!45,densely dashed,line width=0.58pt]
        (0,0.28)--(0,-3.65);
    \end{tikzpicture}%
  \end{minipage}%
}

\twocolumn[{%
  \renewcommand\twocolumn[1][]{#1}%
  \maketitle
  \vspace{-13pt}
  \noindent
  \begin{minipage}[t]{0.418\textwidth}
    \vspace{0pt}\centering
    \irbTeaserSectionTitle{(a)}
    \resizebox{0.98\linewidth}{!}{%
      \begin{tabular}{@{}c@{\hspace{4pt}}c@{}}
        \irbTeaserDiagonal
          {\includegraphics[height=2.78cm]{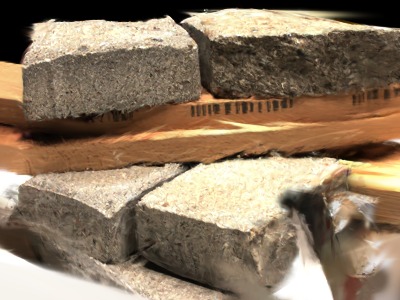}}
          {\includegraphics[height=2.78cm]{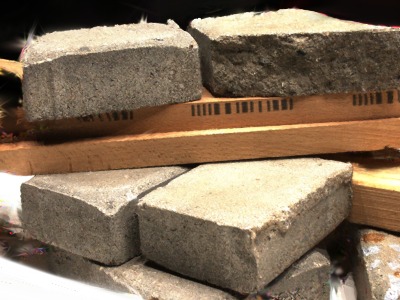}}
          {CoR-GS}{VGGT-$\Omega$ + 3DGS} &
        \irbTeaserDiagonal
          {\includegraphics[height=2.78cm]{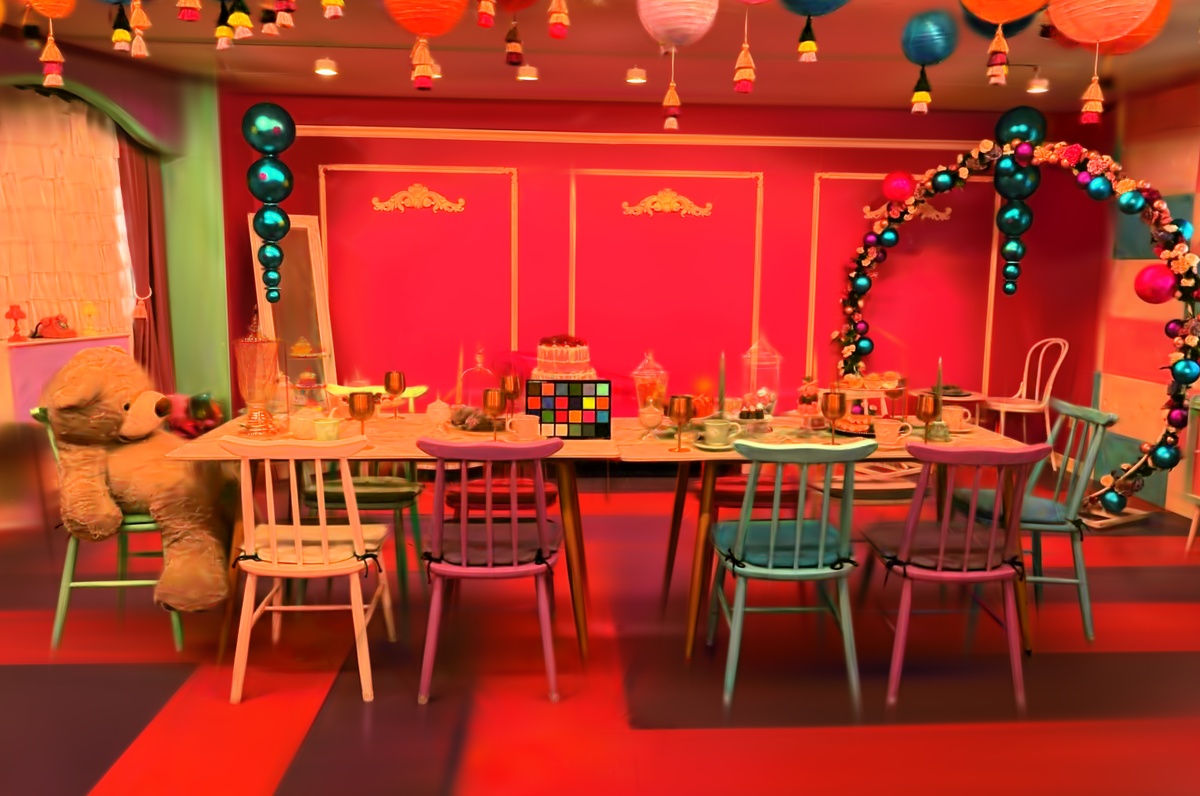}}
          {\includegraphics[height=2.78cm]{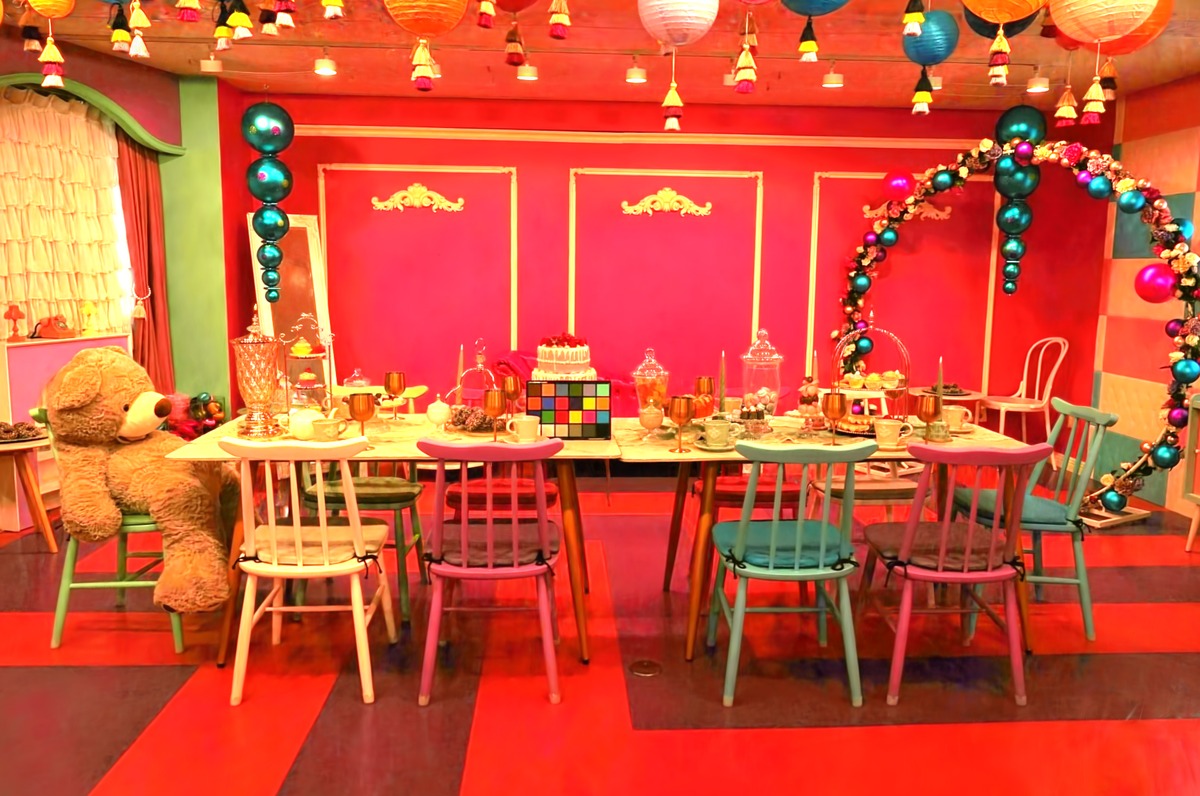}}
          {LL-GS}{DarkIR + 3DGS}\\[-1pt]
        {\fontsize{7.4}{8}\selectfont Sparse Views} &
        {\fontsize{7.4}{8}\selectfont Low Light (Exposure)}
      \end{tabular}%
    }\par
  \end{minipage}\hfill%
  \irbTeaserDivider\hfill%
  \begin{minipage}[t]{0.268\textwidth}
    \vspace{0pt}\centering
    \irbTeaserSectionTitle{(b)}
    \resizebox{\linewidth}{!}{%
    \begin{tikzpicture}[x=1cm,y=1.14cm]
      \path[use as bounding box] (-0.57,-0.34) rectangle (4.86,3.13);
      \fill[irbTeaserOrange] (0.36,2.90) rectangle (0.51,3.04);
      \node[anchor=west,align=center,font=\fontsize{6.7}{7.4}\selectfont]
        at (0.58,2.97) {Original Inputs};
      \fill[irbTeaserBlue] (2.40,2.90) rectangle (2.55,3.04);
      \node[anchor=west,align=center,font=\fontsize{6.7}{7.4}\selectfont]
        at (2.62,2.97) {Improved Inputs};
      \foreach \yy/\val in {2.42/+9,1.34/+4,0.52/0} {%
         \draw[black!12,thin] (0.37,\yy)--(4.74,\yy);
         \node[anchor=east,font=\fontsize{6.5}{7}\selectfont,text=black!65]
            at (0.32,\yy) {\val};
      }
      \draw[black!65,line width=0.6pt] (0.37,0.52)--(4.72,0.52);
      \node[anchor=center,rotate=90,font=\fontsize{6.5}{7.5}\selectfont]
        at (-0.31,1.50) {PSNR Gain over 3DGS (dB)};
      \foreach \xx/\native/\recovered/\task in {
        0.92/2.57/0.78/Sparse Views,
        2.12/1.69/-0.89/Pose-Free,
        3.22/3.99/0.61/Distractors,
        4.32/9.11/-1.38/Exposure
      } {%
        \pgfmathsetmacro{\yNative}{0.52+0.21*\native}
        \pgfmathsetmacro{\yRecovered}{0.52+0.21*\recovered}
        \fill[irbTeaserOrange] ({\xx-0.26},0.52) rectangle ({\xx-0.04},\yNative);
        \fill[irbTeaserBlue] ({\xx+0.04},0.52) rectangle ({\xx+0.26},\yRecovered);
        \node[font=\fontsize{6.6}{7.5}\selectfont,anchor=north]
          at (\xx,0.15) {\task};
      }

    \end{tikzpicture}%
    }\par

  \end{minipage}\hfill%
  \irbTeaserDivider\hfill%
  \begin{minipage}[t]{0.282\textwidth}
    \vspace{0pt}\centering
    \irbTeaserSectionTitle{(c)}
    \resizebox{0.97\linewidth}{!}{%
    \begin{tikzpicture}[x=1cm,y=1cm]
      \path[use as bounding box] (-0.34,-0.41) rectangle (5.14,3.11);
      \fill[irbTeaserOrange!13] (0.55,0.32) rectangle (1.26,2.92);
      \draw[black!15,line width=0.35pt] (0.55,1.485)--(4.72,1.485);
      \draw[black!15,line width=0.35pt] (0.55,2.535)--(4.72,2.535);
      \draw[black!65,line width=0.8pt] (0.55,0.32)--(0.55,2.92);
      \draw[black!65,line width=0.8pt] (0.55,0.32)--(4.72,0.32);
      \draw[black!65,densely dashed,line width=0.6pt] (0.55,0.785)--(4.72,0.785);
      \foreach \yy/\label in {0.435/-1,0.785/0,1.485/+2,2.185/+4,2.885/+6} {%
        \draw[black!55,thin] (0.50,\yy)--(0.55,\yy);
        \node[anchor=east,font=\fontsize{6.3}{7}\selectfont] at (0.45,\yy) {\label};
      }
      \foreach \xx/\label in {0.55/0.05,1.763/0.1,2.976/0.2,4.58/0.5} {%
        \draw[black!55,thin] (\xx,0.32)--(\xx,0.27);
        \node[font=\fontsize{6.3}{7}\selectfont,anchor=north] at (\xx,0.22) {\label};
      }
      \draw[irbTeaserBlue,line width=1.5pt,line cap=round,line join=round]
        (0.609,0.525)--(0.825,0.563)--(1.260,0.782)
        --(2.036,1.217)--(3.032,1.888)--(4.568,2.838);
      \foreach \xx/\yy in {0.609/0.525,0.825/0.563,1.260/0.782,2.036/1.217,3.032/1.888,4.568/2.838} {%
        \fill[irbTeaserBlue] (\xx,\yy) circle (1.4pt);
      }
      \node[font=\fontsize{7}{8}\selectfont\bfseries,text=irbTeaserOrange!88!black,
         anchor=west] at (0.66,2.56) {Hurts};
      \node[font=\fontsize{7}{8}\selectfont,text=irbTeaserOrange!88!black,
         anchor=west] at (0.66,2.33) {\tiny{Accurate Cameras}};
      \draw[->,irbTeaserOrange!85!black,line width=0.55pt] (1.10,2.24)--(0.82,0.72);
      \node[font=\fontsize{7}{8}\selectfont\bfseries,text=irbTeaserBlue,
        anchor=west] at (2.43,2.72) {Helps};
      \node[font=\fontsize{7}{8}\selectfont,text=irbTeaserBlue,
        anchor=west] at (2.43,2.49) {\tiny{Large Errors}};
      \draw[->,irbTeaserBlue,line width=0.55pt] (3.18,2.39)--(3.26,1.98);
      \node[font=\fontsize{7.5}{8}\selectfont,anchor=north]
        at (2.68,-0.10) {Initial Camera Rotation Error ($^\circ$)};
      \node[font=\fontsize{6.2}{7.2}\selectfont,anchor=center,rotate=90]
        at (-0.30,1.34) {PSNR Gain over 3DGS (dB)};
    \end{tikzpicture}%
    }\par

  \end{minipage}
  \par\vspace{-7pt}
  {\fontsize{8}{9.5}\selectfont
    \captionsetup{hypcap=false}
    \captionof{figure}{\teasercaption}
    \label{fig:teaser}
  }
  \vspace{0pt}
}]

  \thispagestyle{paperfirst}
  \begin{abstract}
\iflatexml\lxAddAnnotation{:name=Abstract}\fi

\vspace{-10pt}

3D Gaussian Splatting (3DGS) enables high-quality 3D reconstruction, but often struggles with degraded inputs, such as sparse views, unknown camera poses, or corrupted images.
To handle these problems, many methods add depth guidance, camera refinement, or mechanisms for handling unreliable image regions during 3DGS training.
However, recent vision foundation models can now recover much of the missing or unreliable information before 3DGS training, raising the question of whether these additional components are still needed.

To investigate this question, we use pretrained vision models to improve the inputs, then compare standard 3DGS with task-specific methods on both the original and improved inputs.
The PSNR gain of task-specific methods over standard 3DGS shrinks in 338 of 394 matched comparisons (86\%).
On improved inputs, this gain turns negative on average for unknown camera poses and poor exposure.
Why do some components still help while others hurt?

Controlled experiments show that, as the inputs improve, errors that arise during 3DGS training can limit further gains.
Early rendering errors can drive pose or exposure updates that disturb already accurate cameras or restored images.
Some gains that persist after input recovery can be traced to common 3DGS fitting problems, including floating Gaussians and inconsistent colors across views.
These findings motivate training methods that preserve accurate inputs and address the remaining errors in Gaussian fitting.

\end{abstract}

\fi

\vspace{-15pt}

\section{Introduction}
\label{sec:intro}

\vspace{-5pt}

3D Gaussian Splatting (3DGS)~\cite{kerbl2023gaussians} enables high-quality 3D reconstruction, but often struggles with degraded inputs, such as sparse views, unknown camera poses, or corrupted images.
To handle these problems, many methods add depth guidance, camera refinement, or mechanisms for handling unreliable images during 3DGS training~\cite{li2024dngaussian,fu2024colmapfree,wang2025desplat,cui2025luminance}.
These methods can substantially improve reconstruction on degraded inputs.

When reliable inputs were difficult to obtain before reconstruction, it was natural to address their defects during 3DGS training.
However, recent advances in vision foundation models have changed this situation.
Geometric foundation models can now estimate scene geometry and camera poses from images~\cite{wang2024dust3r,wang2025vggt,wang2026vggtomega}, while segmentation and image restoration models can identify temporary occlusions or improve corrupted images~\cite{carion2026sam3,feijoo2025darkir}.
With these models, we can now improve the inputs first and then train standard 3DGS without adding those extra components.
Once the inputs improve, how much do the task-specific training components still help?

To investigate this question, we study 148 test cases across four settings: sparse views~\cite{zhu2024fsgs}, unknown camera poses~\cite{fu2024colmapfree}, temporary occlusions~\cite{sabour2025spotless}, and poor exposure~\cite{cui2025luminance}.
For each setting, we use pretrained vision models to improve the inputs and then run standard 3DGS without changing its optimizer.
We find that this simple approach completes all 148 test cases and gives results that are competitive with methods designed for each setting (\cref{fig:teaser,tab:main}).
To isolate the effect of the extra components, we then give the task-specific methods and standard 3DGS the same recovered information for each task.
The PSNR gain of the task-specific methods shrinks in 338 of 394 matched comparisons (86\%, \cref{tab:shared}).
For unknown camera poses and poor exposure, the average gain turns negative.

Why do some extra components now hurt reconstruction, and where do the remaining gains come from?
Our controlled experiments show that, as inputs become more accurate, \textbf{errors that arise during 3DGS training can limit reconstruction quality}.
Some components give the optimizer the freedom to refine camera poses or modify the images used for supervision.
When the inputs are poor, these additional degrees of freedom can correct errors in the cameras or images.
Early in training, however, even accurate inputs produce rendering residuals because the Gaussians have not yet learned the scene. Updates that reduce those residuals can move cameras away from accurate poses or distort a restored image.

In our probes, pose optimization improves coarse camera estimates but degrades accurate ones, while a learnable exposure target improves reconstruction from dark images but reduces the contrast of restored targets (\cref{fig:fig5_crossover}).

For components that still help after input recovery, part of the remaining advantage comes from \textbf{mitigating weaknesses in standard 3DGS training itself}.
Consider the temporary-occlusion task, in which people or other objects obscure the static scene in some training views.
One possible explanation for the remaining advantage is that task-specific methods identify temporary objects missed by the recovered masks.
When we exclude more regions suspected of containing missed objects, however, the mean advantage shrinks only slightly in six diagnostic scenes.
In several of these scenes, we identify near-camera floating Gaussians or inconsistent colors across views.
Keeping the masks fixed and changing the standard densification strategy to 3DGS-MCMC~\cite{kheradmand20243d} closes over two-thirds of the mean PSNR gap to RobustSplat across the six scenes (\cref{fig:fig5_survive}).

Together, these findings suggest a further question for 3DGS: how can training make better use of the information that pretrained vision models have already recovered?
This calls for better 3DGS training methods that can correct remaining input errors without disturbing accurate inputs, while also addressing problems in Gaussian fitting itself.
We believe improving this shared training process is a promising direction for future research.

\vspace{-5pt}
\section{Related Work}
\label{sec:related-work}

\iflatexml\else
\definecolor{irbFigInk}{RGB}{29,35,40}
\definecolor{irbFigBlue}{RGB}{20,75,108}
\definecolor{irbFigRust}{RGB}{156,71,39}
\newcommand{\irbFigPhoto}[5]{%
  \begin{scope}
    \clip ({#1-#3/2},{#2-#4/2}) rectangle ({#1+#3/2},{#2+#4/2});
    \node[inner sep=0pt,outer sep=0pt,anchor=center] at (#1,#2)
      {#5};
  \end{scope}
  \draw[irbFigInk,line width=0.55pt]
     ({#1-#3/2},{#2-#4/2}) rectangle ({#1+#3/2},{#2+#4/2});
}
\newcommand{\irbFigSplat}[2]{%
 \begin{scope}[shift={(#1,#2)}]
   \foreach \xx/\yy/\rx/\ang/\cc in {
     -0.42/-0.06/0.20/28/irbFigBlue,
     -0.17/0.24/0.24/-17/irbFigRust,
      0.16/-0.23/0.24/13/irbFigBlue,
      0.40/0.15/0.20/37/irbFigRust,
     -0.37/0.30/0.12/25/irbFigInk,
      0.04/0.12/0.15/-28/irbFigInk} {
     \filldraw[draw=\cc,fill=\cc!28,opacity=0.85,line width=0.48pt,
          rotate around={\ang:(\xx,\yy)}]
        (\xx,\yy) ellipse[x radius=\rx,y radius=0.56*\rx];
   }
 \end{scope}
}
\newcommand{\irbFigCamera}[3]{%
 \begin{scope}[shift={(#1,#2)},scale=#3]
    \fill[irbFigInk] (-0.42,-0.20) rectangle (-0.13,0.20);
    \draw[irbFigInk,line width=0.72pt]
       (-0.13,-0.19)--(0.45,-0.38)--(0.45,0.38)--(-0.13,0.19)--cycle;
    \draw[irbFigBlue,line width=0.72pt] (0.45,-0.38)--(0.45,0.38);
 \end{scope}
}
\fi
\begin{figure*}[t]
\centering
\iflatexml
\includegraphics[width=\linewidth,alt={Four degraded-input tasks, the recovered training signals, and three experiments comparing reconstruction methods.}]{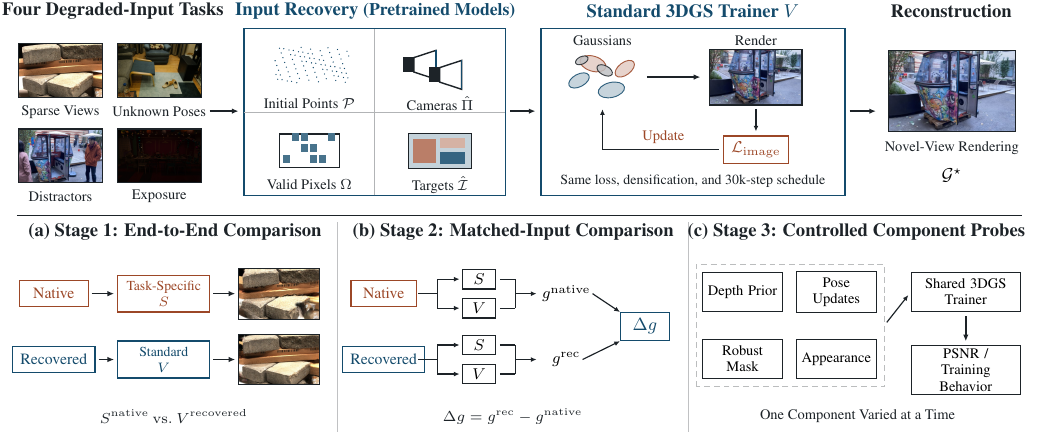}
\else
\resizebox{0.985\textwidth}{!}{%
\begin{tikzpicture}[x=1cm,y=1cm,
    every node/.style={font=\rmfamily\fontsize{8.0}{9.0}\selectfont,inner sep=1pt},
    arrow/.style={-{Latex[length=3.4pt,width=2.8pt]},draw=irbFigInk,line width=0.73pt},
    thinarr/.style={-{Latex[length=2.9pt,width=2.5pt]},draw=irbFigInk,line width=0.60pt},
    heading/.style={font=\rmfamily\fontsize{9.5}{10.4}\selectfont\bfseries,text=irbFigInk},
    captionlab/.style={font=\rmfamily\fontsize{7.4}{8.4}\selectfont,text=irbFigInk},
    rulebox/.style={draw=irbFigInk,line width=0.62pt,fill=white,inner xsep=4pt,inner ysep=3pt,align=center},
    nativebox/.style={rulebox,draw=irbFigRust,text=irbFigRust},
    recbox/.style={rulebox,draw=irbFigBlue,text=irbFigBlue},
  ]
  \path[use as bounding box] (0.00,0.25) rectangle (18.20,8.30);

  \node[heading] at (1.82,8.08) {Four Degraded-Input Tasks};
  \irbFigPhoto{0.88}{7.02}{1.51}{0.96}
    {\includegraphics[width=1.51cm]{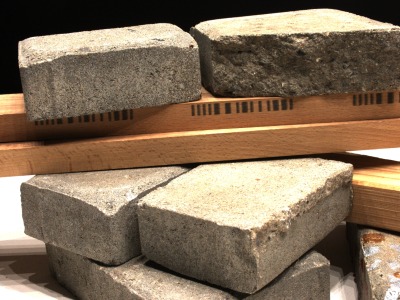}}
  \irbFigPhoto{2.65}{7.02}{1.51}{0.96}
    {\includegraphics[width=1.51cm]{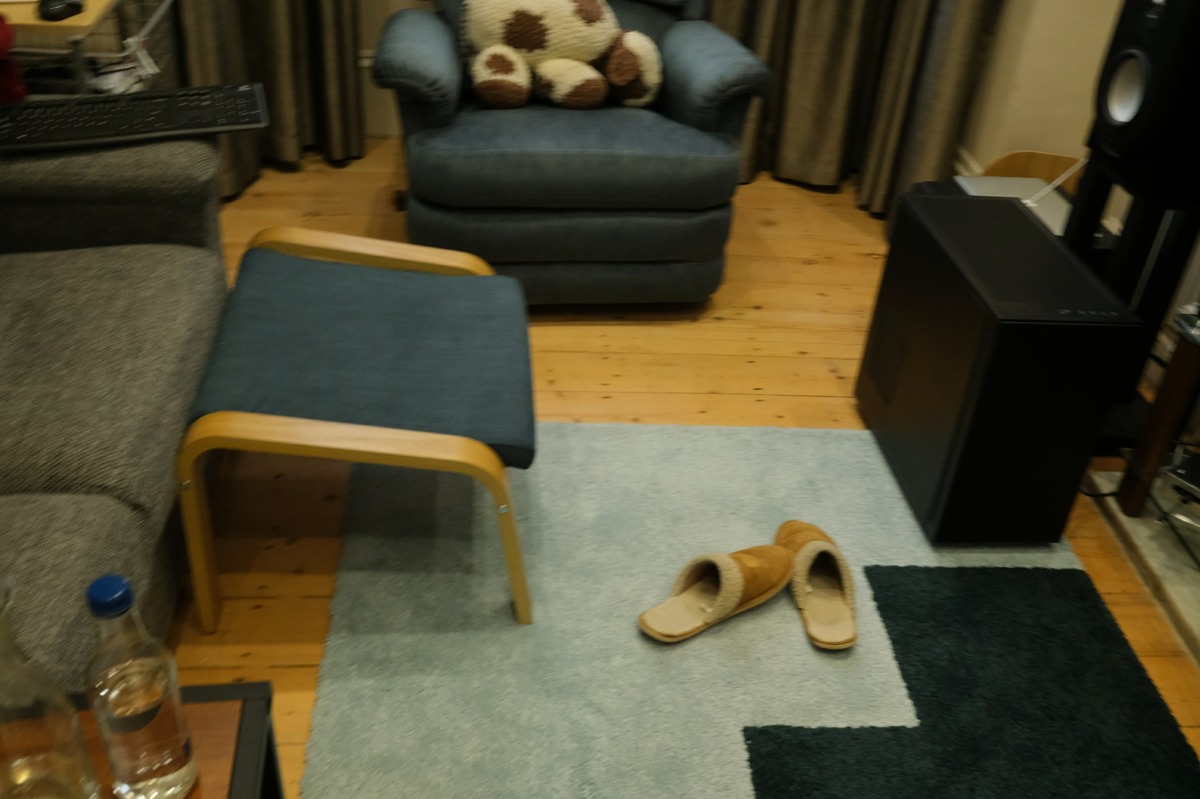}}
  \irbFigPhoto{0.88}{5.49}{1.51}{0.96}
    {\includegraphics[width=1.51cm]{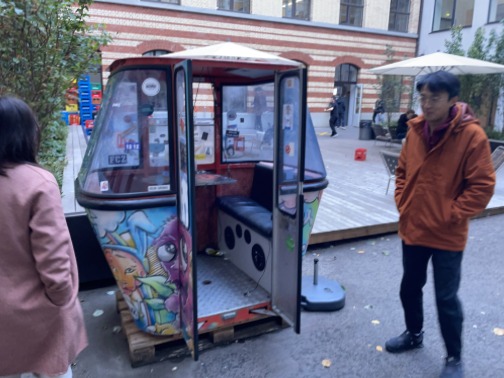}}
  \irbFigPhoto{2.65}{5.49}{1.51}{0.96}
    {\includegraphics[width=1.51cm]{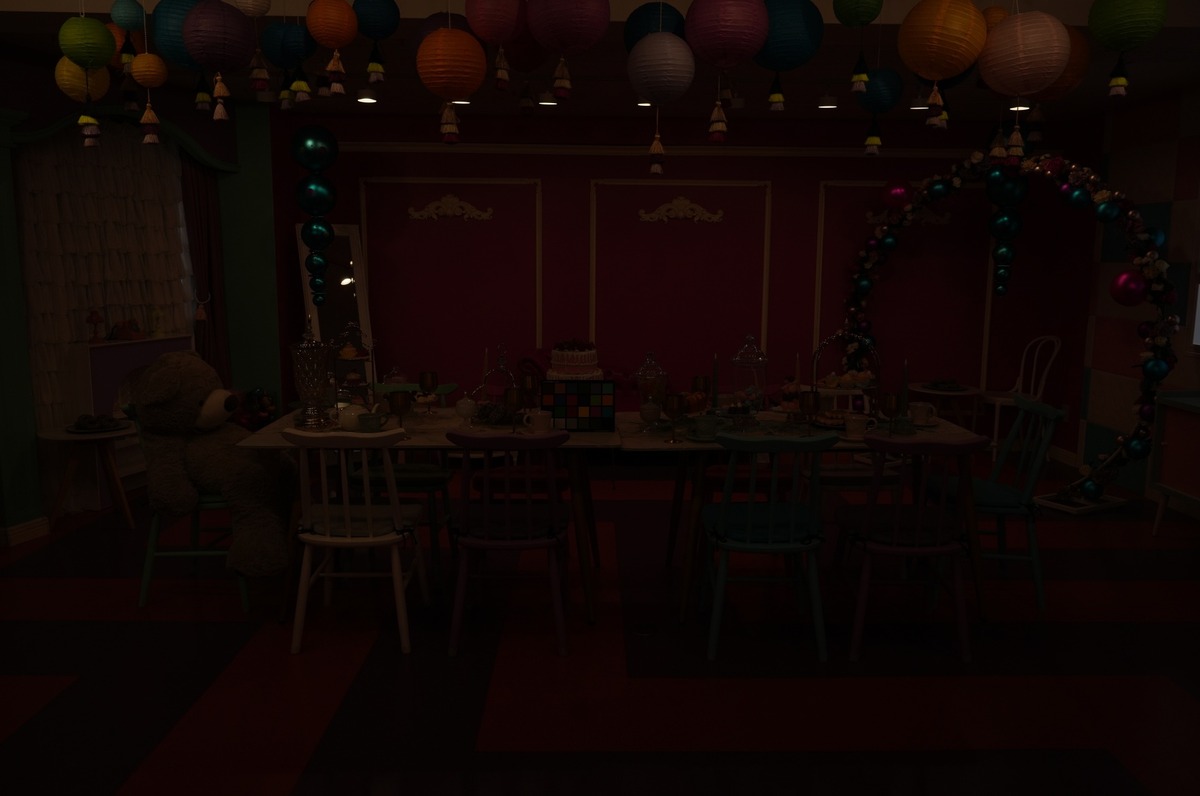}}
  \node[captionlab] at (0.88,6.28) {Sparse Views};
  \node[captionlab] at (2.65,6.28) {Unknown Poses};
  \node[captionlab] at (0.88,4.76) {Distractors};
  \node[captionlab] at (2.65,4.76) {Exposure};
  \draw[arrow] (3.55,6.29)--(4.07,6.29);

  \node[heading,text=irbFigBlue] at (6.52,8.08) {Input Recovery {\small{(Pretrained Models)}}};
  \draw[irbFigBlue,line width=0.85pt] (4.17,4.78) rectangle (8.85,7.77);
  \draw[irbFigInk!34,line width=0.47pt] (6.51,4.78)--(6.51,7.77);
  \draw[irbFigInk!34,line width=0.47pt] (4.17,6.26)--(8.85,6.26);
  \foreach \i in {1,...,83} {
    \pgfmathsetmacro{\xx}{4.71+mod(\i*37,87)/87*1.35}
    \pgfmathsetmacro{\yy}{6.80+mod(\i*49,83)/83*0.64}
    \fill[irbFigBlue] (\xx,\yy) circle (0.012);
  }
  \node[captionlab] at (5.34,6.43) {Initial Points $\mathcal P$};
  \irbFigCamera{7.30}{7.12}{0.68}
  \irbFigCamera{7.76}{6.94}{0.68}
  \node[captionlab] at (7.68,6.43) {Cameras $\hat\Pi$};
  \begin{scope}[shift={(5.34,5.58)}]
    \draw[irbFigInk,line width=0.57pt] (-0.54,-0.30) rectangle (0.54,0.30);
    \foreach \i in {0,...,5} \foreach \j in {0,...,2} {
      \pgfmathtruncatemacro{\on}{mod(\i*7+\j*11,4)}
      \ifnum\on<2
        \fill[irbFigBlue!78] ({-0.47+\i*0.16},{-0.22+\j*0.20}) rectangle
                            ({-0.34+\i*0.16},{-0.08+\j*0.20});
      \fi
    }
  \end{scope}
  \node[captionlab] at (5.34,4.97) {Valid Pixels $\Omega$};
  \begin{scope}[shift={(7.68,5.58)}]
    \fill[irbFigBlue!25] (-0.57,-0.31) rectangle (0.57,0.31);
    \fill[irbFigRust!70] (-0.49,-0.22) rectangle (-0.07,0.21);
    \fill[irbFigBlue!86] (0.00,-0.23) rectangle (0.45,-0.02);
    \fill[irbFigRust!42] (0.00,0.05) rectangle (0.45,0.23);
    \draw[irbFigInk,line width=0.57pt] (-0.57,-0.31) rectangle (0.57,0.31);
  \end{scope}
  \node[captionlab] at (7.68,4.97) {Targets $\hat{\mathcal I}$};
  \draw[arrow] (8.94,6.29)--(9.39,6.29);

  \node[heading,text=irbFigBlue] at (12.21,8.08) {Standard 3DGS Trainer $V$};
  \draw[irbFigBlue,line width=0.85pt] (9.49,4.78) rectangle (14.94,7.77);
  \irbFigSplat{10.59}{6.90}
  \node[captionlab] at (10.59,7.55) {Gaussians};
  \draw[arrow,shorten <=1pt,shorten >=1pt]
    (11.35,6.90)--(12.38,6.90);
  \irbFigPhoto{13.34}{6.90}{1.65}{1.00}
    {\includegraphics[width=1.65cm]{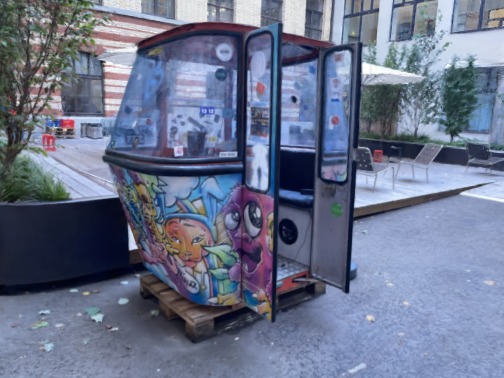}}
  \node[captionlab] at (13.34,7.55) {Render};
  \node[rulebox,draw=irbFigRust,text=irbFigRust,
        minimum width=1.10cm,minimum height=0.42cm]
        (loss) at (13.34,5.60) {$\mathcal L_{\mathrm{image}}$};
  \coordinate (irbRenderBottom) at (13.34,6.40);
  \coordinate (irbGaussianBottom) at (10.59,6.36);
  \draw[thinarr,shorten <=2pt,shorten >=2pt]
    (irbRenderBottom)--(loss.north);
  \draw[thinarr,shorten <=2pt,shorten >=1pt]
    (loss.west)--(10.59,5.60)--(irbGaussianBottom);
  \node[captionlab,text=irbFigRust] at (11.68,5.83) {Update};
  \node[font=\rmfamily\fontsize{7.1}{8.0}\selectfont,text=irbFigInk]
     at (12.21,5.03) {Same loss, densification, and 30k-step schedule};
  \draw[arrow] (15.02,6.29)--(15.49,6.29);

  \node[heading] at (16.84,8.08) {Reconstruction};
  \irbFigPhoto{16.85}{6.65}{2.27}{1.46}
    {\includegraphics[width=2.27cm]{figures/qualitative/t3_transient_onthego_patio_high/ours.jpg}}
  \node[captionlab] at (16.85,5.62) {Novel-View Rendering};
  \node[font=\rmfamily\fontsize{8.8}{9.4}\selectfont]
    at (16.85,5.14) {$\mathcal G^\star$};

  \draw[irbFigInk,line width=0.68pt] (0.10,4.42)--(18.10,4.42);
  \draw[irbFigInk!28,line width=0.52pt] (5.84,0.55)--(5.84,4.30);
  \draw[irbFigInk!28,line width=0.52pt] (12.13,0.55)--(12.13,4.30);
  \node[heading] at (2.93,4.13) {(a) Stage 1: End-to-End Comparison};
  \node[heading] at (8.99,4.13) {(b) Stage 2: Matched-Input Comparison};
  \node[heading] at (15.14,4.13) {(c) Stage 3: Controlled Component Probes};

  \node[nativebox,minimum width=1.23cm,minimum height=0.46cm]
        (irbANative) at (0.76,3.02) {Native};
  \node[recbox,minimum width=1.23cm,minimum height=0.46cm]
        (irbARecovered) at (0.76,1.84) {Recovered};

  \node[nativebox,align=center,font=\scriptsize,
        text width=1.45cm,inner xsep=2pt,inner ysep=2pt,
        minimum width=1.65cm,minimum height=0.66cm]
        (irbASpecialist) at (2.73,3.02) {Task-Specific\\$S$};
  \node[recbox,align=center,font=\scriptsize,
        text width=1.45cm,inner xsep=2pt,inner ysep=2pt,
        minimum width=1.65cm,minimum height=0.66cm]
        (irbAStandard) at (2.73,1.84) {Standard\\$V$};

  \draw[arrow,shorten <=1.5pt,shorten >=1.5pt]
        (irbANative.east)--(irbASpecialist.west);
  \draw[arrow,shorten <=1.5pt,shorten >=1.5pt]
        (irbARecovered.east)--(irbAStandard.west);

  \irbFigPhoto{4.79}{3.02}{1.45}{0.91}
    {\includegraphics[width=1.45cm]{figures/qualitative/t1_sparse_dtu_scan38/specialist.jpg}}
  \irbFigPhoto{4.79}{1.84}{1.45}{0.91}
    {\includegraphics[width=1.45cm]{figures/qualitative/t1_sparse_dtu_scan38/ours.jpg}}

  \coordinate (irbANativeOutput) at (4.065,3.02);
  \coordinate (irbARecoveredOutput) at (4.065,1.84);
  \draw[arrow,shorten <=1.5pt,shorten >=1.5pt]
        (irbASpecialist.east)--(irbANativeOutput);
  \draw[arrow,shorten <=1.5pt,shorten >=1.5pt]
        (irbAStandard.east)--(irbARecoveredOutput);

  \node[captionlab] at (2.91,0.82)
    {$S^{\mathrm{native}}\ \mathrm{vs.}\ V^{\mathrm{recovered}}$};

  \node[nativebox,minimum width=1.17cm,minimum height=0.46cm] at (6.67,3.02)
        {Native};
  \node[recbox,minimum width=1.17cm,minimum height=0.46cm] at (6.67,1.84)
        {Recovered};
  \foreach \yy in {3.02,1.84} {
    \draw[irbFigInk,line width=0.60pt] (7.29,\yy)--(7.63,\yy);
    \draw[irbFigInk,line width=0.60pt]
          (7.63,{\yy-0.26})--(7.63,{\yy+0.26});
    \draw[thinarr] (7.63,{\yy+0.26})--(8.03,{\yy+0.26});
    \draw[thinarr] (7.63,{\yy-0.26})--(8.03,{\yy-0.26});
    \node[rulebox,minimum width=0.59cm,minimum height=0.35cm,
          inner xsep=2pt,inner ysep=1pt]
       at (8.38,{\yy+0.26}) {$S$};
    \node[rulebox,minimum width=0.59cm,minimum height=0.35cm,
          inner xsep=2pt,inner ysep=1pt]
       at (8.38,{\yy-0.26}) {$V$};
    \draw[irbFigInk,line width=0.55pt] (8.75,{\yy+0.26})--(9.03,{\yy+0.26})
          (8.75,{\yy-0.26})--(9.03,{\yy-0.26})
          (9.03,{\yy-0.26})--(9.03,{\yy+0.26});
    \draw[thinarr] (9.03,\yy)--(9.41,\yy);
  }
  \node[font=\rmfamily\fontsize{8.0}{9.0}\selectfont] (gn) at (9.94,3.02)
        {$g^{\mathrm{native}}$};
  \node[font=\rmfamily\fontsize{8.0}{9.0}\selectfont] (gr) at (9.94,1.84)
        {$g^{\mathrm{rec}}$};
  \node[recbox,minimum width=0.87cm,minimum height=0.50cm,
        font=\rmfamily\fontsize{9.0}{9.4}\selectfont\bfseries]
        (delta) at (11.35,2.43) {$\Delta g$};
  \draw[thinarr] (gn.east)--(delta.north west);
  \draw[thinarr] (gr.east)--(delta.south west);
  \node[captionlab] at (8.98,0.82)
        {$\Delta g=g^{\mathrm{rec}}-g^{\mathrm{native}}$};

  \draw[irbFigInk!30,densely dashed,line width=0.45pt]
    (12.27,1.35) rectangle (15.64,3.55);
  \node[rulebox,minimum width=1.41cm,minimum height=0.70cm,
        text width=1.30cm,inner xsep=2pt,align=center,
        font=\rmfamily\fontsize{7.4}{8.3}\selectfont]
        at (13.10,3.05) {Depth Prior};
  \node[rulebox,minimum width=1.41cm,minimum height=0.70cm,
        text width=1.30cm,inner xsep=2pt,align=center,
        font=\rmfamily\fontsize{7.4}{8.3}\selectfont]
        at (14.78,3.05) {Pose Updates};
  \node[rulebox,minimum width=1.41cm,minimum height=0.70cm,
        text width=1.30cm,inner xsep=2pt,align=center,
        font=\rmfamily\fontsize{7.4}{8.3}\selectfont]
        at (13.10,1.85) {Robust Mask};
  \node[rulebox,minimum width=1.41cm,minimum height=0.70cm,
        text width=1.30cm,inner xsep=2pt,align=center,
        font=\rmfamily\fontsize{7.4}{8.3}\selectfont]
        at (14.78,1.85) {Appearance};

  \node[rulebox,minimum width=1.95cm,minimum height=0.70cm,
        text width=1.75cm,inner xsep=2pt,inner ysep=1.5pt,
        font=\rmfamily\fontsize{7.4}{8.3}\selectfont,align=center]
        (trainer) at (17.10,3.05) {Shared 3DGS\\Trainer};
  \node[rulebox,minimum width=1.95cm,minimum height=0.70cm,
        text width=1.75cm,inner xsep=2pt,inner ysep=1.5pt,
        font=\rmfamily\fontsize{7.4}{8.3}\selectfont,align=center]
        (measure) at (17.10,1.65) {PSNR / Training\\Behavior};
  \coordinate (irbComponentOutput) at (15.64,2.45);
  \draw[thinarr,shorten <=2pt,shorten >=2pt]
        (irbComponentOutput)--(trainer.west);
  \draw[arrow,shorten <=2pt,shorten >=2pt]
        (trainer.south)--(measure.north);
  \node[captionlab] at (15.16,0.82) {One Component Varied at a Time};
\end{tikzpicture}%
}
\fi
\vspace{-10pt}
\caption{\textbf{Overview of our study design separating input recovery from 3DGS training.}
  Depending on the task, pretrained vision models recover initial points $\mathcal{P}$, camera poses $\hat{\Pi}$, valid pixel masks $\Omega$, or target images $\hat{\mathcal{I}}$ before standard 3DGS training.
  (a) \textbf{Stage 1} compares task-specific methods on the original degraded inputs ($S^{\text{native}}$) against standard 3DGS on recovered inputs ($V^{\text{recovered}}$).
  (b) \textbf{Stage 2} compares the relative PSNR gain ($g$) under native inputs and under matched recovered inputs.
  (c) \textbf{Stage 3} tests individual components inside a shared trainer to explain why some hurt while others help.}
\label{fig:study-pipeline}

\vspace{-15pt}
\end{figure*}

\subsection{3D Reconstruction for Degraded Inputs}

Standard 3D Gaussian Splatting (3DGS)~\cite{kerbl2023gaussians} reconstructs a scene from images, calibrated cameras, and an initial point cloud.
When these inputs are incomplete, task-specific methods introduce additional constraints or modify scene optimization, including geometry regularization and frequency annealing~\cite{niemeyer2022regnerf,yang2023freenerf}.
For sparse views, depth priors are widely used~\cite{deng2022dsnerf,wang2023sparsenerf,chung2024depthreggs}, as in DNGaussian~\cite{li2024dngaussian} and FSGS~\cite{zhu2024fsgs}, while CoR-GS~\cite{zhang2024corgs} regularizes agreement between Gaussian fields and DropGaussian~\cite{park2025dropgaussian} uses stochastic dropout.
Related work uses stereo and feature correspondences~\cite{han2024binocular3dgs,stereogs,peng2024scgaussian,gu2026sparsesurf}.
These designs, together with structured updates and density-aware regularization~\cite{paliwal2024coherentgs,song2026d2gs}, address geometric ambiguity that is poorly constrained by the observed views alone.

When camera poses are unknown, pose and scene estimation become coupled.
BARF~\cite{lin2021barf} and NoPe-NeRF~\cite{bian2023nopenerf} optimize camera parameters together with radiance fields, while SPARF~\cite{truong2023sparf} uses multiview correspondences to refine camera estimates.
For 3DGS, CF-3DGS~\cite{fu2024colmapfree}, HT-3DGS~\cite{ji2025htgs}, and 3R-GS~\cite{huang2025threergs} combine pose recovery with Gaussian fitting, while GloSplat~\cite{xiong2026glosplat} uses SfM tracks as geometric anchors.
These methods differ in the constraints and initialization available to pose optimization.

Unreliable images introduce a separate challenge.
For transient distractors, methods use appearance and uncertainty cues~\cite{martinbrualla2021nerfw,sabour2023robustnerf,ren2024onthego,zhang2024gsw} or static--transient separation~\cite{lin2025hybridgs,li2025asymgs}, while SpotLessSplats~\cite{sabour2025spotless}, WildGaussians~\cite{kulhanek2024wildgaussians}, DeSplat~\cite{wang2025desplat}, and RobustSplat~\cite{fu2025robustsplat} provide representative Gaussian formulations.
Under adverse exposure, methods explore sensor-aware and illumination-based reconstruction~\cite{mildenhall2022rawnerf,wang2023llnerf}, as well as attenuation and appearance modeling in Aleth-NeRF~\cite{cui2024aleth}, Luminance-GS~\cite{cui2025luminance}, LITA-GS~\cite{zhou2025litags}, LL-Gaussian~\cite{sun2025llgaussian}, and related Gaussian methods~\cite{ye2024gaussiandk,wang2025llgs}.
Together, these lines of work show how task-specific optimization compensates for unreliable geometric and photometric inputs.

\vspace{-5pt}
\subsection{Input Recovery and Reconstruction Evaluation}

\vspace{-3pt}

Recent vision models make it possible to recover useful scene information before 3DGS training begins.
DUSt3R~\cite{wang2024dust3r}, MASt3R~\cite{leroy2024mast3r}, VGGT~\cite{wang2025vggt}, and VGGT-$\Omega$~\cite{wang2026vggtomega} recover geometric information directly from images, alongside other feed-forward approaches~\cite{cabon2025must3r,yang2025fast3r,keetha2026mapanything}.
VGGSfM~\cite{wang2024vggsfm}, MASt3R-SfM~\cite{duisterhof2025mast3rsfm}, and related methods~\cite{pan2026global} further combine learned predictions with geometric optimization.
For unreliable images, SAM~\cite{ravi2025sam2,carion2026sam3} and DINOv2~\cite{oquab2023dinov2} provide segmentation and visual features, while DarkIR~\cite{feijoo2025darkir}, Retinexformer~\cite{cai2023retinexformer}, and LCDPNet~\cite{wang2022lcdpnet} help restore image quality.

These advances have already influenced 3DGS reconstruction.
InstantSplat~\cite{fan2024instantsplat} and ZeroGS~\cite{chen2024zerogs} use learned geometric predictions to reconstruct scenes from sparse or unposed images, while other methods improve the initial geometry through point-cloud augmentation and multiview information~\cite{dai2025eapgs,jang2025comapgs}.
NexusGS~\cite{zheng2025nexusgs} constructs stronger sparse-view geometry using epipolar depth priors, and studies of initialization~\cite{zhou2026initialize,desiatov2026role} further show that the starting geometry influences subsequent 3DGS training.
Other work addresses distractors and difficult lighting conditions~\cite{gu2026sparseviewdistractor,zhu2026nakags,liu2026eloggs}.
Meanwhile, NerfBaselines~\cite{kulhanek2026nerfbaselines}, RealX3D~\cite{realx3d2025}, and related benchmarks~\cite{liu2026ntire3drr,lin2025hqgs,yoon2026resplat,cong2025e3dbench} support broader and more consistent reconstruction evaluation.

These studies show that improving the inputs can substantially change the results of 3DGS reconstruction.
Building on these studies, we ask how input recovery changes the value of task-specific 3DGS training components.

\vspace{3pt}

\section{Study Design}
\label{sec:method}

\vspace{-4pt}

We now describe our study design.
To understand whether extra training components are still needed in the era of foundation models, our core idea is to \textbf{separate the benefits of improving the inputs from those of modifying 3DGS training}.
As illustrated in \cref{fig:study-pipeline}, we study four representative degraded-input tasks: sparse views~\cite{zhu2024fsgs}, unknown camera poses~\cite{fu2024colmapfree}, temporary occlusions (distractors)~\cite{sabour2025spotless}, and poor exposure~\cite{cui2025luminance}.
In \cref{sec:method:separation}, we first review the inputs required by standard 3DGS and describe how we use pretrained vision models to recover the degraded inputs before 3DGS training.
Then, in \cref{sec:method:benchmark}, we introduce our three-stage evaluation protocol (\cref{fig:study-pipeline}a–c), which progressively tests (1) whether standard 3DGS with recovered inputs can compete with specially-designed methods, (2) how much extra gain these methods still provide, when given the same recovered inputs, and (3) why certain components remain helpful while others become harmful.

\vspace{-3pt}
\subsection{Recovering Degraded Inputs with Models}
\label{sec:method:assumptions}
\label{sec:method:separation}

Standard 3DGS~\cite{kerbl2023gaussians} reconstructs a 3D scene $\mathcal{G}^\star$ from a set of input images.
Specifically, a standard 3DGS training pipeline $V$ takes four inputs (\cref{fig:study-pipeline}, top):
(1) an initial point cloud $\mathcal{P}$ (typically from Structure-from-Motion~\cite{schoenberger2016sfm}),
(2) calibrated camera parameters $\hat{\Pi}$,
(3) training target images $\hat{\mathcal{I}}$, and
(4) valid pixel masks $\Omega$ indicating which pixels are used for loss computation (all pixels by default):
\begin{equation}
    \mathcal{G}^\star = V(\mathcal{P}, \hat{\Pi}, \hat{\mathcal{I}}, \Omega).
    \label{eq:interface}
\end{equation}
In degraded-input tasks, standard 3DGS struggles because one or more of these four inputs are missing or unreliable.
We use the same rendering loss, densification strategy, and 30k-step training schedule across tasks. We recover each task's inputs from the training observations before Gaussian optimization begins.

\begin{itemize}
    \item \textbf{Sparse Views:} When only a few training images are available (with known camera poses), traditional SfM fails or produces an overly sparse point cloud.
    We use the geometric foundation model VGGT-$\Omega$~\cite{wang2026vggtomega} to predict dense depth maps and confidence scores from the training views, and unproject high-confidence pixels using known camera poses to obtain a dense initial point cloud $\mathcal{P}$.

    \item \textbf{Unknown Camera Poses:} When camera extrinsics are unknown (with known intrinsics), both camera poses and the initial point cloud are missing.
    We use VGGT-$\Omega$~\cite{wang2026vggtomega} to predict initial camera poses and dense depth maps from the unposed images, refine the predicted extrinsics via fast bundle adjustment (BA, \cite{triggs1999bundle}) with fixed intrinsics, and supply the refined cameras $\hat{\Pi}$ and unprojected dense point cloud $\mathcal{P}$ to standard 3DGS.

    \item \textbf{Temporary Occlusions (Distractors):} When moving people or temporary objects occlude parts of the scene in training images, they corrupt both the initial SfM points and the image supervision.
    We use SAM~3.1~\cite{carion2026sam3} to segment candidate regions and identify temporary objects using DINOv2~\cite{oquab2023dinov2} features and multiview geometric consistency.
    We exclude the identified distractors from the supervision mask $\Omega$. For the end-to-end comparison, we construct $\mathcal{P}$ by back-projecting VGGT-$\Omega$ depth at retained pixels and filtering the resulting points using multiview evidence. The matched-input comparison uses a common training-only SfM cloud to isolate the effect of the masks.

    \item \textbf{Poor Exposure:} When training images are severely under- or over-exposed, the supervision targets $\hat{\mathcal{I}}$ become unreliable.
    We restore the degraded training images using DarkIR~\cite{feijoo2025darkir} for low-light scenes and LCDPNet~\cite{wang2022lcdpnet} for over-exposed scenes, and scale each restored image to a common mean luminance before clipping to produce the target images $\hat{\mathcal{I}}$. For the end-to-end comparison, the initial points are obtained from VGGT-$\Omega$ depth on the restored images.
\end{itemize}
All input recovery steps use only task-permitted training observations, and finish completely before 3DGS training begins.
\Cref{sec:supp:input} gives the recovery parameters, including the scale-relative near plane used for nine pose-free scenes.

\begin{table*}[t]
  \centering
  \begingroup
  \setlength{\tabcolsep}{2.25pt}
  \renewcommand{\arraystretch}{0.96}
  \iflatexml
    \begin{tabular}{@{}l*{4}{ccc}@{}}
\multicolumn{13}{l}{\bfseries (a) Sparse Views (91)} \\
\toprule
 & \multicolumn{3}{c}{LLFF (24)} & \multicolumn{3}{c}{DTU (45)} & \multicolumn{3}{c}{Mip360 (14)} & \multicolumn{3}{c}{Blender (8)} \\
\cmidrule(lr){2-4}\cmidrule(lr){5-7}\cmidrule(lr){8-10}\cmidrule(lr){11-13}
Method & PSNR$\uparrow$ & SSIM$\uparrow$ & LPIPS$\downarrow$ & PSNR$\uparrow$ & SSIM$\uparrow$ & LPIPS$\downarrow$ & PSNR$\uparrow$ & SSIM$\uparrow$ & LPIPS$\downarrow$ & PSNR$\uparrow$ & SSIM$\uparrow$ & LPIPS$\downarrow$ \\
\midrule
3DGS~\cite{kerbl2023gaussians} & 18.00 & 0.579 & 0.305 & 19.21$_{43}$ & 0.821 & 0.149 & 19.51 & 0.574 & \cellcolor{best2}0.345 & 22.84$_{6}$ & 0.866 & \cellcolor{best3}0.106 \\
DNGaussian~\cite{li2024dngaussian} & 20.95 & 0.671 & 0.272 & 13.61$_{3}$ & 0.684 & 0.280 & 14.59$_{3}$ & 0.364 & 0.669 & \cellcolor{best1}\textbf{24.10} & \cellcolor{best1}\textbf{0.884} & \cellcolor{best1}\textbf{0.090} \\
FSGS~\cite{zhu2024fsgs} & \cellcolor{best1}\textbf{21.58} & \cellcolor{best2}0.702 & \cellcolor{best2}0.210 & 21.96$_{43}$ & \cellcolor{best3}0.865 & 0.133 & \cellcolor{best3}20.90 & \cellcolor{best3}0.633 & \cellcolor{best3}0.353 & 18.88$_{1}$ & 0.837 & 0.147 \\
CoR-GS~\cite{zhang2024corgs} & \cellcolor{best3}21.11 & \cellcolor{best3}0.680 & \cellcolor{best3}0.218 & \cellcolor{best3}22.11$_{43}$ & \cellcolor{best2}0.881 & \cellcolor{best2}0.102 & \cellcolor{best2}20.91 & \cellcolor{best2}0.635 & 0.354 & \cellcolor{best2}23.00$_{6}$ & \cellcolor{best3}0.869 & 0.111 \\
DropGS~\cite{park2025dropgaussian} & \cellcolor{best2}21.57 & \cellcolor{best1}\textbf{0.704} & \cellcolor{best1}\textbf{0.202} & \cellcolor{best2}22.35$_{43}$ & \cellcolor{best2}0.881 & \cellcolor{best3}0.113 & \cellcolor{best1}\textbf{20.94} & \cellcolor{best1}\textbf{0.638} & 0.357 & 22.80$_{5}$ & 0.867 & 0.112 \\
Ours & 19.77 & 0.678 & 0.226 & \cellcolor{best1}\textbf{23.61} & \cellcolor{best1}\textbf{0.899} & \cellcolor{best1}\textbf{0.077} & 19.97 & 0.605 & \cellcolor{best1}\textbf{0.309} & \cellcolor{best3}22.98 & \cellcolor{best2}0.875 & \cellcolor{best2}0.101 \\
\bottomrule
\end{tabular}
\par
    \begin{tabular}{@{}l*{4}{ccc}@{}}
\multicolumn{13}{l}{\bfseries (b) Pose-Free (21)} \\
\toprule
 & \multicolumn{3}{c}{T\&T (8)} & \multicolumn{3}{c}{CO3D (5)} & \multicolumn{3}{c}{Mip360 (4)} & \multicolumn{3}{c}{DTU (4)} \\
\cmidrule(lr){2-4}\cmidrule(lr){5-7}\cmidrule(lr){8-10}\cmidrule(lr){11-13}
Method & PSNR$\uparrow$ & SSIM$\uparrow$ & LPIPS$\downarrow$ & PSNR$\uparrow$ & SSIM$\uparrow$ & LPIPS$\downarrow$ & PSNR$\uparrow$ & SSIM$\uparrow$ & LPIPS$\downarrow$ & PSNR$\uparrow$ & SSIM$\uparrow$ & LPIPS$\downarrow$ \\
\midrule
\textcolor{gray}{COLMAP}~\cite{schoenberger2016sfm} & \textcolor{gray}{28.62} & \textcolor{gray}{0.895} & \textcolor{gray}{0.101} & \textcolor{gray}{32.30} & \textcolor{gray}{0.932} & \textcolor{gray}{0.134} & \textcolor{gray}{28.32} & \textcolor{gray}{0.866} & \textcolor{gray}{0.172} & \textcolor{gray}{31.74} & \textcolor{gray}{0.943} & \textcolor{gray}{0.067} \\
CF-3DGS~\cite{fu2024colmapfree} & 16.42 & 0.442 & 0.332 & 14.04 & 0.460 & 0.553 & 14.48$_{0}$ & 0.271 & 0.469 & 16.83 & 0.628 & 0.315 \\
HT-3DGS~\cite{ji2025htgs} & 16.49 & 0.441 & 0.327 & 14.55 & 0.465 & 0.530 & 14.48$_{2}$ & 0.271 & 0.469 & 16.87 & 0.577 & 0.338 \\
MASt3R+GS~\cite{leroy2024mast3r} & \cellcolor{best3}20.12 & \cellcolor{best3}0.604 & \cellcolor{best3}0.214 & \cellcolor{best3}24.52 & \cellcolor{best3}0.734 & \cellcolor{best3}0.215 & \cellcolor{best3}18.35 & \cellcolor{best3}0.480 & \cellcolor{best3}0.355 & \cellcolor{best3}25.12 & \cellcolor{best3}0.848 & \cellcolor{best3}0.117 \\
3R-GS~\cite{huang2025threergs} & \cellcolor{best2}22.12 & \cellcolor{best2}0.663 & \cellcolor{best2}0.160 & \cellcolor{best2}25.92 & \cellcolor{best2}0.764 & \cellcolor{best2}0.192 & \cellcolor{best2}20.68 & \cellcolor{best2}0.605 & \cellcolor{best2}0.331 & \cellcolor{best2}25.89 & \cellcolor{best2}0.862 & \cellcolor{best2}0.111 \\
Ours & \cellcolor{best1}\textbf{27.89} & \cellcolor{best1}\textbf{0.890} & \cellcolor{best1}\textbf{0.090} & \cellcolor{best1}\textbf{30.25} & \cellcolor{best1}\textbf{0.900} & \cellcolor{best1}\textbf{0.140} & \cellcolor{best1}\textbf{27.98} & \cellcolor{best1}\textbf{0.859} & \cellcolor{best1}\textbf{0.166} & \cellcolor{best1}\textbf{32.14} & \cellcolor{best1}\textbf{0.945} & \cellcolor{best1}\textbf{0.058} \\
\bottomrule
\end{tabular}
\par
    \begin{tabular}{@{}l*{4}{ccc}@{}}
\multicolumn{13}{l}{\bfseries (c) Transient (19)} \\
\toprule
 & \multicolumn{3}{c}{RobustNeRF (5)} & \multicolumn{3}{c}{On-the-Go (6)} & \multicolumn{3}{c}{RealX3D (8)} & \multicolumn{3}{c}{All (19)} \\
\cmidrule(lr){2-4}\cmidrule(lr){5-7}\cmidrule(lr){8-10}\cmidrule(lr){11-13}
Method & PSNR$\uparrow$ & SSIM$\uparrow$ & LPIPS$\downarrow$ & PSNR$\uparrow$ & SSIM$\uparrow$ & LPIPS$\downarrow$ & PSNR$\uparrow$ & SSIM$\uparrow$ & LPIPS$\downarrow$ & PSNR$\uparrow$ & SSIM$\uparrow$ & LPIPS$\downarrow$ \\
\midrule
3DGS~\cite{kerbl2023gaussians} & 27.17 & 0.885 & 0.161 & 18.35 & 0.661 & 0.316 & 20.55 & 0.780 & 0.244 & 21.60 & 0.770 & 0.245 \\
SpotLess~\cite{sabour2025spotless} & 29.70 & 0.908 & 0.148 & 22.02 & 0.742 & 0.267 & 25.25 & 0.825 & 0.240 & 25.40 & 0.821 & 0.224 \\
WildGS~\cite{kulhanek2024wildgaussians} & 28.20 & 0.903 & 0.156 & \cellcolor{best3}22.43 & 0.771 & 0.221 & 23.72 & 0.824 & 0.225 & 24.49 & 0.828 & 0.205 \\
DeSplat~\cite{wang2025desplat} & \cellcolor{best2}31.03 & \cellcolor{best2}0.921 & \cellcolor{best2}0.106 & 22.32 & \cellcolor{best3}0.783 & \cellcolor{best3}0.186 & \cellcolor{best2}25.81 & \cellcolor{best1}\textbf{0.846} & \cellcolor{best2}0.173 & \cellcolor{best2}26.08 & \cellcolor{best2}0.846 & \cellcolor{best2}0.159 \\
RobustSplat~\cite{fu2025robustsplat} & \cellcolor{best1}\textbf{31.25} & \cellcolor{best1}\textbf{0.926} & \cellcolor{best3}0.117 & \cellcolor{best1}\textbf{23.01} & \cellcolor{best1}\textbf{0.801} & \cellcolor{best2}0.179 & \cellcolor{best1}\textbf{25.90} & \cellcolor{best2}0.843 & \cellcolor{best3}0.191 & \cellcolor{best1}\textbf{26.39} & \cellcolor{best1}\textbf{0.852} & \cellcolor{best3}0.168 \\
Ours & \cellcolor{best3}30.14 & \cellcolor{best3}0.917 & \cellcolor{best1}\textbf{0.097} & \cellcolor{best2}22.70 & \cellcolor{best2}0.787 & \cellcolor{best1}\textbf{0.171} & \cellcolor{best3}25.30 & \cellcolor{best3}0.836 & \cellcolor{best1}\textbf{0.153} & \cellcolor{best3}25.75 & \cellcolor{best3}0.842 & \cellcolor{best1}\textbf{0.144} \\
\bottomrule

\end{tabular}
\par
    \begin{tabular}{@{}l*{4}{ccc}@{}}
\multicolumn{13}{l}{\bfseries (d) Exposure (17)} \\
\toprule
 & \multicolumn{3}{c}{LOM Low (5)} & \multicolumn{3}{c}{RealX3D Low (9)} & \multicolumn{3}{c}{Low Light (14)} & \multicolumn{3}{c}{Over-Exp.\ (3)} \\
\cmidrule(lr){2-4}\cmidrule(lr){5-7}\cmidrule(lr){8-10}\cmidrule(lr){11-13}
Method & PSNR$\uparrow$ & SSIM$\uparrow$ & LPIPS$\downarrow$ & PSNR$\uparrow$ & SSIM$\uparrow$ & LPIPS$\downarrow$ & PSNR$\uparrow$ & SSIM$\uparrow$ & LPIPS$\downarrow$ & PSNR$\uparrow$ & SSIM$\uparrow$ & LPIPS$\downarrow$ \\
\midrule
3DGS~\cite{kerbl2023gaussians} & \phantom{0}6.88 & 0.155 & 0.541 & \phantom{0}6.70 & 0.061 & 0.676 & \phantom{0}6.77 & 0.095 & 0.628 & 10.36 & 0.733 & \cellcolor{best3}0.242 \\
Luminance-GS~\cite{cui2025luminance} & \cellcolor{best2}19.14 & \cellcolor{best2}0.826 & \cellcolor{best2}0.284 & \cellcolor{best3}12.98 & \cellcolor{best2}0.454 & \cellcolor{best2}0.498 & \cellcolor{best3}15.18 & \cellcolor{best2}0.587 & \cellcolor{best2}0.421 & \cellcolor{best2}20.85 & \cellcolor{best2}0.843 & \cellcolor{best2}0.158 \\
LITA-GS~\cite{zhou2025litags} & 17.85 & 0.767 & 0.366 & -- & -- & -- & -- & -- & -- & -- & -- & -- \\
LL-GS~\cite{sun2025llgaussian} & \cellcolor{best3}18.30 & \cellcolor{best3}0.798 & \cellcolor{best3}0.297 & \cellcolor{best2}13.88 & \cellcolor{best3}0.424 & \cellcolor{best3}0.519 & \cellcolor{best2}15.46 & \cellcolor{best3}0.558 & \cellcolor{best3}0.440 & -- & -- & -- \\
Aleth-NeRF~\cite{cui2024aleth} & -- & -- & -- & -- & -- & -- & -- & -- & -- & \cellcolor{best3}19.46 & \cellcolor{best3}0.788 & 0.311 \\
Ours & \cellcolor{best1}\textbf{23.78} & \cellcolor{best1}\textbf{0.871} & \cellcolor{best1}\textbf{0.225} & \cellcolor{best1}\textbf{17.33} & \cellcolor{best1}\textbf{0.671} & \cellcolor{best1}\textbf{0.369} & \cellcolor{best1}\textbf{19.63} & \cellcolor{best1}\textbf{0.743} & \cellcolor{best1}\textbf{0.318} & \cellcolor{best1}\textbf{23.97} & \cellcolor{best1}\textbf{0.888} & \cellcolor{best1}\textbf{0.121} \\
\bottomrule
\end{tabular}

  \else
  \resizebox{\textwidth}{!}{%
    \begin{tabular}{@{}c@{\hspace{8pt}}c@{}}
       &
       \\[12pt]
       &
      
    \end{tabular}%
  }
  \fi
  \endgroup
  \vspace{-7pt}
    \caption{\textbf{Stage 1 (End-to-End): Standard 3DGS with recovered inputs (Ours) is highly competitive with methods specially designed for each degraded-input task.}
    It leads all three reported metrics on average for sparse-view DTU and for poor exposure, and approaches the COLMAP-pose reference on pose-free scenes.
    Parentheses show the number of test cases. Subscripts show valid-run counts as defined in \cref{sec:supp:protocol}.
    \colorbox{best1}{Best}, \colorbox{best2}{Second}, and \colorbox{best3}{Third} mark the top three results per column (excluding the gray COLMAP-pose reference).
    \Cref{sec:supp:perscene} gives the per-scene results.}
\label{tab:main}
  \vspace{-10pt}
\end{table*}

\iflatexml\else
\newcommand{\irbDenseTile}[4]{%
  \begin{scope}
    \clip ({#1-1.325},{#2-0.85}) rectangle ({#1+1.325},{#2+0.85});
    \node[inner sep=0pt,anchor=center] at (#1,#2)
      {#3};
  \end{scope}
  \draw[black!35,line width=0.32pt]
    ({#1-1.325},{#2-0.85}) rectangle ({#1+1.325},{#2+0.85});
  \node[anchor=north west,inner xsep=3.2pt,inner ysep=2.1pt,
        rounded corners=0.8pt,fill=black,fill opacity=0.67,
        text=white,text opacity=1,
        font=\fontsize{7.0}{7.6}\selectfont\bfseries]
    at ({#1-1.285},{#2+0.810}) {#4};
}
\newcommand{\irbDenseTask}[6]{%
  \fill[black!78,rounded corners=1.2pt]
    (#1,{#2-0.85}) rectangle ({#1+0.44},{#2+0.85});
  \node[rotate=90,text=white,font=\fontsize{8.0}{8.5}\selectfont\bfseries]
    at ({#1+0.22},#2) {#3};
  \irbDenseTile{#1+1.88}{#2}{#4}{Specially-Designed}
  \irbDenseTile{#1+4.60}{#2}{#5}{Ours (Rec. + 3DGS)}
  \irbDenseTile{#1+7.32}{#2}{#6}{GT}
}
\fi
\begin{figure*}[t]
  \centering
\iflatexml
\includegraphics[width=\linewidth,alt={Specialized methods, standard 3DGS with improved inputs, and reference views for sparse views, unknown poses, occlusions, and poor exposure.}]{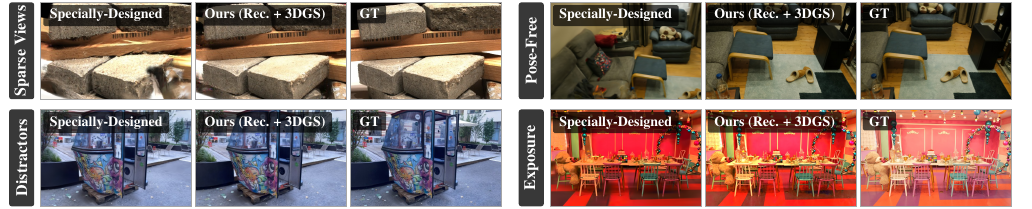}
\else
  \resizebox{0.985\linewidth}{!}{%
    \begin{tikzpicture}[x=1cm,y=1cm]
      \path[use as bounding box] (-0.07,-0.02) rectangle (17.77,3.71);
      \irbDenseTask{0.00}{2.77}{Sparse Views}
        {\includegraphics[width=2.65cm]{figures/qualitative/t1_sparse_dtu_scan38/specialist.jpg}}
        {\includegraphics[width=2.65cm]{figures/qualitative/t1_sparse_dtu_scan38/ours.jpg}}
        {\includegraphics[width=2.65cm]{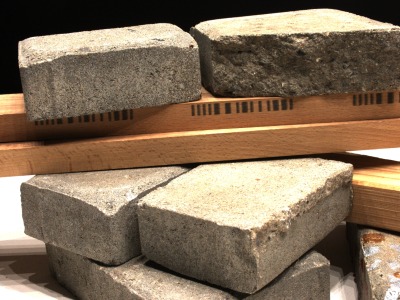}}
      \irbDenseTask{8.95}{2.77}{Pose-Free}
        {\includegraphics[width=2.65cm]{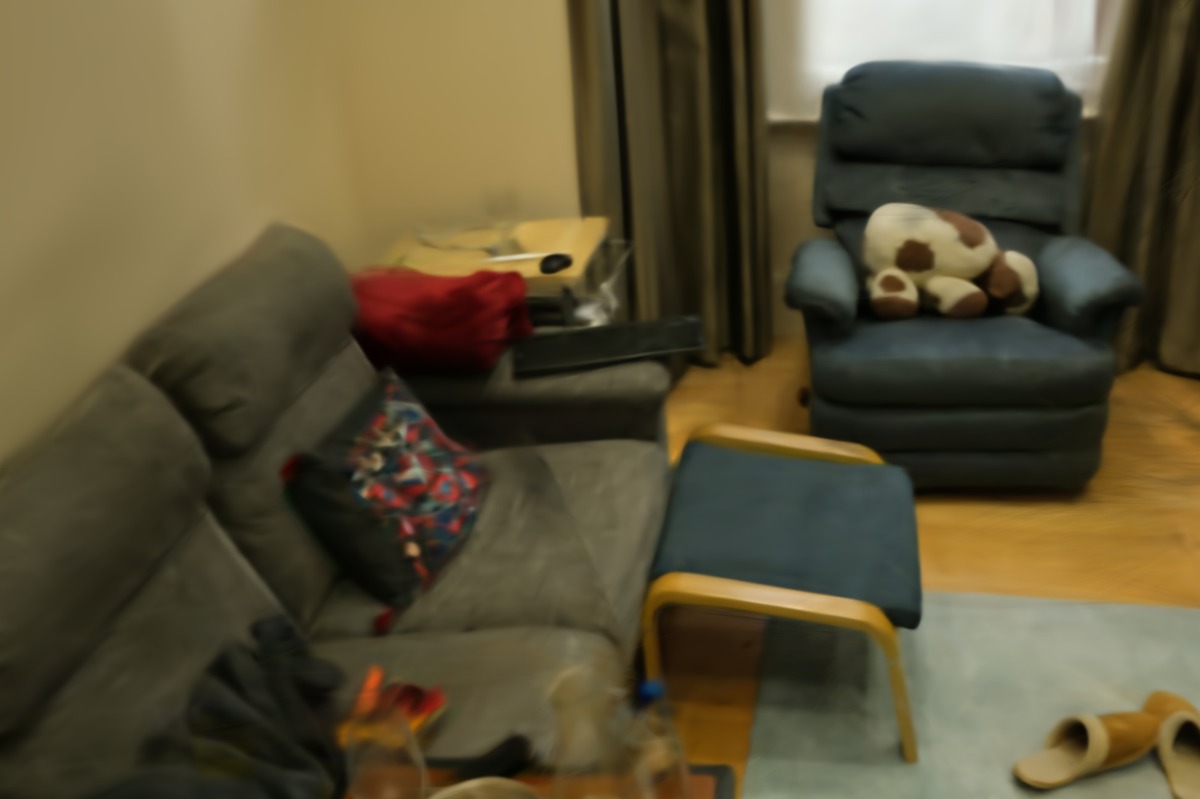}}
        {\includegraphics[width=2.65cm]{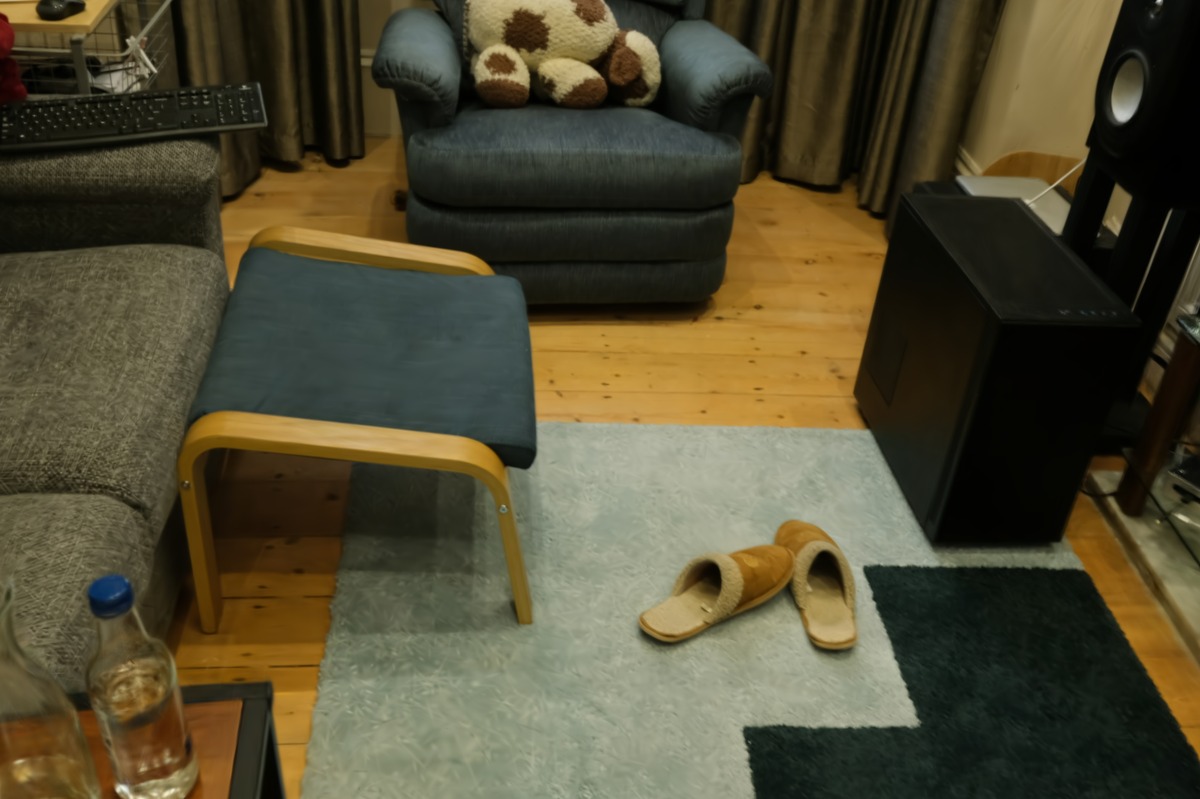}}
        {\includegraphics[width=2.65cm]{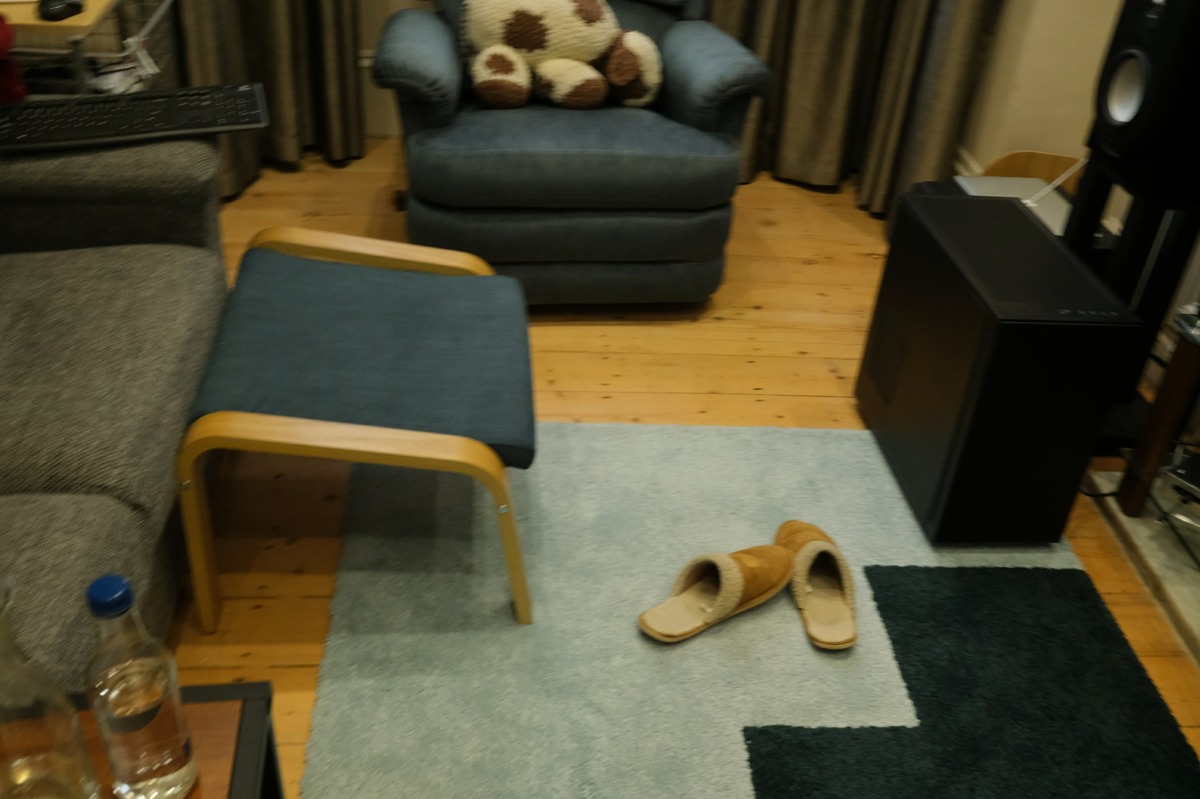}}
      \irbDenseTask{0.00}{0.89}{Distractors}
        {\includegraphics[width=2.65cm]{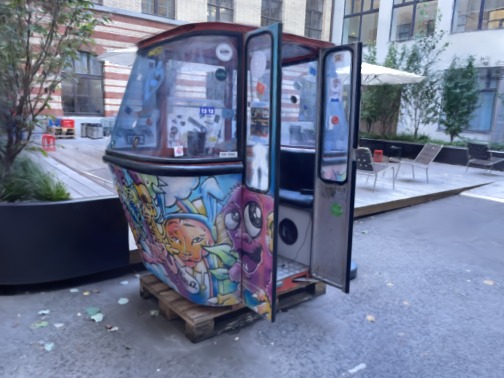}}
        {\includegraphics[width=2.65cm]{figures/qualitative/t3_transient_onthego_patio_high/ours.jpg}}
        {\includegraphics[width=2.65cm]{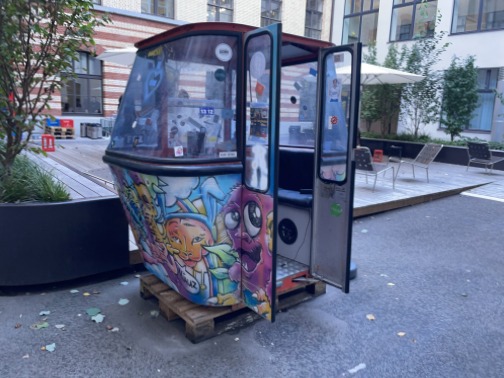}}
      \irbDenseTask{8.95}{0.89}{Exposure}
        {\includegraphics[width=2.65cm]{figures/qualitative/t4_lowlight_realx3d_cupcake/specialist.jpg}}
        {\includegraphics[width=2.65cm]{figures/qualitative/t4_lowlight_realx3d_cupcake/ours.jpg}}
        {\includegraphics[width=2.65cm]{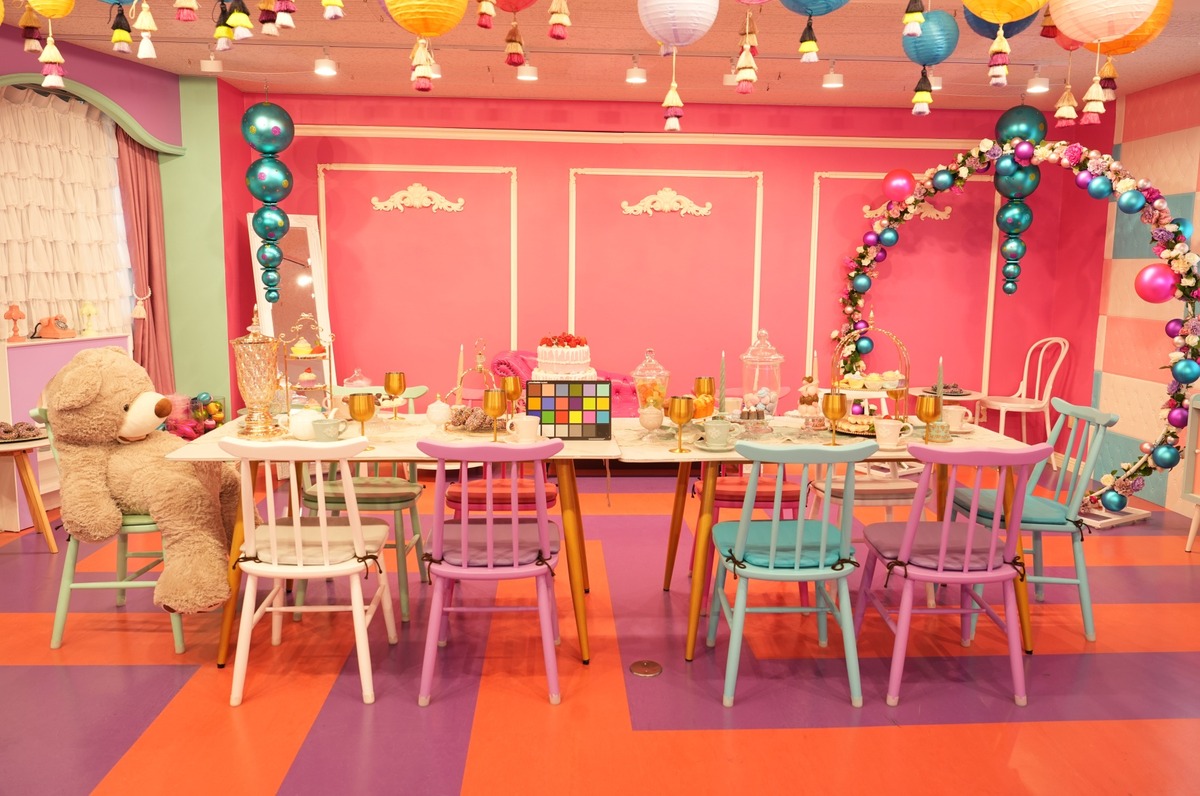}}
    \end{tikzpicture}%
  }
\fi
\vspace{-8pt}
  \caption{\textbf{With inputs recovered by foundation models, unmodified standard 3DGS (Ours) produces sharper geometry and cleaner appearance than specially-designed methods.}
    Each group compares a representative method trained on the original degraded inputs (CoR-GS, 3R-GS, RobustSplat, and LL-GS) against standard 3DGS trained on recovered inputs.
    \Cref{fig:supp:qualitative} shows the input or reference image and the standard 3DGS rendering for each example.}
    \label{fig:qualitative}
  \vspace{-17pt}
\end{figure*}

\subsection{Three-Stage Evaluation Protocol}
\label{sec:method:benchmark}

With our input recovery pipeline in place, both standard 3DGS ($V$) and specially-designed methods ($S$) can be trained on either the original degraded inputs (\textit{native}) or the foundation-model-improved inputs (\textit{recovered}).
Using this setup, we evaluate reconstructions on held-out test views across three progressive stages (\cref{fig:study-pipeline}a-c):

\vspace{-10pt}
\paragraph{Stage 1: End-to-End Comparison (\cref{fig:study-pipeline}a).}
We first test whether simply recovering the inputs before training makes standard 3DGS competitive with specially-designed methods.
In this stage, we run standard 3DGS on the recovered inputs ($V^{\text{recovered}}$) and compare its held-out rendering quality against specially-designed methods running on the original degraded inputs ($S^{\text{native}}$).

\vspace{-10pt}
\paragraph{Stage 2: Matched-Input Comparison (\cref{fig:study-pipeline}b).}
Since the methods in Stage 1 use different inputs, that comparison cannot tell us how much their extra training components still help.
We therefore compare $S$ and $V$ under the same inputs and measure the PSNR gain for each condition $c \in \{\text{native}, \text{recovered}\}$:
\begin{equation}
    \begin{gathered}
        g^{c} = \operatorname{PSNR}\bigl(S^{c}\bigr) - \operatorname{PSNR}\bigl(V^{c}\bigr), \\
        \Delta g = g^{\text{recovered}} - g^{\text{native}}.
    \end{gathered}
    \label{eq:specialized_gain}
\end{equation}
Here, $g^{c} > 0$ means $S$ outperforms standard 3DGS under input condition $c$, while $g^{c} < 0$ means it performs worse.
A negative $\Delta g$ means that the advantage of $S$ shrinks after input recovery, even if its absolute PSNR improves.
For each task, both methods receive the same designated recovered input: dense points $\mathcal{P}$ for sparse views, refined cameras $\hat{\Pi}$ and points $\mathcal{P}$ for unknown poses, static masks $\Omega$ for temporary occlusions, or restored images $\hat{\mathcal{I}}$ for poor exposure.

\vspace{-10pt}
\paragraph{Stage 3: Controlled Component Probes (\cref{fig:study-pipeline}c).}
Further, as we will show in \cref{sec:experiments}, when given recovered inputs, some specially-designed methods outperform standard 3DGS ($g^{\text{recovered}} > 0$), but others actually perform worse ($g^{\text{recovered}} < 0$).
However, because existing method codebases often bundle multiple components and hyperparameter changes together, pipeline-level comparisons alone cannot isolate why a specific design helps or hurts.
Therefore, in our third stage (\cref{sec:analysis}), we implement representative training components, such as depth priors, camera pose optimization, robust occlusion masking, and learnable exposure adjustment, individually inside the same shared 3DGS trainer $V$.
By varying one component and one input-quality factor at a time, we examine why some components damage already accurate inputs while others continue to help.

\Cref{sec:supp:protocol,sec:supp:matched,sec:supp:analysis} give the evaluation protocol, matched-input comparisons, and component-probe settings, respectively.

\section{Experiments}
\label{sec:experiments}

\subsection{Experimental Setup}
\label{sec:exp:setup}

\paragraph{Datasets and baselines.}
We evaluate 148 test cases (configurations) across our four degraded-input tasks on 10 datasets (\cref{tab:main}).
Specifically, we use LLFF~\cite{mildenhall2019llff}, DTU~\cite{jensen2014dtu}, Mip-NeRF 360~\cite{barron2022mipnerf360}, and Blender~\cite{mildenhall2020nerf} for sparse views;
Tanks and Temples~\cite{knapitsch2017tanks}, CO3D~\cite{reizenstein2021co3d}, Mip-NeRF 360, and DTU for unknown camera poses (pose-free);
RobustNeRF~\cite{sabour2023robustnerf}, On-the-go~\cite{ren2024onthego}, and RealX3D~\cite{realx3d2025} for temporary occlusions (transient distractors);
and LOM~\cite{cui2024aleth} and RealX3D for poor exposure.

\paragraph{Evaluation details.}
We evaluate all methods on held-out test views using PSNR, SSIM, and LPIPS.
To prevent test-view leakage, initial SfM points are built from training views only, and pose-free test cameras are aligned via a similarity transform fitted solely on training cameras.
We compare our unmodified 30k-step 3DGS pipeline against official implementations of specially-designed methods. If a baseline fails completely on a test case, we assign it the worst score among compared methods on that case.

\Cref{sec:supp:protocol} details the splits, metrics, adapters, and handling of failed runs.

\vspace{-2pt}
\subsection{Reconstruction with Recovered Inputs}
\label{sec:exp:main}
\vspace{-1pt}

We begin with Stage~1 (\cref{fig:study-pipeline}a), comparing standard 3DGS on recovered inputs against specially-designed methods running on the original degraded inputs.
As shown in \cref{tab:main} and \cref{fig:qualitative}, this simple approach completes all 148 test cases without failure and achieves results highly competitive with methods designed for each task.
Even when compared against the best-performing specially-designed method on each individual test case, standard 3DGS with recovered inputs still achieves higher PSNR on 43\% (63/146) of comparable cases and lower LPIPS on 69\% (101/146). The two excluded cases had no valid specialized baseline (\cref{sec:supp:perscene}).

Looking closer across tasks (\cref{tab:main}), we find that when foundation models can cleanly resolve the input defects before training, standard 3DGS alone outperforms specially-designed methods.
For unknown camera poses, refining VGGT-$\Omega$ predictions with fixed-intrinsics BA reduces the mean camera rotation error from $0.567^\circ$ to $0.068^\circ$.
With these accurate cameras, standard 3DGS reaches 29.28\,dB across all pose-free scenes, beating the strongest pose-free baseline 3R-GS by $+5.81$\,dB and coming within $0.75$\,dB of the COLMAP-pose reference.
For poor exposure, standard 3DGS with restored inputs yields the best average PSNR, SSIM, and LPIPS on both low-light and over-exposed scenes. Even scalar brightness alignment alone improves on LL-Gaussian by $1.43$\,dB across the 14 low-light scenes (\cref{sec:supp:perscene}).

Meanwhile, sparse views and temporary occlusions remain more challenging, as unobserved regions and occluded backgrounds still leave 3D ambiguities during Gaussian fitting.
Even so, input recovery closes most of the gap (\cref{tab:main}).
In sparse views, dense initialization prevents sparse-point failures and achieves the best results on DTU, though it trails FSGS on forward-facing LLFF.
For temporary occlusions, using recovered static masks and a filtered VGGT-$\Omega$ point cloud boosts standard 3DGS by $+4.15$\,dB and yields the best overall LPIPS (0.144), trailing the top baseline RobustSplat by only $0.64$\,dB in PSNR.

\begin{figure*}[t]
\centering
\iflatexml
\includegraphics[width=\linewidth,alt={Pose refinement and exposure target updates lose their PSNR advantage as input errors shrink. The pose crossover varies with the learning rate.}]{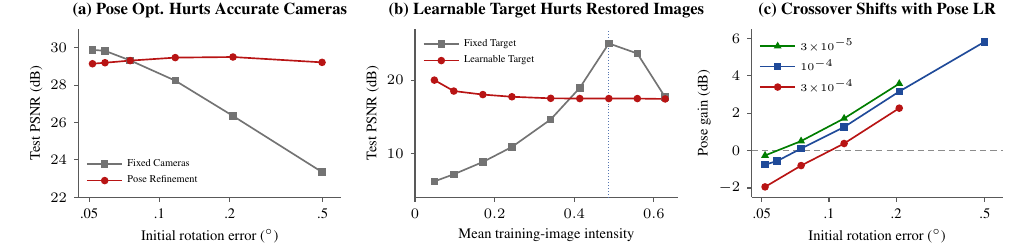}
\else
\begin{tikzpicture}
\begin{axis}[
  irb, name=pose,scale only axis,width=4.45cm,height=2.85cm,
  xmode=log,xmin=0.045,xmax=0.60,
  xtick={0.05,0.1,0.2,0.5},xticklabels={.05,.1,.2,.5},
  ymin=22,ymax=31,
  xlabel={Initial rotation error ($^\circ$)},ylabel={Test PSNR (dB)},
  title={(a) Pose Opt. Hurts Accurate Cameras},
  legend style={at={(0.02,0.03)},anchor=south west,font=\tiny}
]
  \addplot[irbGray,irb line,mark=square*]
    coordinates {(0.0517,29.889)(0.0585,29.834)(0.0750,29.322)
      (0.1169,28.241)(0.2065,26.354)(0.4968,23.353)};
  \addlegendentry{Fixed Cameras}
  \addplot[irbRed,irb line,mark=*]
    coordinates {(0.0517,29.145)(0.0585,29.200)(0.0750,29.313)
      (0.1169,29.473)(0.2065,29.505)(0.4968,29.219)};
  \addlegendentry{Pose Refinement}
\end{axis}
\begin{axis}[
  irb,name=exposure,at={(pose.east)},anchor=west,xshift=1.25cm,
  scale only axis,width=4.45cm,height=2.85cm,
  xmin=0,xmax=0.66,ymin=4,ymax=27,
  xlabel={Mean training-image intensity},ylabel={Test PSNR (dB)},
  title={(b) Learnable Target Hurts Restored Images},
  legend style={at={(0.02,0.98)},anchor=north west,font=\tiny}
]
  \addplot[irbGray,irb line,mark=square*]
    coordinates {(0.0487,6.185)(0.0974,7.154)(0.1704,8.835)
      (0.2434,10.888)(0.3408,14.650)(0.4138,18.965)
      (0.4868,25.037)(0.5584,23.659)(0.6283,17.686)};
  \addlegendentry{Fixed Target}
  \addplot[irbRed,irb line,mark=*]
    coordinates {(0.0487,20.022)(0.0974,18.527)(0.1704,18.046)
      (0.2434,17.747)(0.3408,17.536)(0.4138,17.504)
      (0.4868,17.504)(0.5584,17.502)(0.6283,17.447)};
  \addlegendentry{Learnable Target}
  \draw[irbBlue!75,densely dotted,thin]
    (axis cs:0.4868,4)--(axis cs:0.4868,27);
\end{axis}
\begin{axis}[
  irb,at={(exposure.east)},anchor=west,xshift=1.25cm,
  scale only axis,width=4.25cm,height=2.85cm,
  xmode=log,xmin=0.045,xmax=0.60,
  xtick={0.05,0.1,0.2,0.5},xticklabels={.05,.1,.2,.5},
  ymin=-2.5,ymax=6.5,ytick={-2,0,2,4,6},
  xlabel={Initial rotation error ($^\circ$)},ylabel={Pose gain (dB)},
  title={(c) Crossover Shifts with Pose LR},
  legend style={at={(0.02,0.98)},anchor=north west,font=\tiny}
]
  \draw[black!45,densely dashed,thin]
    (axis cs:0.045,0)--(axis cs:0.60,0);
  \addplot[irbGreen,irb line,mark=triangle*]
    coordinates {(0.0517,-0.26)(0.0750,0.51)(0.1169,1.72)(0.2065,3.58)};
  \addlegendentry{$3\!\times\!10^{-5}$}
  \addplot[irbBlue,irb line,mark=square*]
    coordinates {(0.0517,-0.75)(0.0585,-0.56)(0.0750,0.11)
      (0.1169,1.26)(0.2065,3.16)(0.4968,5.81)};
  \addlegendentry{$10^{-4}$}
  \addplot[irbRed,irb line,mark=*]
    coordinates {(0.0517,-1.94)(0.0750,-0.80)(0.1169,0.38)(0.2065,2.27)};
  \addlegendentry{$3\!\times\!10^{-4}$}
\end{axis}
\end{tikzpicture}
\fi
\vspace{-4pt}
\caption{\textbf{Extra components that modify cameras or target images switch from helpful to harmful as input errors decrease.}
  Because unconverged Gaussians produce large rendering errors early in training, joint optimization can mistakenly alter already accurate inputs:
  (a) pose refinement improves coarse cameras but pushes accurate BA cameras away from their true poses;
  (b) learnable exposure targets help dark images but degrade restored ones; and
  (c) larger pose learning rates shift the zero-gain crossover to larger initial rotation errors.}
\label{fig:fig5_crossover}

\vspace{-15pt}
\end{figure*}

\vspace{-2pt}
\subsection{Gains from Extra Components}
\label{sec:exp:shared}
\vspace{-1pt}

Since standard 3DGS with recovered inputs has already matched or beaten many specially-designed methods, we next move to Stage~2 (\cref{fig:study-pipeline}b): \emph{How much do the extra training components help when both methods receive the same recovered information for a given task?}
As shown in \cref{tab:shared}, comparing each method against standard 3DGS with the designated inputs shared (\cref{eq:specialized_gain}) reveals that the PSNR advantage of specially-designed methods shrinks ($\Delta g < 0$) in 338 out of 394 matched comparisons (86\%).
More strikingly, the four tasks split into two clear regimes:

First, for unknown camera poses and poor exposure, \textbf{the average extra gain turns negative} ($g^{\text{recovered}} < 0$), meaning that specially-designed methods perform even worse than standard 3DGS on the same improved inputs.
Notably, better inputs do improve the absolute performance of these baselines—for example, giving recovered cameras and points to 3R-GS raises its PSNR from 23.47 to 28.39\,dB.
However, standard 3DGS improves even more on the same inputs (reaching 29.28\,dB), flipping 3R-GS's relative gain from $+1.69$ to $-0.89$\,dB.
Likewise, on restored images, the average relative gain of exposure-specific methods drops from $+9.11$ to $-1.38$\,dB across 36 comparisons.

Second, for sparse views and temporary occlusions, specially-designed methods retain a positive average gain ($g^{\text{recovered}} > 0$), but this advantage shrinks substantially (\cref{tab:shared}).
In sparse views, dense VGGT-$\Omega$ points resolve 19 baseline initialization failures. On the 261 matched pairs, sharing the dense points cuts the mean PSNR advantage from $+2.57$ to $+0.78$\,dB (\cref{sec:supp:matched}).
For temporary occlusions, providing the same recovered static masks to both sides (while keeping the original SfM points fixed) reduces the mean advantage of robust methods from $+3.99$ to $+0.61$\,dB.
These results raise two key questions: \emph{Why do some extra components hurt reconstruction once the inputs are accurate, and where do the remaining positive gains come from?} We investigate both in \cref{sec:analysis}.

\begin{table}[t]
  \centering
  \footnotesize
  \begingroup
  \setlength{\tabcolsep}{1.65pt}
  \renewcommand{\arraystretch}{1.04}
  \iflatexml
    \begin{tabular}{@{}l@{\hspace{6pt}}l@{\hspace{5pt}}r@{\hspace{6pt}}r@{\hspace{6pt}}r@{\hspace{6pt}}r@{}}
    \toprule
    & & & \multicolumn{2}{c}{Task-Specific PSNR Gain (dB)} & \\
    \cmidrule(lr){4-5}
    Setting & Shared Input & Pairs & Native & Recovered & Shrinks \\
    \midrule
    Sparse Views & Points & 261 & +2.57 & +0.78 & 214 \\
    Pose-Free & Cams., Points & 21 & +1.69 & $-$0.89 & 18 \\
    Transient & Masks & 76 & +3.99 & +0.61 & 70 \\
    Exposure & Images & 36 & +9.11 & $-$1.38 & 36 \\
    \midrule
    \multicolumn{2}{@{}l}{Total Matched Pairs} & 394 & \multicolumn{2}{c}{--} & 338 \\
    \bottomrule
  \end{tabular}%

  \else
    \begin{adjustbox}{max width=\columnwidth}
      \begin{tabular}{@{}l@{\hspace{6pt}}l@{\hspace{5pt}}r@{\hspace{6pt}}r@{\hspace{6pt}}r@{\hspace{6pt}}r@{}}
    \toprule
    & & & \multicolumn{2}{c}{Task-Specific PSNR Gain (dB)} & \\
    \cmidrule(lr){4-5}
    Setting & Shared Input & Pairs & Native & Recovered & Shrinks \\
    \midrule
    Sparse Views & Points & 261 & +2.57 & +0.78 & 214 \\
    Pose-Free & Cams., Points & 21 & +1.69 & $-$0.89 & 18 \\
    Transient & Masks & 76 & +3.99 & +0.61 & 70 \\
    Exposure & Images & 36 & +9.11 & $-$1.38 & 36 \\
    \midrule
    \multicolumn{2}{@{}l}{Total Matched Pairs} & 394 & \multicolumn{2}{c}{--} & 338 \\
    \bottomrule
  \end{tabular}%
    \end{adjustbox}
  \fi
  \endgroup
  \vspace{-3pt}
  \caption{\textbf{Stage 2 (Matched Inputs): When both methods share the recovered input listed for each task, the PSNR gain over standard 3DGS shrinks in 338 of 394 comparisons (86\%).}
    Notably, average relative gains even turn negative ($g^{\text{recovered}} < 0$) for pose-free reconstruction and poor exposure, meaning standard 3DGS performs better on the same improved inputs.
    Here, positive values indicate an advantage over standard 3DGS under the same inputs (\cref{eq:specialized_gain}), and \emph{Shrinks} counts comparisons where $\Delta g < 0$. \Cref{fig:supp:stage2} shows all 394 pairs.}
\label{tab:shared}
  \vspace{-20pt}
\end{table}

\vspace{-7pt}

\section{Analysis of Extra Components}
\label{sec:analysis}

\vspace{-3pt}

\begin{figure*}[t]
\centering
\iflatexml
\includegraphics[width=\linewidth,alt={Curves and image comparisons show how early rendering errors reduce the contrast of restored supervision targets.}]{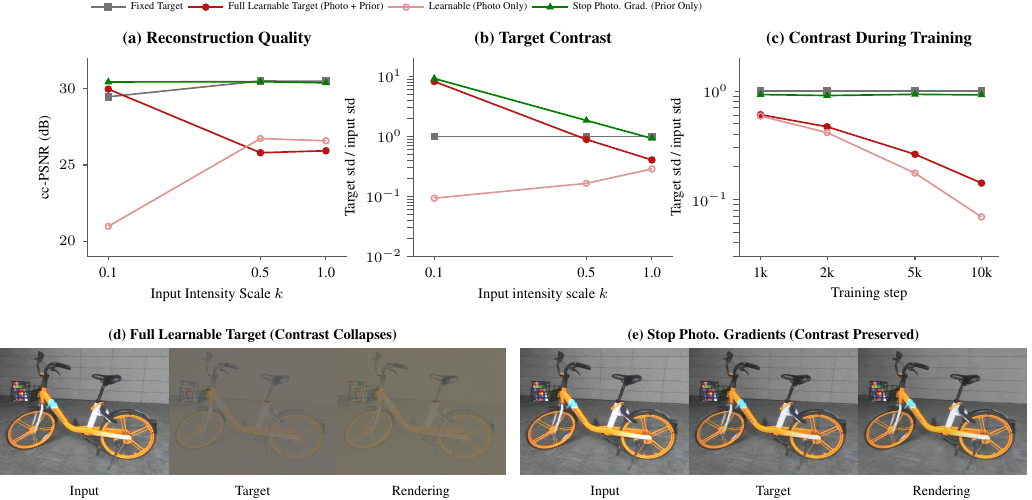}
\else
\begin{tikzpicture}
\begin{axis}[irb, name=a, scale only axis, width=4.4cm, height=3.35cm, xmode=log, xmin=0.08, xmax=1.25, xtick={0.1,0.5,1}, xticklabels={0.1,0.5,1.0}, ymin=19, ymax=32, xlabel={Input Intensity Scale $k$}, ylabel={cc-PSNR (dB)}, title={(a) Reconstruction Quality}, legend style={at={(0.0,1.2)}, anchor=south west, legend columns=4, font=\tiny}]
  \addplot[irbGray, irb line, mark=square*] coordinates {(0.1,29.481) (0.5,30.516) (1.0,30.505)};
  \addlegendentry{Fixed Target}
  \addplot[irbRed, irb line, mark=*] coordinates {(0.1,29.986) (0.5,25.806) (1.0,25.933)};
  \addlegendentry{Full Learnable Target (Photo + Prior)}
  \addplot[irbRed!45, irb line, mark=o] coordinates {(0.1,20.965) (0.5,26.736) (1.0,26.589)};
  \addlegendentry{Learnable (Photo Only)}
  \addplot[irbGreen, irb line, mark=triangle*] coordinates {(0.1,30.455) (0.5,30.474) (1.0,30.399)};
  \addlegendentry{Stop Photo. Grad. (Prior Only)}
\end{axis}
\begin{axis}[irb, name=b, at={(a.east)}, anchor=west, xshift=1.12cm, scale only axis, width=4.4cm, height=3.35cm, xmode=log, ymode=log, xmin=0.08, xmax=1.25, xtick={0.1,0.5,1}, xticklabels={0.1,0.5,1.0}, ymin=0.01, ymax=20, xlabel={Input intensity scale $k$}, ylabel={Target std / input std}, title={(b) Target Contrast}]
  \addplot[irbGray, irb line, mark=square*, forget plot] coordinates {(0.1,1.0000) (0.5,1.0000) (1.0,1.0000)};
  \addplot[irbRed, irb line, mark=*, forget plot] coordinates {(0.1,8.2475) (0.5,0.8898) (1.0,0.4049)};
  \addplot[irbRed!45, irb line, mark=o, forget plot] coordinates {(0.1,0.0932) (0.5,0.1649) (1.0,0.2870)};
  \addplot[irbGreen, irb line, mark=triangle*, forget plot] coordinates {(0.1,9.2456) (0.5,1.8488) (1.0,0.9258)};
\end{axis}
\begin{axis}[irb, name=c, at={(b.east)}, anchor=west, xshift=1.12cm, scale only axis, width=4.4cm, height=3.35cm, xmode=log, ymode=log, xmin=800, xmax=12000, xtick={1000,2000,5000,10000}, xticklabels={1k,2k,5k,10k}, ymin=0.03, ymax=2, xlabel={Training step}, ylabel={Target std / input std}, title={(c) Contrast During Training}]
  \addplot[irbGray, irb line, mark=square*, forget plot] coordinates {(1000,1.0000) (2000,1.0000) (5000,1.0000) (10000,1.0000)};
  \addplot[irbRed, irb line, mark=*, forget plot] coordinates {(1000,0.6062) (2000,0.4696) (5000,0.2616) (10000,0.1419)};
  \addplot[irbRed!45, irb line, mark=o, forget plot] coordinates {(1000,0.5886) (2000,0.4144) (5000,0.1754) (10000,0.0692)};
  \addplot[irbGreen, irb line, mark=triangle*, forget plot] coordinates {(1000,0.9300) (2000,0.9102) (5000,0.9334) (10000,0.9223)};
\end{axis}
\end{tikzpicture}

\vspace{3pt}
\noindent
\begin{minipage}[t]{0.49\textwidth}
  \centering
  {\scriptsize\bfseries (d) Full Learnable Target (Contrast Collapses)}\par\vspace{2pt}
  \includegraphics[width=\linewidth]{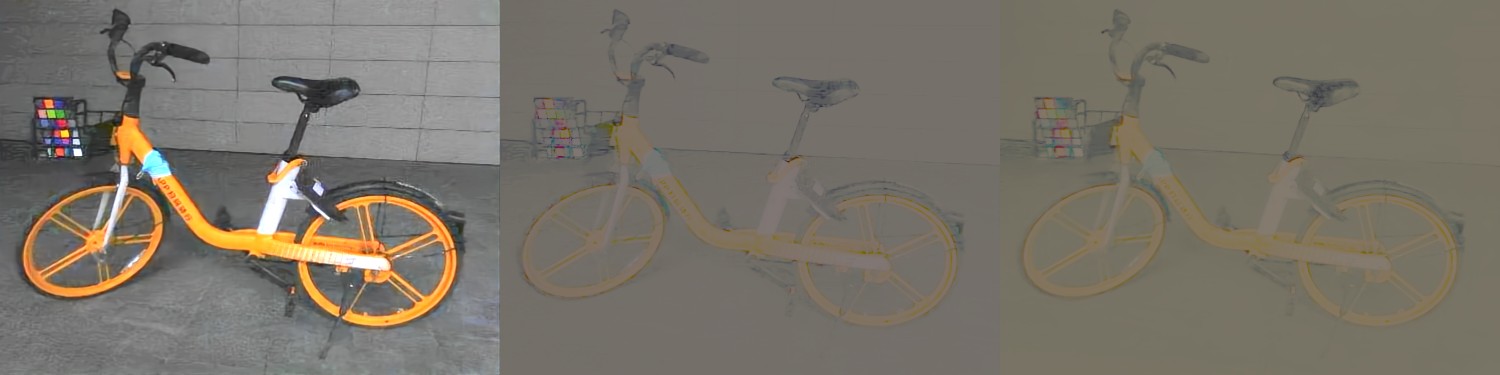}\par
  \vspace{-2pt}
  {\fontsize{6.7}{7.5}\selectfont
   \makebox[0.333\linewidth][c]{Input}%
   \makebox[0.333\linewidth][c]{Target}%
   \makebox[0.333\linewidth][c]{Rendering}}
\end{minipage}\hfill
\begin{minipage}[t]{0.49\textwidth}
  \centering
  {\scriptsize\bfseries (e) Stop Photo. Gradients (Contrast Preserved)}\par\vspace{2pt}
  \includegraphics[width=\linewidth]{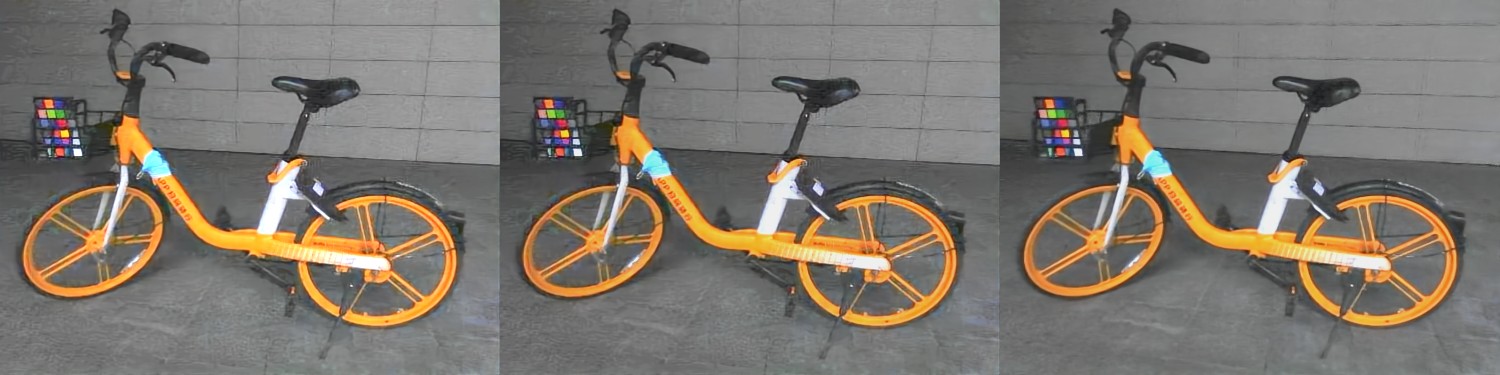}\par
  \vspace{-2pt}
  {\fontsize{6.7}{7.5}\selectfont
   \makebox[0.333\linewidth][c]{Input}%
   \makebox[0.333\linewidth][c]{Target}%
   \makebox[0.333\linewidth][c]{Rendering}}
\end{minipage}
\fi
\vspace{-5pt}
\caption{\textbf{Allowing photometric gradients to update learnable target images collapses image contrast toward blurry early renderings.}
  On restored LOM images scaled by brightness $k$:
  (a) the full learnable target lowers mean color-calibrated PSNR (cc-PSNR) at $k=0.5$ and $k=1.0$.
  (b) Target contrast varies with exposure. (c--d) On restored \texttt{bike} images, contrast declines during training and the target loses visible detail.
  (e) Simply stopping photometric gradients to the target (updating it only via the brightness prior) prevents contrast collapse and restores reconstruction quality across brightness levels.}
\label{fig:t4_moving_target}
\vspace{-17pt}
\end{figure*}

To understand the split observed in \cref{tab:shared}, we move to Stage~3 (\cref{fig:study-pipeline}c) and test individual training components inside our shared 3DGS trainer.
The probes show how \textbf{3DGS training errors can limit reconstruction once the inputs improve}. We examine when extra updates disturb accurate inputs (\cref{sec:analysis_degrade}) and which fitting artifacts account for some of the remaining gains (\cref{sec:analysis_survive}).
The probe settings and supporting results appear in \cref{sec:supp:analysis}.

\subsection{Why Do Some Components Become Harmful?}
\label{sec:analysis_degrade}

We first vary the input quality while keeping the training component and scenes fixed, which cleanly reproduces the reversal from \cref{tab:shared}.
As shown in \cref{fig:fig5_crossover}a, joint camera pose optimization improves test PSNR by $+5.87$\,dB when initial cameras have coarse errors ($0.497^\circ$), but reduces PSNR by $0.74$\,dB when the BA cameras are already accurate ($0.052^\circ$).
Similarly, adding a Luminance-GS-style learnable image target~\cite{cui2025luminance} boosts PSNR by $+13.84$\,dB on severely dark images (intensity scale $k=0.1$), yet loses $7.53$\,dB once exposure is recovered ($k=1.0$, \cref{fig:fig5_crossover}b).
In the official pipelines, 3R-GS~\cite{huang2025threergs} increases BA rotation error in 14 of 21 scenes. In converged Luminance-GS runs on restored LOM \texttt{chair} and \texttt{sofa}, enabling the learnable target reduces PSNR by $2.2$--$3.6$\,dB (\cref{sec:supp:pose,sec:supp:target}).

Why does an optimization that helps poor inputs hurt accurate ones?
Early in 3DGS training, unconverged Gaussians produce blurry renderings with large photometric residuals even with accurate cameras or restored images.
Jointly optimizing camera poses or target images lets the optimizer reduce those residuals by altering the supplied inputs.

\vspace{-10pt}
\paragraph{Pose updates overshoot before the scene adapts.}
In camera optimization, one might suspect that early pose gradients from unconverged Gaussians point in wrong directions.
Surprisingly, during the first 100 steps on accurate BA cameras, the negative pose gradient on the sampled view still correlates positively with the true correction direction (mean cosine $+0.28$), and about six in ten sampled-camera updates reduce its error.
Instead, the problem lies in the \emph{step size} of the executed gradient-descent updates.
For a local rotation error $e$ and an update step $u$ making an angle $\phi$ with the ideal correction $-e$, the squared pose error changes by:
\begin{equation}
  \|e+u\|^2-\|e\|^2
  \approx \|u\|^2-2\|e\|\|u\|\cos\phi.
  \label{eq:pose_overshoot}
\end{equation}
Thus, even a correctly directed update ($\cos\phi > 0$) increases camera error whenever its step size $\|u\|$ exceeds $2\|e\|\cos\phi$.
When initial cameras are accurate, $\|e\|$ is small. Accumulated Adam~\cite{kingma2014adam} momentum also moves unsampled cameras. During the first 100 steps, $81\%$ of these moves oppose the correction direction. By step 500, mean BA rotation error rises from $0.052^\circ$ to $0.081^\circ$ (\cref{sec:supp:pose}).

Consequently, raising the pose learning rate shifts the zero-gain crossover toward larger initial errors (\cref{fig:fig5_crossover}c).
Waiting for the Gaussians to converge before updating cameras is not a simple fix either: after roughly 3k steps, the Gaussians have already adapted to the initial cameras, so pose gradients lose alignment with the true correction direction (cosine near zero) and the gain on coarse VGGT-$\Omega$ cameras drops from $+2.38$ to $+0.66$\,dB.

\begin{figure*}[t]
\centering
\iflatexml
\includegraphics[width=\linewidth,alt={Quantitative and visual comparisons of depth supervision, Gaussian dropout, and densification after input recovery.}]{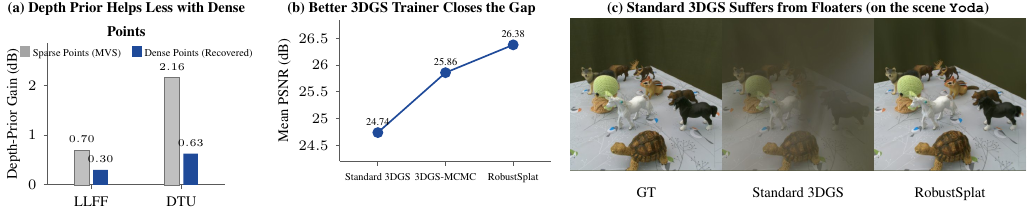}
\else
\begin{minipage}[t]{0.245\textwidth}
  \centering\vspace{0pt}
  {\scriptsize\bfseries (a) Depth Prior Helps Less with Dense Points}\par\vspace{1pt}
  \begin{tikzpicture}
    \begin{axis}[
      irb,scale only axis,width=0.70\linewidth,height=2.4cm,
      ybar,bar width=7pt,ymin=0,ymax=2.85,
      symbolic x coords={LLFF,DTU},xtick=data,
      enlarge x limits=0.48,ylabel={Depth-Prior Gain (dB)},
      nodes near coords,
      every node near coord/.append style={font=\tiny,
        /pgf/number format/fixed,/pgf/number format/precision=2,
        /pgf/number format/fixed zerofill}
    ]
      \addplot[fill=irbGray!45,draw=irbGray,line width=0.3pt]
        coordinates {(LLFF,0.70)(DTU,2.16)};
      \addplot[fill=irbBlue,draw=none]
        coordinates {(LLFF,0.30)(DTU,0.63)};
      \node[anchor=north,font=\tiny,inner sep=0pt]
        at (rel axis cs:0.5,0.98) {%
          \textcolor{irbGray!65}{\rule{5pt}{5pt}}\ Sparse Points (MVS)\quad
          \textcolor{irbBlue}{\rule{5pt}{5pt}}\ Dense Points (Recovered)};
    \end{axis}
  \end{tikzpicture}
\end{minipage}\hfill
\begin{minipage}[t]{0.273\textwidth}
  \centering\vspace{0pt}
  {\scriptsize\bfseries (b) Better 3DGS Trainer Closes the Gap}\par\vspace{1pt}
  \begin{tikzpicture}
    \begin{axis}[
      irb,scale only axis,width=0.75\linewidth,height=2.4cm,
      ymin=24.2,ymax=26.85,
      symbolic x coords={Standard 3DGS,3DGS-MCMC,RobustSplat},xtick=data,
      x tick label style={font=\tiny,rotate=0},
      enlarge x limits=0.28,
      ylabel={Mean PSNR (dB)},
      ytick={24.5,25,25.5,26,26.5},
    ]
      \addplot[irbBlue,irb line,mark=*,mark size=2.2pt]
        coordinates {(Standard 3DGS,24.74)(3DGS-MCMC,25.86)(RobustSplat,26.38)};
      \node[anchor=south,font=\tiny] at (axis cs:Standard 3DGS,24.74) {24.74};
      \node[anchor=south,font=\tiny] at (axis cs:3DGS-MCMC,25.86) {25.86};
      \node[anchor=south,font=\tiny] at (axis cs:RobustSplat,26.38) {26.38};
    \end{axis}
  \end{tikzpicture}
\end{minipage}\hfill
\begin{minipage}[t]{0.442\textwidth}
  \centering\vspace{0pt}
  {\scriptsize\bfseries (c) Standard 3DGS Suffers from Floaters (on the scene \texttt{Yoda})}\par\vspace{1pt}
  \includegraphics[width=\linewidth,
    trim={0bp 864bp 1296bp 0bp},clip]%
    {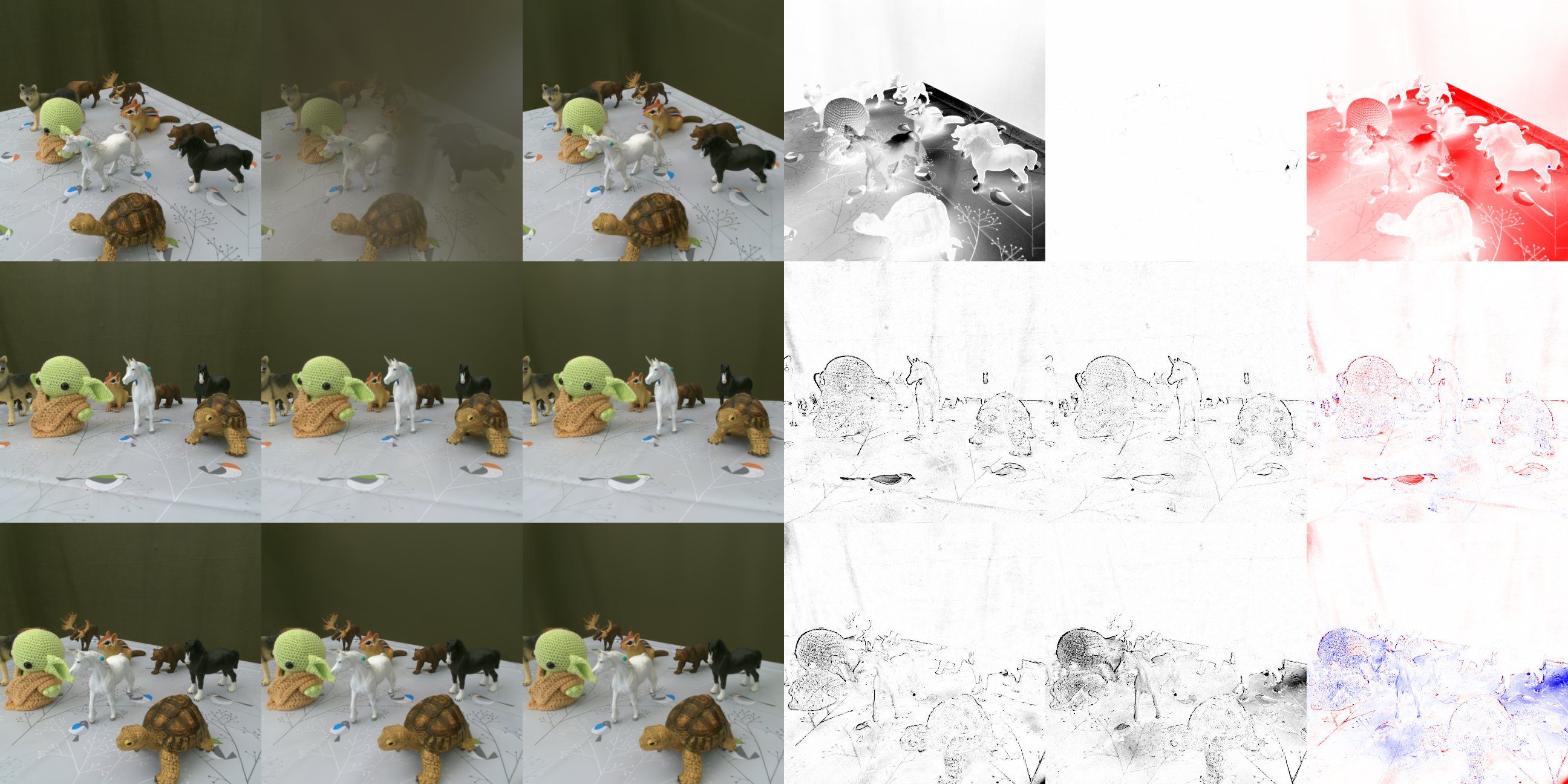}
  \par\vspace{1pt}
  {\fontsize{7}{7.8}\selectfont
    \makebox[0.333\linewidth][c]{GT}%
    \makebox[0.333\linewidth][c]{Standard 3DGS}%
    \makebox[0.333\linewidth][c]{RobustSplat}}
\end{minipage}
\fi
\vspace{-5pt}
\caption{\textbf{The remaining gains of specially-designed methods after input recovery largely come from mitigating general 3DGS training weaknesses.}
  (a) Once dense initial points are recovered, the tested depth-correlation loss provides less gain in sparse views.
  (b) With the same recovered masks, 3DGS-MCMC closes over two-thirds of the mean PSNR gap to RobustSplat across six scenes.
  (c) One example of such fitting problems is shown here, where near-camera floaters plague standard 3DGS.}
\label{fig:fig5_survive}
\vspace{-18pt}
\end{figure*}

\vspace{-10pt}
\paragraph{Learnable targets collapse toward early renderings.}
Modifying supervision images during training offers an even more direct failure path.
In methods like Luminance-GS~\cite{cui2025luminance}, a learnable tone curve $\Phi_\theta(I)$ adjusts the training image under a brightness prior while the photometric loss $\mathcal{L}_{\mathrm{photo}}(\mathcal{R}(G),\Phi_\theta(I))$ updates both the Gaussians $G$ and $\theta$.
Early in training, the easiest way for $\theta$ to reduce $\mathcal{L}_{\mathrm{photo}}$ against a blurry rendering $\mathcal{R}(G)$ is to flatten the contrast of the target image itself.

Testing this across three LOM scenes (\cref{fig:t4_moving_target}), we find that on restored images ($k=1.0$), enabling the full learnable target lowers mean color-calibrated PSNR (cc-PSNR) by about $4.6$\,dB (\cref{fig:t4_moving_target}a and \cref{tab:supp:collapse}).
On \texttt{bike}, the target image's contrast collapses to $14\%$ of the input by step 10k, washing out image details to match the blurry early rendering (\cref{fig:t4_moving_target}b--d).
Simply stopping the photometric gradient to $\theta$—so the tone curve is updated only by the brightness prior—preserves $92\%$ of the target contrast and restores cc-PSNR across brightness levels (\cref{fig:t4_moving_target}a, e).

\subsection{Why Do Other Components Remain Helpful?}
\label{sec:analysis_survive}

We now turn to our second question: for sparse views and temporary occlusions, why do specially-designed methods still outperform standard 3DGS after input recovery?
The following probes link some of the remaining gains to two weaknesses of standard 3DGS: overfitting under sparse views and fitting artifacts under temporary occlusions.

\vspace{-10pt}
\paragraph{From recovering missing geometry to preventing overfitting.}
In sparse views, monocular depth priors (\eg, FSGS~\cite{zhu2024fsgs}) are designed to fill in missing geometry when starting from sparse SfM points.
Our controlled probe uses an FSGS-style depth-correlation loss with VGGT-$\Omega$ training-view depth (\cref{sec:supp:sparse}). Once initialized with dense VGGT-$\Omega$ points, the initial geometry already covers all training pixels and correlates strongly with the depth prior ($r\geq0.82$ in 7 of 8 scenes), cutting the depth-loss PSNR gain by more than half on both LLFF and DTU (\cref{fig:fig5_survive}a).
Instead, the main challenge shifts to preventing dense Gaussians from overfitting to the few training views: random Gaussian dropout~\cite{park2025dropgaussian}, which hurts under sparse points ($-1.41$\,dB on LLFF), becomes beneficial ($+0.51$\,dB) once dense initialization is provided.

\vspace{-10pt}
\paragraph{From finding missed occlusions to fixing generic 3DGS artifacts.}
Temporary occlusions reveal this shift even more clearly.
Across six diagnostic scenes where both methods receive the exact same recovered masks and initial SfM points, RobustSplat~\cite{fu2025robustsplat} still leads standard 3DGS by $+1.64$\,dB.
One natural hypothesis is that RobustSplat's online masking catches moving objects missed by our recovered masks.
A 102-run control study tests several ways of expanding the masks. Excluding all high-residual regions flagged by the missed-distractor proxy narrows the mean gap by $0.31$\,dB (\cref{tab:supp:maskdose}).

If missed occlusions are not the main cause, where does the remaining gap come from?
The worst $10\%$ of test views account for approximately $91\%$ of the MSE gap on \texttt{yoda} and $83\%$ on \texttt{Cupcake}.
Per-view inspection reveals near-camera floating Gaussians that obscure the scene on \texttt{yoda} (\cref{fig:fig5_survive}c) and cross-view color inconsistencies on \texttt{MilkCookie} (\cref{fig:supp:errmap}).
To verify this, we keep the recovered masks and initial points completely fixed, and simply replace standard 3DGS training strategy with the task-agnostic 3DGS-MCMC one~\cite{kheradmand20243d}.
Without any occlusion-specific modules, this general training improvement closes over $2/3$ of the PSNR gap to RobustSplat (\cref{fig:fig5_survive}b and \cref{tab:supp:mcmc}) and achieves a lower mean LPIPS ($0.118$ vs.\ $0.138$).
Likewise, adding a generic per-view affine color adjustment on \texttt{MilkCookie} boosts standard 3DGS by $+2.74$\,dB, outperforming RobustSplat without altering the occlusion masks at all.

Together, these findings suggest that once foundation models recover degraded inputs, future progress depends less on adding task-specific input fixes during training, and more on improving the optimization dynamics of standard 3DGS itself—such as preserving accurate inputs early in training and suppressing floaters, overfitting, and cross-view color drift.

\vspace{-7pt}

\section{Conclusion}
\label{sec:conclusion}

\vspace{-4pt}

In this paper, we revisited 3DGS reconstruction for degraded inputs in the era of vision foundation models.
Across 148 test cases spanning four degraded-input tasks, we found that recovering the inputs with pretrained vision models makes standard 3DGS competitive with task-specific methods.

When both sides receive the same recovered information, the PSNR gain of task-specific methods shrinks in 86\% of matched comparisons. The average gain becomes negative for unknown camera poses and poor exposure.
Controlled experiments illustrate how this reversal can arise. Early rendering errors from unconverged Gaussians can drive joint optimization to alter accurate camera poses or reduce contrast in restored images. Where gains persist, some can be traced to general 3DGS fitting problems, including overfitting, floating Gaussians, and cross-view color differences.

These findings point to two priorities for 3DGS training: preserving accurate inputs during early optimization and reducing artifacts in Gaussian fitting.

\section*{LLM Usage Statement}

We used OpenAI GPT models, Google Gemini, and Anthropic Claude to assist with manuscript drafting and language polishing, narrative organization, LaTeX and figure preparation, and technical and citation consistency checks. These tools also assisted with experimental scripting, debugging, result aggregation, and analysis, and provided reviewer-style feedback on clarity, technical consistency, and presentation. The authors designed the study, determined the experimental protocol, verified the reported results and references, and take responsibility for the final manuscript.

{\small
\bibliographystyle{ieeenat_fullname}
\bibliography{main}
}

\clearpage
\appendix
\setcounter{table}{0}
\setcounter{figure}{0}
\setcounter{equation}{0}
\renewcommand{\thetable}{A\arabic{table}}
\renewcommand{\thefigure}{A\arabic{figure}}
\renewcommand{\theequation}{A\arabic{equation}}
\renewcommand{\theHtable}{supp.\arabic{table}}
\renewcommand{\theHfigure}{supp.\arabic{figure}}
\renewcommand{\theHequation}{supp.\arabic{equation}}
\iflatexml
  \section*{Supplementary Material}
\else
  \twocolumn[{\centering\Large\bfseries Supplementary Material\par\vspace{14pt}}]
\fi
\raggedbottom
\section*{Overview}
The following sections document the experiments reported in the main text.

\begin{table}[h]
    \centering
    \footnotesize
    \setlength{\tabcolsep}{4pt}
    \begin{tabular}{@{}p{0.43\columnwidth}p{0.49\columnwidth}@{}}
        \toprule
        \textbf{Main text} & \textbf{Supporting material} \\
        \midrule
        Input recovery (\cref{sec:method:separation}) & \cref{sec:supp:input} \\
        Evaluation protocol (\cref{sec:method:benchmark,sec:exp:setup}) & \cref{sec:supp:protocol} \\
        End-to-end results (\cref{tab:main}) & \cref{sec:supp:perscene} \\
        Matched inputs (\cref{tab:shared}) & \cref{sec:supp:matched} \\
        Component probes (\cref{sec:analysis}) & \cref{sec:supp:analysis} \\
        Qualitative examples (\cref{fig:qualitative}) & \cref{sec:supp:qualitative} \\
        \bottomrule
    \end{tabular}
\end{table}

\section{Input Recovery Details}
\label{sec:supp:input}

\begin{table}[htbp]
    \centering
    \footnotesize
    \setlength{\tabcolsep}{3pt}
    \begin{tabular}{@{}p{1.45cm}p{2.1cm}p{3.45cm}@{}}
        \toprule
        Task & Available Inputs & Missing or Unreliable Information \\
        \midrule
        Sparse Views & Few RGB views and known cameras & Incomplete geometric initialization \\
        Pose-Free & RGB views and known intrinsics & Camera extrinsics and initial geometry \\
        Transient & RGB views and known cameras & Static-scene supervision obscured by distractors \\
        Exposure & Low-light or over-exposed RGB views and known cameras & Photometrically degraded supervision \\
        \bottomrule
    \end{tabular}
    \caption{\textbf{Task definitions.} Inputs refer to the training views only. Every task outputs a static 3D reconstruction evaluated on held-out images.}
    \label{tab:supp:tasks}
\end{table}

\begin{figure*}[t]
\centering
\iflatexml
\includegraphics[width=\linewidth,alt={Examples of the geometric and image inputs recovered before Gaussian splatting for the four tasks.}]{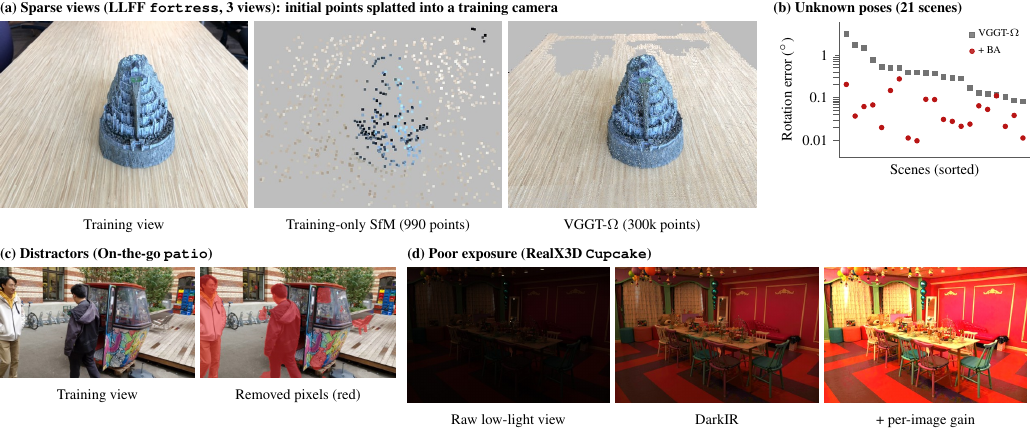}
\else
\setlength{\tabcolsep}{1.5pt}
\renewcommand{\arraystretch}{0.9}
\begin{minipage}[t]{0.735\linewidth}
  \raggedright{\scriptsize\bfseries (a) Sparse views (LLFF \texttt{fortress}, 3 views): initial points splatted into a training camera}\par\vspace{2pt}
  \begin{tabular}{@{}ccc@{}}
  \includegraphics[width=0.327\linewidth]{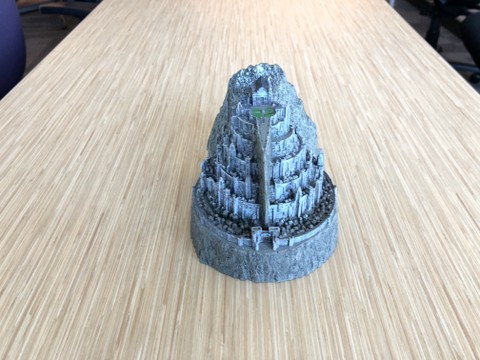} &
  \includegraphics[width=0.327\linewidth]{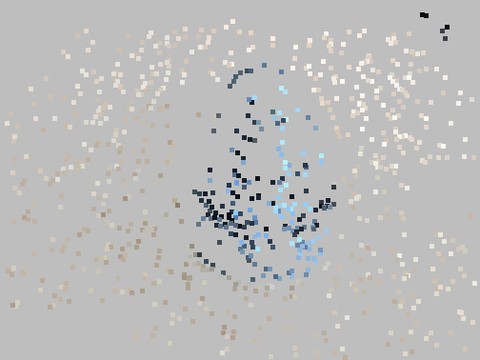} &
  \includegraphics[width=0.327\linewidth]{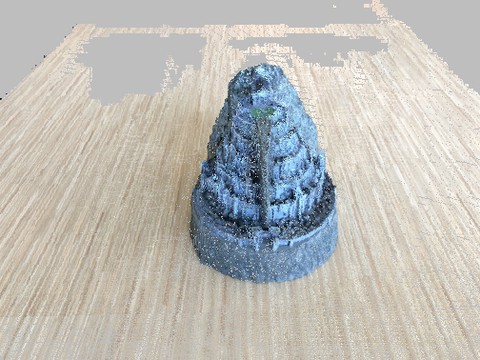} \\
  {\scriptsize Training view} & {\scriptsize Training-only SfM (990 points)} & {\scriptsize VGGT-$\Omega$ (300k points)} \\
  \end{tabular}
\end{minipage}\hfill
\begin{minipage}[t]{0.25\linewidth}
  \raggedright{\scriptsize\bfseries (b) Unknown poses (21 scenes)}\par\vspace{2pt}
  \begin{tikzpicture}
  \begin{axis}[irb, scale only axis, width=0.74\linewidth, height=2.3cm, ymode=log,
    ymin=0.004, ymax=6, xmin=-0.8, xmax=20.8, xtick=\empty,
    ytick={0.01,0.1,1}, yticklabels={0.01,0.1,1},
    xlabel={Scenes (sorted)}, ylabel={Rotation error ($^\circ$)},
    legend style={at={(0.98,0.99)},anchor=north east,font=\tiny,fill=white,fill opacity=0.85,text opacity=1}]
    \addplot[irbGray, only marks, mark=square*, mark size=1.1pt] coordinates {(0,3.1643) (1,1.7239) (2,1.4800) (3,0.7750) (4,0.5263) (5,0.5003) (6,0.4979) (7,0.3920) (8,0.3905) (9,0.3781) (10,0.3729) (11,0.3058) (12,0.2894) (13,0.2817) (14,0.1725) (15,0.1315) (16,0.1251) (17,0.1180) (18,0.1065) (19,0.0863) (20,0.0829)};
    \addlegendentry{VGGT-$\Omega$}
    \addplot[irbRed, only marks, mark=*, mark size=1.1pt] coordinates {(0,0.2059) (1,0.0375) (2,0.0625) (3,0.0682) (4,0.0202) (5,0.1484) (6,0.2771) (7,0.0115) (8,0.0099) (9,0.0924) (10,0.0911) (11,0.0313) (12,0.0282) (13,0.0216) (14,0.0243) (15,0.0651) (16,0.0536) (17,0.1115) (18,0.0215) (19,0.0388) (20,0.0115)};
    \addlegendentry{+ BA}
  \end{axis}
  \end{tikzpicture}
\end{minipage}
\par\vspace{5pt}
\begin{minipage}[t]{0.385\linewidth}
  \raggedright{\scriptsize\bfseries (c) Distractors (On-the-go \texttt{patio})}\par\vspace{2pt}
  \begin{tabular}{@{}cc@{}}
  \includegraphics[width=0.49\linewidth]{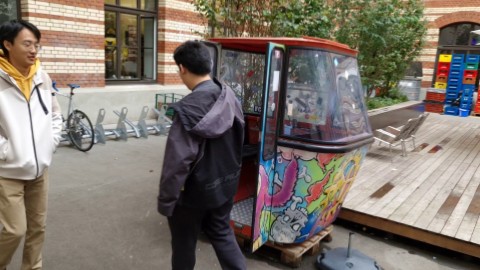} &
  \includegraphics[width=0.49\linewidth]{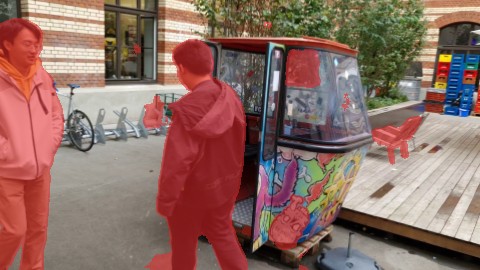} \\
  {\scriptsize Training view} & {\scriptsize Removed pixels (red)} \\
  \end{tabular}
\end{minipage}\hfill
\begin{minipage}[t]{0.6\linewidth}
  \raggedright{\scriptsize\bfseries (d) Poor exposure (RealX3D \texttt{Cupcake})}\par\vspace{2pt}
  \begin{tabular}{@{}ccc@{}}
  \includegraphics[width=0.327\linewidth]{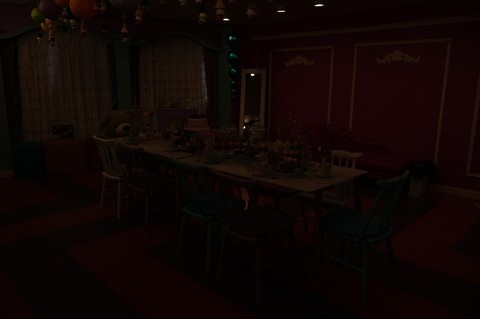} &
  \includegraphics[width=0.327\linewidth]{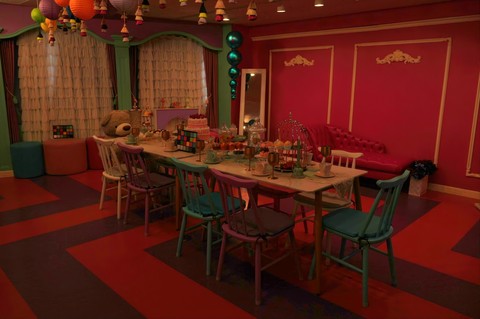} &
  \includegraphics[width=0.327\linewidth]{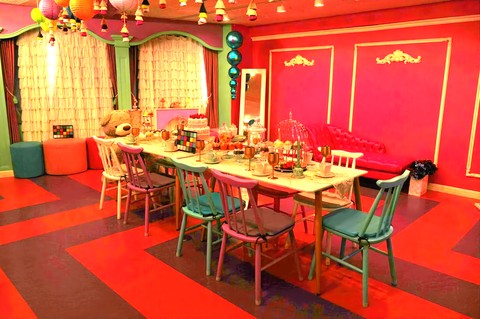} \\
  {\scriptsize Raw low-light view} & {\scriptsize DarkIR} & {\scriptsize + per-image gain} \\
  \end{tabular}
\end{minipage}
\fi
\vspace{-4pt}
\caption{\textbf{Inputs recovered by pretrained vision models before standard 3DGS training} (Sec.~\ref{sec:method:separation} in the main text; training views only).
(a) Dense VGGT-$\Omega$ points replace the sparse training-only SfM cloud (gray: no point).
(b) Fixed-intrinsics BA reduces the mean rotation error of the VGGT-$\Omega$ training cameras from $0.567^\circ$ to $0.068^\circ$ and lowers it on every pose-free scene.
(c) Object-level masks remove distractors from the photometric loss.
(d) Restoration followed by a per-image gain gives well-exposed targets.}
\label{fig:supp:recovery}
\vspace{-6pt}
\end{figure*}

\cref{tab:supp:tasks} summarizes what each task provides and what is missing, and \cref{fig:supp:recovery} shows an example of the recovered
  inputs for each task.

Each task uses the same recovery procedure across its scenes. All procedures use only the permitted training observations and finish before Gaussian optimization begins.
The recovery steps never read test images or test cameras.
All recovered inputs are then passed to the same unmodified 3DGS trainer (\cref{sec:supp:protocol}).

\paragraph{Sparse views.}
We run VGGT-$\Omega$~\cite{wang2026vggtomega} on the training views only and obtain per-view depth and confidence maps.
The predicted depth is brought to the scale of the known cameras with one global factor per scene.
When a training-only SfM cloud projects onto at least 50 training pixels, the factor is the median ratio between SfM depth and predicted depth on those pixels; otherwise it is the scale of a Sim(3) (Umeyama) fit between predicted and known training camera centers.
Each training pixel is back-projected with the known intrinsics and pose of its view.
We drop the 20\% of pixels with the lowest confidence in each scene, take colors from the original training RGB, voxel-downsample with a voxel equal to one processed-pixel footprint at the median depth, and keep at most 300k points with a deterministic subsampling.
For Blender, pure-white background pixels are not back-projected, since otherwise they form a white shell around the object.

\paragraph{Unknown camera poses (pose-free).}
VGGT-$\Omega$ predicts initial extrinsics and depth from the unposed training images.
Because VGGT-$\Omega$ also predicts its own intrinsics, errors in its focal length are partly absorbed into the poses.
We therefore refine the extrinsics with the dataset's known intrinsics held fixed.
We extract SIFT features on the training images and match pairs that are pose neighbors or depth co-visibility neighbors under the VGGT-$\Omega$ prediction.
Tracks are triangulated with the VGGT-$\Omega$ poses held fixed, followed by two rounds of bundle adjustment~\cite{triggs1999bundle} over poses and points with intrinsics fixed, re-triangulating between rounds and removing points with mean reprojection error above 4 pixels.
Views that cannot be registered keep their VGGT-$\Omega$ pose.
The initial point cloud back-projects the VGGT-$\Omega$ depth with the known intrinsics and the refined poses, with each view's depth rescaled by its median ratio to the BA points (at most 300k points).
This step reduces the mean rotation error of the training cameras from $0.567^\circ$ to $0.068^\circ$ over the 21 scenes.
The trainer uses near plane 0.2 as in the other tasks.
On the nine scenes whose predicted scene extent is far below 1 (all eight Tanks and Temples scenes and CO3D 110), this near plane clips the scene, so Ours is trained on them with a near plane relative to the extent ($0.2\times\min(\text{extent},1)$); on the other scenes the extent is at least 0.59, so the two settings are nearly the same (and identical when the extent is at least 1).

\paragraph{Temporary occlusions (distractors).}
For temporary occlusions, we build binary object-level masks from the training images only.
SAM~3.1~\cite{carion2026sam3} proposes complete candidate object regions; semantic labels only propose regions and never decide deletion on their own.
Each candidate is then accepted or rejected as a distractor from multiview evidence computed with VGGT-$\Omega$ depth and DINOv2~\cite{oquab2023dinov2} features: RGB and feature correspondence across training views, and geometric consistency, where a projected point within 15\% of the observed depth in another view counts as support and a point more than 20\% in front of reliable observed depth counts as a contradiction.
A geometric contradiction is considered strong only if it appears in at least three views from at least two pose groups, with a group-balanced contradiction fraction of at least 0.35.
Candidates without sufficient evidence are kept as supervision, so missing evidence is never a reason to delete.
The accepted masks remove distractor pixels from the photometric loss.
Specifically, in the end-to-end comparison (Stage~1), the initial cloud back-projects VGGT-$\Omega$ depth from masked-in training pixels only, so points on removed distractors never enter it.
We keep only pixels with confidence above the 20th percentile and small relative depth gradient ($<0.05$), require support from at least two other views and reject points with strong geometric contradictions (same thresholds as above), and voxel-downsample to at most 600k points.
In the matched-input comparison (Stage~2) and the transient probes of Sec.~\ref{sec:analysis_survive} in the main text, both sides instead use the same training-only SfM cloud, so that only the masks are shared.
The same models, thresholds, and seed are used for all 19 scenes, without per-scene selection.

\paragraph{Poor exposure.}
For low light, each training image is enhanced by a frozen DarkIR~\cite{feijoo2025darkir} model.
For over-exposure, we use a frozen LCDPNet~\cite{wang2022lcdpnet} model.
Each restored image is then multiplied by its own scalar gain, chosen so that its Rec.~709 mean luminance would be 0.49 before clipping to $[0,1]$; this removes exposure differences between views.
The initial point cloud comes from VGGT-$\Omega$ depth on the restored images, scaled with a fixed-camera training-only SfM cloud of the same images, with confidence filtering as in the sparse-view setting.
Evaluation uses the normal-exposure ground truth of each dataset.

\section{Evaluation Protocol}
\label{sec:supp:protocol}

\begin{table}[t]
    \centering
    \footnotesize
    \setlength{\tabcolsep}{3pt}
    \begin{tabular}{@{}l p{2.0cm} p{3.55cm}@{}}
        \toprule
        Setting (Configs) & Datasets & Compared Methods \\
        \midrule
        Sparse Views (91) & \raggedright LLFF, DTU, Mip360, Blender & \raggedright DNGaussian~\cite{li2024dngaussian}, FSGS~\cite{zhu2024fsgs}, CoR-GS~\cite{zhang2024corgs}, DropGaussian~\cite{park2025dropgaussian} \tabularnewline
        Pose-Free (21) & \raggedright T\&T, CO3D, Mip360, DTU & \raggedright CF-3DGS~\cite{fu2024colmapfree}, HT-3DGS~\cite{ji2025htgs}, MASt3R~\cite{leroy2024mast3r}+3DGS, 3R-GS~\cite{huang2025threergs} \tabularnewline
        Transient (19) & \raggedright RobustNeRF, On-the-go, RealX3D & \raggedright SpotLessSplats~\cite{sabour2025spotless}, WildGaussians~\cite{kulhanek2024wildgaussians}, DeSplat~\cite{wang2025desplat}, RobustSplat~\cite{fu2025robustsplat} \tabularnewline
        Low Light (14) & \raggedright LOM, RealX3D & \raggedright Luminance-GS~\cite{cui2025luminance}, LITA-GS~\cite{zhou2025litags} (LOM), LL-Gaussian~\cite{sun2025llgaussian} \tabularnewline
        Over-Exposure (3) & \raggedright LOM & \raggedright Luminance-GS, Aleth-NeRF~\cite{cui2024aleth} \tabularnewline
        \bottomrule
    \end{tabular}
    \caption{\textbf{Evaluation coverage across 148 test cases and 10 distinct datasets.}
    A test case (configuration) is one scene, or one scene at one training-view count for sparse views.
    Standard 3DGS is included in every setting; pose-free reconstruction additionally reports a COLMAP-pose reference~\cite{schoenberger2016sfm}.}
    \label{tab:supp:setup}
\end{table}

\paragraph{Datasets and splits.}
\cref{tab:supp:setup} lists the datasets and compared methods.
For sparse views, LLFF (8 scenes) and DTU (15 scans) use 3, 6, and 9 training views, Mip-NeRF 360 (7 scenes) uses 12 and 24 views, and Blender (8 scenes) uses 8 views.
LLFF and Mip-NeRF 360 follow the FSGS split: every 8th image is a test view and the training views are sampled uniformly from the rest.
DTU uses the standard training-view order and 25 test views, and is evaluated with object masks at 400$\times$300.
Blender uses the 8 training views of DietNeRF and every 8th image of the official test set, at 400$\times$400 with a white background.
For pose-free reconstruction, Tanks and Temples (8) and CO3D (5) follow the CF-3DGS/HT-3DGS protocol, and Mip-NeRF 360 (4) and DTU (4) follow the 3R-GS protocol.
Temporary-occlusion scenes use the official clean test views of RobustNeRF (5), On-the-go (6), and the RealX3D dynamic subset (8).
Poor-exposure scenes use LOM (5 low-light, 3 over-exposed) and the RealX3D extreme low-light subset (9), evaluated against the normal-exposure images.
Two further LOM over-exposure scenes (\texttt{buu}, \texttt{sofa}) were used to choose the over-exposure restorer and are excluded from the reported over-exposure results.

\paragraph{Metrics.}
All methods are evaluated by the same evaluator on the same held-out views.
We report PSNR, SSIM (11$\times$11 Gaussian window, $\sigma{=}1.5$), and LPIPS with a VGG backbone, averaged over test views and then over scenes.

\paragraph{Pose-free evaluation.}
Test cameras are never optimized.
For each method, a Sim(3) transform is fitted between its predicted training cameras and the reference training cameras, and the same transform maps the reference test cameras into the method's frame.
Rotation errors are measured on training cameras after this alignment.
Predicted reconstructions have arbitrary scale, so a fixed near plane can clip scenes whose predicted scale is small.
We therefore re-rendered the saved models of all pose-free methods on Tanks and Temples, CO3D, and two Mip-NeRF 360 scenes (\texttt{counter}, \texttt{garden}) with a near plane equal to 1\% of the median distance between training cameras, converted to each method's frame by the training-only alignment.
The remaining six scenes (all four DTU scans, Mip-NeRF 360 \texttt{bicycle} and \texttt{room}) keep the original evaluation with near plane 0.2; their predicted scene extents are between 0.78 and 1.15, where this choice has little effect (on the re-rendered Mip-NeRF 360 scenes, the two evaluations differ by 0.08\,dB on average).
Within a scene, all methods use the same evaluation.
Failed re-renderings follow the failure rule below.

\paragraph{Failures.}
A run is \emph{failed} when it crashes, cannot run on the test case (\eg, its training-only SfM initialization is empty), or produces no renderable output.
A failed run takes the worst score per metric among the compared methods on the same test case (excluding the COLMAP reference).
Runs that finish but collapse keep their measured scores.
Subscripts in Tab.~\ref{tab:main} report coverage. For sparse views, this count also excludes completed runs whose PSNR is more than 3\,dB below standard 3DGS on the same test case. These runs retain their measured scores in the reported means.
Failed cells are marked in the per-scene tables (\cref{sec:supp:perscene}).

\paragraph{Baseline implementations.}
We use the authors' official code for every specially-designed method, with adapters that only convert our data format, splits, and cameras to each repository's input format.
Each method keeps its official hyperparameters, losses, and number of iterations.
We checked each adapter's camera projection against our reference cameras.
This check revealed that several Gaussian splatting repositories build the projection from the field of view only and assume a centered principal point.
On DTU, LOM, and RealX3D, whose calibrations are off-center by up to 26 pixels, we therefore pass the full intrinsic matrix to FSGS, CoR-GS, DropGaussian, DNGaussian, LITA-GS, and LL-Gaussian; all reported runs use these corrected adapters.
LITA-GS is reported on LOM only because of its training cost at RealX3D resolution, and Aleth-NeRF is reported on over-exposure only.

\paragraph{Shared 3DGS trainer.}
Ours and every standard 3DGS entry use the same gsplat-based trainer with the original 3DGS recipe: 30k steps, SH degree 3, loss $0.8\,\mathcal{L}_1 + 0.2\,\mathcal{L}_{\text{D-SSIM}}$, densification every 100 steps from step 500 to 15k with gradient threshold $2\times10^{-4}$, opacity reset every 3k steps, and the original learning rates.
Pose and appearance optimization are disabled, and the trainer, its settings, and the seed are identical across tasks; only the inputs change, apart from the scale-relative near plane on nine pose-free scenes (\cref{sec:supp:input}).

\section{Per-Scene Results}
\label{sec:supp:perscene}

\cref{tab:supp:sparse-a,tab:supp:sparse-b,tab:supp:posefree,tab:supp:transient,tab:supp:exposure} report PSNR, SSIM, and LPIPS for every test case in Tab.~\ref{tab:main}.
The dataset means in these tables reproduce the values in Tab.~\ref{tab:main}.

\paragraph{Per-case best method.}
Sec.~\ref{sec:exp:main} in the main text also compares standard 3DGS with recovered inputs (Ours) against the best-performing specially-designed method on each test case, chosen separately for PSNR and for LPIPS.
Two of the 148 test cases, DTU \texttt{scan30} and \texttt{scan110} with three training views, have no completed run of any specialized method. Excluding these leaves 146 comparable cases.
Ours achieves higher PSNR on 63 of them (sparse views 27/89, unknown poses 19/21, temporary occlusions 1/19, poor exposure 16/17) and better LPIPS on 101 (52/89, 20/21, 13/19, 16/17).

On the 14 low-light scenes, an additional baseline applies only scalar brightness alignment to the training images before standard 3DGS. Its mean PSNR is 16.89\,dB, compared with 15.46\,dB for LL-Gaussian and 19.63\,dB for the full input-recovery pipeline.

\begin{table*}[t]
\centering
\begingroup
\setlength{\tabcolsep}{2.25pt}
\renewcommand{\arraystretch}{0.96}
\resizebox{\textwidth}{!}{%
\begin{tabular}{@{}l@{\hspace{6pt}}cccccc@{\hspace{6pt}}cccccc@{\hspace{6pt}}cccccc@{}}
\toprule
 & \multicolumn{6}{c}{PSNR$\uparrow$} & \multicolumn{6}{c}{SSIM$\uparrow$} & \multicolumn{6}{c}{LPIPS$\downarrow$} \\
\cmidrule(lr){2-7}\cmidrule(lr){8-13}\cmidrule(lr){14-19}
Scene & 3DGS & DNGS & FSGS & CoR & Drop & Ours & 3DGS & DNGS & FSGS & CoR & Drop & Ours & 3DGS & DNGS & FSGS & CoR & Drop & Ours \\
\midrule
\multicolumn{19}{@{}l}{\textit{LLFF (24)}} \\
fern (3) & 14.16 & \textbf{20.95} & 16.29 & 16.71 & 17.73 & 17.51 & 0.353 & \textbf{0.674} & 0.445 & 0.470 & 0.496 & 0.546 & 0.465 & \textbf{0.268} & 0.433 & 0.392 & 0.369 & 0.320 \\
fern (6) & 21.13 & 23.26 & 24.76 & 24.88 & \textbf{25.31} & 21.56 & 0.678 & 0.766 & 0.832 & 0.835 & \textbf{0.845} & 0.732 & 0.239 & 0.230 & 0.127 & 0.126 & \textbf{0.121} & 0.194 \\
fern (9) & 23.07 & 23.96 & \textbf{26.24} & 26.12 & 26.18 & 23.52 & 0.771 & 0.788 & 0.875 & 0.875 & \textbf{0.877} & 0.792 & 0.177 & 0.216 & 0.102 & \textbf{0.095} & 0.101 & 0.154 \\
room (3) & 12.04 & 16.67 & \textbf{18.68} & 15.43 & 17.56 & 18.48 & 0.383 & 0.622 & 0.669 & 0.559 & 0.650 & \textbf{0.723} & 0.569 & 0.425 & 0.337 & 0.415 & 0.340 & \textbf{0.277} \\
room (6) & 22.71 & 25.66 & 27.70 & 28.80 & \textbf{28.82} & 26.29 & 0.800 & 0.885 & 0.922 & \textbf{0.934} & 0.933 & 0.897 & 0.229 & 0.185 & 0.103 & \textbf{0.085} & 0.088 & 0.115 \\
room (9) & 22.11 & 24.98 & 28.60 & \textbf{29.69} & 29.28 & 25.75 & 0.801 & 0.890 & 0.938 & \textbf{0.947} & 0.943 & 0.907 & 0.245 & 0.183 & 0.092 & \textbf{0.075} & 0.079 & 0.115 \\
trex (3) & 12.21 & 19.39 & 19.79 & 19.35 & \textbf{20.12} & 14.16 & 0.383 & 0.689 & 0.720 & 0.711 & \textbf{0.741} & 0.526 & 0.468 & 0.276 & 0.222 & 0.197 & \textbf{0.186} & 0.362 \\
trex (6) & 14.76 & 21.82 & \textbf{23.56} & 23.39 & 23.20 & 19.48 & 0.586 & 0.785 & 0.854 & \textbf{0.858} & 0.856 & 0.754 & 0.366 & 0.194 & 0.132 & \textbf{0.115} & 0.118 & 0.213 \\
trex (9) & 19.96 & 23.00 & 25.22 & 26.32 & \textbf{26.44} & 22.32 & 0.793 & 0.820 & 0.897 & \textbf{0.916} & \textbf{0.916} & 0.839 & 0.204 & 0.177 & 0.103 & 0.079 & \textbf{0.077} & 0.151 \\
horns (3) & 15.24 & \textbf{18.45} & 17.98 & 16.92 & 17.31 & 16.72 & 0.408 & 0.608 & \textbf{0.618} & 0.587 & 0.581 & 0.603 & 0.433 & 0.351 & \textbf{0.283} & 0.304 & 0.316 & 0.294 \\
horns (6) & 17.94 & 21.92 & \textbf{23.94} & 22.97 & 22.93 & 20.05 & 0.591 & 0.751 & \textbf{0.840} & 0.826 & 0.824 & 0.707 & 0.320 & 0.269 & \textbf{0.150} & 0.154 & 0.155 & 0.239 \\
horns (9) & 22.08 & 23.26 & \textbf{25.06} & 24.55 & 24.84 & 22.49 & 0.774 & 0.788 & 0.881 & 0.881 & \textbf{0.884} & 0.808 & 0.194 & 0.254 & 0.127 & 0.114 & \textbf{0.112} & 0.161 \\
flower (3) & 16.02 & \textbf{18.80} & 15.77 & 15.37 & 16.90 & 17.45 & 0.387 & \textbf{0.571} & 0.355 & 0.365 & 0.419 & 0.475 & 0.424 & \textbf{0.312} & 0.432 & 0.420 & 0.384 & 0.365 \\
flower (6) & 22.37 & \textbf{23.32} & 23.26 & 22.28 & 22.74 & 22.93 & 0.729 & 0.756 & \textbf{0.778} & 0.755 & 0.768 & 0.743 & 0.178 & 0.192 & \textbf{0.146} & 0.158 & 0.153 & 0.170 \\
flower (9) & 24.79 & 24.59 & \textbf{26.25} & 25.32 & 25.49 & 24.57 & 0.814 & 0.794 & \textbf{0.854} & 0.828 & 0.833 & 0.805 & 0.124 & 0.176 & \textbf{0.098} & 0.122 & 0.117 & 0.133 \\
leaves (3) & 12.12 & \textbf{15.26} & 10.84 & 11.73 & 11.86 & 13.55 & 0.228 & \textbf{0.465} & 0.108 & 0.165 & 0.179 & 0.357 & 0.465 & \textbf{0.320} & 0.546 & 0.481 & 0.470 & 0.377 \\
leaves (6) & 14.05 & \textbf{17.70} & 15.15 & 13.42 & 15.23 & 15.22 & 0.392 & \textbf{0.605} & 0.445 & 0.289 & 0.473 & 0.478 & 0.391 & \textbf{0.266} & 0.314 & 0.429 & 0.311 & 0.326 \\
leaves (9) & 15.83 & \textbf{17.95} & 17.05 & 13.92 & 15.39 & 16.60 & 0.523 & \textbf{0.630} & 0.575 & 0.316 & 0.427 & 0.579 & 0.314 & 0.262 & \textbf{0.255} & 0.447 & 0.334 & 0.272 \\
orchids (3) & 14.30 & 15.35 & \textbf{16.04} & 15.27 & 15.80 & 15.50 & 0.392 & 0.440 & 0.488 & 0.460 & 0.484 & \textbf{0.490} & 0.347 & 0.319 & 0.271 & 0.284 & 0.270 & \textbf{0.259} \\
orchids (6) & 15.96 & 17.15 & \textbf{17.81} & 16.96 & 17.36 & 16.13 & 0.519 & 0.543 & \textbf{0.602} & 0.579 & 0.594 & 0.534 & 0.280 & 0.273 & 0.215 & 0.225 & \textbf{0.213} & 0.250 \\
orchids (9) & 16.79 & 18.03 & \textbf{18.53} & 18.13 & 18.40 & 16.47 & 0.574 & 0.590 & \textbf{0.653} & 0.640 & \textbf{0.653} & 0.563 & 0.249 & 0.259 & 0.190 & 0.194 & \textbf{0.184} & 0.251 \\
fortress (3) & 17.08 & 20.97 & 23.62 & \textbf{23.91} & 23.61 & 19.78 & 0.479 & 0.429 & 0.746 & \textbf{0.775} & 0.760 & 0.761 & 0.317 & 0.327 & 0.169 & \textbf{0.149} & 0.163 & 0.159 \\
fortress (6) & 21.17 & 24.80 & 27.09 & \textbf{27.34} & 27.15 & 22.71 & 0.741 & 0.596 & 0.865 & \textbf{0.873} & 0.871 & 0.827 & 0.181 & 0.389 & 0.106 & \textbf{0.088} & 0.098 & 0.130 \\
fortress (9) & 24.04 & 25.58 & \textbf{28.74} & 27.78 & 27.99 & 25.33 & 0.790 & 0.623 & \textbf{0.881} & 0.876 & 0.880 & 0.816 & 0.147 & 0.402 & 0.086 & \textbf{0.082} & 0.092 & 0.126 \\
\textit{Mean} & 18.00 & 20.95 & 21.58 & 21.11 & 21.57 & 19.77 & 0.579 & 0.671 & 0.702 & 0.680 & 0.704 & 0.678 & 0.305 & 0.272 & 0.210 & 0.218 & 0.202 & 0.226 \\
\midrule
\multicolumn{19}{@{}l}{\textit{Mip-NeRF 360 (14)}} \\
room (12) & 19.08 & 12.88 & 20.07 & \textbf{20.24} & 19.49 & 19.98 & 0.637 & 0.436 & 0.708 & \textbf{0.728} & 0.709 & 0.705 & 0.371 & 0.618 & 0.344 & 0.325 & 0.349 & \textbf{0.301} \\
room (24) & 23.42 & 13.96 & 24.55 & \textbf{24.76} & 24.39 & 24.17 & 0.788 & 0.493 & 0.834 & \textbf{0.840} & 0.830 & 0.812 & 0.248 & 0.617 & 0.214 & \textbf{0.210} & 0.228 & 0.212 \\
stump (12) & 15.32 & 16.98 & \textbf{17.50} & 17.31 & 17.12 & 16.32 & 0.193 & \textbf{0.325} & 0.309 & 0.304 & 0.289 & 0.204 & 0.580 & 0.674 & 0.624 & 0.631 & 0.626 & \textbf{0.551} \\
stump (24) & 21.07 & 18.43 & 22.24 & 21.68 & \textbf{22.55} & 21.54 & 0.520 & 0.348 & 0.574 & 0.547 & \textbf{0.582} & 0.559 & 0.367 & 0.656 & 0.405 & 0.420 & 0.406 & \textbf{0.338} \\
bonsai (12) & 16.96 & 13.85 & 19.14 & \textbf{19.56} & 19.10 & 17.65 & 0.592 & 0.404 & 0.649 & \textbf{0.658} & \textbf{0.658} & 0.635 & 0.375 & 0.601 & 0.352 & 0.360 & 0.350 & \textbf{0.325} \\
bonsai (24) & 21.80 & 15.18 & \textbf{24.27} & 23.88 & 24.08 & 22.80 & 0.790 & 0.478 & \textbf{0.827} & 0.819 & 0.823 & 0.823 & 0.231 & 0.596 & 0.215 & 0.222 & 0.229 & \textbf{0.190} \\
garden (12) & 17.93 & 14.17 & 19.42 & 19.51 & \textbf{19.74} & 18.75 & 0.447 & 0.286 & 0.508 & 0.507 & \textbf{0.522} & 0.498 & 0.380 & 0.749 & 0.411 & 0.427 & 0.414 & \textbf{0.334} \\
garden (24) & 22.36 & 15.91 & 23.46 & 23.18 & \textbf{23.60} & 22.05 & 0.705 & 0.305 & 0.731 & 0.723 & \textbf{0.732} & 0.702 & 0.204 & 0.717 & 0.225 & 0.234 & 0.250 & \textbf{0.198} \\
bicycle (12) & 15.71 & 14.51 & 18.03 & \textbf{18.35} & 17.93 & 16.21 & 0.252 & 0.273 & 0.365 & \textbf{0.369} & 0.367 & 0.239 & 0.516 & 0.713 & 0.565 & 0.577 & 0.572 & \textbf{0.500} \\
bicycle (24) & 18.78 & 15.65 & 20.65 & 19.50 & \textbf{20.93} & 19.69 & 0.396 & 0.299 & 0.462 & 0.448 & \textbf{0.471} & 0.457 & 0.413 & 0.709 & 0.495 & 0.503 & 0.491 & \textbf{0.349} \\
counter (12) & 16.73 & 12.22 & 18.31 & \textbf{19.07} & 18.06 & 17.61 & 0.504 & 0.357 & 0.592 & \textbf{0.622} & 0.609 & 0.585 & 0.416 & 0.671 & 0.390 & 0.377 & 0.383 & \textbf{0.348} \\
counter (24) & 21.20 & 13.12 & 23.03 & \textbf{23.10} & 22.91 & 22.42 & 0.720 & 0.409 & 0.768 & 0.767 & \textbf{0.774} & 0.755 & 0.257 & 0.668 & 0.244 & 0.236 & 0.248 & \textbf{0.219} \\
kitchen (12) & 19.68 & 13.18 & 18.85 & 19.73 & \textbf{19.81} & 19.61 & 0.665 & 0.330 & 0.694 & 0.722 & \textbf{0.729} & 0.715 & 0.299 & 0.692 & 0.293 & 0.273 & 0.274 & \textbf{0.266} \\
kitchen (24) & 23.03 & 14.18 & 23.05 & 22.92 & \textbf{23.47} & 20.73 & 0.825 & 0.355 & 0.835 & 0.835 & \textbf{0.842} & 0.786 & 0.170 & 0.679 & \textbf{0.164} & \textbf{0.164} & 0.173 & 0.201 \\
\textit{Mean} & 19.51 & 14.59 & 20.90 & 20.91 & 20.94 & 19.97 & 0.574 & 0.364 & 0.633 & 0.635 & 0.638 & 0.605 & 0.345 & 0.669 & 0.353 & 0.354 & 0.357 & 0.309 \\
\midrule
\multicolumn{19}{@{}l}{\textit{Blender (8)}} \\
mic (8) & 28.03 & \textbf{28.09} & 21.79 & 28.07 & 27.93 & 27.72 & 0.959 & 0.959 & 0.922 & \textbf{0.960} & 0.958 & 0.958 & \textbf{0.034} & \textbf{0.034} & 0.074 & 0.035 & 0.036 & 0.035 \\
lego (8) & 22.58 & 23.74 & 17.28 & 22.03 & 23.19 & \textbf{25.10} & 0.849 & 0.866 & 0.814 & 0.859 & 0.866 & \textbf{0.897} & 0.121 & 0.109 & 0.141 & 0.118 & 0.111 & \textbf{0.084} \\
ship (8) & 21.57$^\dagger$ & 21.57 & 21.57$^\dagger$ & 21.57$^\dagger$ & 21.57$^\dagger$ & \textbf{22.24} & 0.768$^\dagger$ & 0.768 & 0.768$^\dagger$ & 0.768$^\dagger$ & 0.768$^\dagger$ & \textbf{0.787} & 0.183$^\dagger$ & 0.183 & 0.183$^\dagger$ & 0.183$^\dagger$ & 0.183$^\dagger$ & \textbf{0.160} \\
chair (8) & 25.33 & 25.05 & 17.69 & \textbf{25.92} & 25.09 & 25.46 & 0.910 & 0.907 & 0.855 & \textbf{0.921} & 0.916 & 0.912 & \textbf{0.069} & 0.071 & 0.174 & 0.071 & 0.077 & 0.074 \\
drums (8) & 19.23 & \textbf{20.34} & 15.34 & 17.02 & 15.79 & 18.98 & 0.842 & \textbf{0.860} & 0.769 & 0.789 & 0.782 & 0.839 & 0.119 & \textbf{0.101} & 0.230 & 0.196 & 0.204 & 0.126 \\
ficus (8) & 20.11$^\dagger$ & \textbf{25.10} & 20.11$^\dagger$ & 20.11$^\dagger$ & 20.11$^\dagger$ & 20.11 & 0.870$^\dagger$ & \textbf{0.935} & 0.870$^\dagger$ & 0.870$^\dagger$ & 0.870$^\dagger$ & 0.870 & 0.106$^\dagger$ & \textbf{0.049} & 0.106$^\dagger$ & 0.106$^\dagger$ & 0.106$^\dagger$ & 0.106 \\
hotdog (8) & 27.45 & 28.88 & 20.00 & \textbf{29.36} & 28.65 & 25.51 & 0.930 & 0.944 & 0.905 & \textbf{0.946} & 0.942 & 0.921 & 0.066 & \textbf{0.055} & 0.099 & \textbf{0.055} & 0.063 & 0.079 \\
materials (8) & 18.43 & \textbf{20.05} & 17.23 & 19.94 & \textbf{20.05} & 18.71 & 0.803 & 0.833 & 0.793 & \textbf{0.838} & 0.836 & 0.814 & 0.151 & 0.123 & 0.171 & 0.124 & \textbf{0.120} & 0.144 \\
\textit{Mean} & 22.84 & 24.10 & 18.88 & 23.00 & 22.80 & 22.98 & 0.866 & 0.884 & 0.837 & 0.869 & 0.867 & 0.875 & 0.106 & 0.090 & 0.147 & 0.111 & 0.112 & 0.101 \\
\bottomrule
\end{tabular}%
}
\endgroup
\caption{\textbf{Per-scene results: sparse views (LLFF, Mip-NeRF 360, Blender).} Bold marks the best method in each row. $^\dagger$ marks failed runs, which take the worst score among the compared methods on that test case (Sec.~\ref{sec:exp:setup} in the main text); dataset means include these values and match Tab.~\ref{tab:main}. The number in parentheses after each scene is the number of training views. DNGS, CoR and Drop denote DNGaussian, CoR-GS and DropGaussian.}
\label{tab:supp:sparse-a}
\end{table*}

\begin{table*}[t]
\centering
\begingroup
\setlength{\tabcolsep}{2.25pt}
\renewcommand{\arraystretch}{0.96}
\resizebox{\textwidth}{!}{%
\begin{tabular}{@{}l@{\hspace{6pt}}cccccc@{\hspace{6pt}}cccccc@{\hspace{6pt}}cccccc@{}}
\toprule
 & \multicolumn{6}{c}{PSNR$\uparrow$} & \multicolumn{6}{c}{SSIM$\uparrow$} & \multicolumn{6}{c}{LPIPS$\downarrow$} \\
\cmidrule(lr){2-7}\cmidrule(lr){8-13}\cmidrule(lr){14-19}
Scene & 3DGS & DNGS & FSGS & CoR & Drop & Ours & 3DGS & DNGS & FSGS & CoR & Drop & Ours & 3DGS & DNGS & FSGS & CoR & Drop & Ours \\
\midrule
\multicolumn{19}{@{}l}{\textit{DTU (45)}} \\
scan8 (3) & 8.48 & 8.00$^\dagger$ & 8.00 & 16.44 & 15.04 & \textbf{21.06} & 0.587 & 0.587$^\dagger$ & 0.606 & 0.769 & 0.747 & \textbf{0.850} & 0.375 & 0.403$^\dagger$ & 0.403 & 0.250 & 0.276 & \textbf{0.124} \\
scan8 (6) & 18.89 & 18.89$^\dagger$ & \textbf{25.11} & 21.61 & 21.92 & 24.28 & 0.781 & 0.781$^\dagger$ & \textbf{0.877} & 0.842 & 0.859 & 0.868 & 0.197 & 0.197$^\dagger$ & 0.156 & 0.157 & 0.155 & \textbf{0.102} \\
scan8 (9) & 22.27 & 13.62 & \textbf{27.11} & 26.29 & 27.08 & 26.55 & 0.864 & 0.709 & \textbf{0.913} & 0.890 & 0.902 & 0.889 & 0.132 & 0.308 & 0.117 & 0.121 & 0.115 & \textbf{0.092} \\
scan21 (3) & 10.01 & 10.01$^\dagger$ & 12.99 & 12.76 & 13.25 & \textbf{17.06} & 0.412 & 0.412$^\dagger$ & 0.521 & 0.577 & 0.591 & \textbf{0.722} & 0.459 & 0.459$^\dagger$ & 0.337 & 0.295 & 0.304 & \textbf{0.185} \\
scan21 (6) & 17.10 & 10.93 & 17.46 & 17.70 & 17.47 & \textbf{20.58} & 0.763 & 0.369 & 0.793 & 0.807 & 0.804 & \textbf{0.850} & 0.175 & 0.527 & 0.165 & 0.138 & 0.151 & \textbf{0.111} \\
scan21 (9) & 19.80 & 11.41 & 21.71 & 21.49 & 21.81 & \textbf{22.79} & 0.853 & 0.416 & 0.883 & 0.891 & 0.894 & \textbf{0.899} & 0.110 & 0.499 & 0.108 & 0.081 & 0.088 & \textbf{0.077} \\
scan30 (3) & \textbf{19.86}$^\dagger$ & \textbf{19.86}$^\dagger$ & \textbf{19.86}$^\dagger$ & \textbf{19.86}$^\dagger$ & \textbf{19.86}$^\dagger$ & \textbf{19.86} & \textbf{0.933}$^\dagger$ & \textbf{0.933}$^\dagger$ & \textbf{0.933}$^\dagger$ & \textbf{0.933}$^\dagger$ & \textbf{0.933}$^\dagger$ & \textbf{0.933} & \textbf{0.052}$^\dagger$ & \textbf{0.052}$^\dagger$ & \textbf{0.052}$^\dagger$ & \textbf{0.052}$^\dagger$ & \textbf{0.052}$^\dagger$ & \textbf{0.052} \\
scan30 (6) & 23.52 & 15.73 & 26.72 & 25.99 & \textbf{27.82} & 26.80 & 0.943 & 0.891 & 0.959 & \textbf{0.964} & \textbf{0.964} & 0.963 & 0.045 & 0.106 & 0.051 & 0.035 & 0.042 & \textbf{0.029} \\
scan30 (9) & 24.09 & 8.45 & 29.85 & 29.41 & \textbf{30.27} & 28.20 & 0.956 & 0.842 & \textbf{0.977} & 0.976 & \textbf{0.977} & 0.969 & 0.038 & 0.164 & 0.030 & \textbf{0.022} & 0.024 & 0.024 \\
scan31 (3) & 10.66 & 8.08 & 14.34 & 15.38 & 14.17 & \textbf{17.81} & 0.620 & 0.490 & 0.702 & 0.745 & 0.732 & \textbf{0.816} & 0.317 & 0.462 & 0.242 & 0.196 & 0.218 & \textbf{0.128} \\
scan31 (6) & 19.65 & 14.42 & 20.52 & 20.83 & 21.05 & \textbf{21.74} & 0.822 & 0.653 & 0.867 & 0.869 & 0.875 & \textbf{0.878} & 0.135 & 0.308 & 0.108 & 0.093 & 0.098 & \textbf{0.085} \\
scan31 (9) & 19.52 & 15.27 & 22.70 & \textbf{23.54} & 22.54 & 23.36 & 0.873 & 0.680 & 0.908 & \textbf{0.913} & 0.911 & 0.903 & 0.094 & 0.300 & 0.080 & \textbf{0.063} & 0.072 & 0.069 \\
scan34 (3) & 12.35 & 8.17 & 16.16 & 17.81 & 17.17 & \textbf{21.35} & 0.552 & 0.428 & 0.707 & 0.771 & 0.765 & \textbf{0.808} & 0.339 & 0.484 & 0.252 & 0.179 & 0.205 & \textbf{0.132} \\
scan34 (6) & 21.32 & 13.96 & 22.56 & 21.84 & 21.24 & \textbf{24.25} & 0.807 & 0.481 & 0.841 & 0.835 & 0.832 & \textbf{0.860} & 0.133 & 0.382 & 0.157 & 0.127 & 0.158 & \textbf{0.098} \\
scan34 (9) & 23.57 & 15.81 & 25.51 & 24.44 & 25.23 & \textbf{25.66} & 0.875 & 0.525 & 0.900 & \textbf{0.903} & 0.897 & 0.890 & 0.087 & 0.339 & 0.115 & 0.079 & 0.108 & \textbf{0.077} \\
scan38 (3) & 10.98 & 7.36 & 15.26 & 15.03 & 16.37 & \textbf{18.60} & 0.449 & 0.347 & 0.625 & 0.674 & 0.683 & \textbf{0.775} & 0.433 & 0.617 & 0.339 & 0.264 & 0.279 & \textbf{0.168} \\
scan38 (6) & 21.17 & 16.08 & 22.34 & 21.02 & 21.93 & \textbf{22.98} & 0.785 & 0.544 & 0.839 & 0.831 & 0.838 & \textbf{0.851} & 0.161 & 0.384 & 0.184 & 0.139 & 0.158 & \textbf{0.109} \\
scan38 (9) & 23.03 & 14.90 & 24.51 & 24.37 & \textbf{25.22} & 24.28 & 0.866 & 0.463 & 0.891 & 0.895 & \textbf{0.896} & 0.888 & 0.101 & 0.432 & 0.135 & 0.089 & 0.111 & \textbf{0.083} \\
scan40 (3) & 8.36 & 8.36$^\dagger$ & 15.97 & 14.58 & 16.92 & \textbf{22.55} & 0.561 & 0.561$^\dagger$ & 0.635 & 0.641 & 0.651 & \textbf{0.840} & 0.374 & 0.374$^\dagger$ & 0.333 & 0.284 & 0.294 & \textbf{0.127} \\
scan40 (6) & 19.33 & 13.93 & 19.12 & 21.30 & 22.18 & \textbf{23.99} & 0.813 & 0.610 & 0.838 & 0.862 & 0.862 & \textbf{0.879} & 0.156 & 0.310 & 0.189 & 0.139 & 0.152 & \textbf{0.101} \\
scan40 (9) & 17.08 & 17.22 & 25.48 & 24.15 & 25.40 & \textbf{25.56} & 0.825 & 0.683 & 0.894 & 0.904 & \textbf{0.910} & 0.900 & 0.158 & 0.256 & 0.150 & 0.101 & 0.106 & \textbf{0.081} \\
scan41 (3) & 10.19 & 10.19$^\dagger$ & 15.61 & 18.21 & 17.18 & \textbf{19.75} & 0.587 & 0.587$^\dagger$ & 0.722 & 0.803 & 0.786 & \textbf{0.826} & 0.358 & 0.358$^\dagger$ & 0.275 & 0.169 & 0.207 & \textbf{0.130} \\
scan41 (6) & 21.98 & 16.03 & \textbf{24.77} & 24.42 & 24.11 & 24.58 & 0.820 & 0.712 & \textbf{0.883} & 0.880 & 0.876 & 0.882 & 0.142 & 0.309 & 0.137 & 0.105 & 0.129 & \textbf{0.086} \\
scan41 (9) & 23.76 & 19.08 & 26.81 & 26.29 & \textbf{26.99} & 26.51 & 0.892 & 0.766 & 0.927 & 0.926 & \textbf{0.931} & 0.910 & 0.093 & 0.238 & 0.099 & 0.072 & 0.084 & \textbf{0.067} \\
scan45 (3) & 8.42 & 6.25 & 14.45 & \textbf{16.82} & 16.65 & 15.39 & 0.641 & 0.452 & 0.767 & \textbf{0.831} & 0.825 & 0.819 & 0.306 & 0.466 & 0.176 & \textbf{0.125} & 0.135 & \textbf{0.125} \\
scan45 (6) & 20.39 & 12.59 & 21.14 & 20.79 & 19.77 & \textbf{21.53} & 0.878 & 0.603 & 0.897 & 0.896 & 0.890 & \textbf{0.903} & 0.090 & 0.355 & 0.080 & 0.076 & 0.084 & \textbf{0.068} \\
scan45 (9) & 22.67 & 14.11 & 23.14 & 22.53 & 22.98 & \textbf{23.29} & 0.920 & 0.686 & 0.933 & 0.927 & \textbf{0.935} & 0.924 & 0.057 & 0.283 & 0.056 & 0.054 & \textbf{0.052} & 0.053 \\
scan55 (3) & 18.20 & 8.77 & 18.16 & 20.15 & \textbf{21.14} & 20.46 & 0.832 & 0.634 & 0.844 & 0.888 & \textbf{0.897} & 0.874 & 0.134 & 0.265 & 0.140 & 0.098 & 0.101 & \textbf{0.090} \\
scan55 (6) & 23.09 & 14.61 & 24.84 & 25.17 & \textbf{25.18} & 24.96 & 0.936 & 0.717 & \textbf{0.954} & 0.950 & 0.953 & 0.952 & 0.060 & 0.188 & 0.053 & 0.048 & 0.049 & \textbf{0.042} \\
scan55 (9) & 25.95 & 20.25 & 27.84 & 27.33 & \textbf{27.92} & 27.07 & 0.957 & 0.865 & 0.969 & 0.967 & \textbf{0.970} & 0.964 & 0.039 & 0.127 & 0.041 & 0.034 & 0.036 & \textbf{0.032} \\
scan63 (3) & 8.34 & 8.34$^\dagger$ & 16.96 & 20.18 & 20.91 & \textbf{21.09} & 0.812 & 0.812$^\dagger$ & 0.896 & 0.936 & 0.939 & \textbf{0.945} & 0.178 & 0.178$^\dagger$ & 0.103 & 0.058 & 0.063 & \textbf{0.042} \\
scan63 (6) & 20.75 & 13.65 & 25.47 & 23.33 & 25.44 & \textbf{25.76} & 0.933 & 0.854 & 0.960 & 0.957 & 0.962 & \textbf{0.963} & 0.058 & 0.141 & 0.048 & 0.041 & 0.041 & \textbf{0.028} \\
scan63 (9) & 24.70 & 8.17 & \textbf{29.19} & 26.83 & 28.93 & 27.83 & 0.960 & 0.805 & 0.976 & 0.975 & \textbf{0.979} & 0.972 & 0.030 & 0.214 & 0.032 & 0.023 & 0.025 & \textbf{0.020} \\
scan82 (3) & 6.95 & 6.95$^\dagger$ & 15.19 & 13.91 & 10.86 & \textbf{20.35} & 0.785 & 0.785$^\dagger$ & 0.861 & 0.883 & 0.844 & \textbf{0.944} & 0.198 & 0.198$^\dagger$ & 0.124 & 0.101 & 0.137 & \textbf{0.047} \\
scan82 (6) & 22.10 & 9.52 & 25.69 & 21.10 & 25.67 & \textbf{26.03} & 0.943 & 0.834 & 0.955 & 0.949 & \textbf{0.964} & 0.963 & 0.045 & 0.178 & 0.059 & 0.046 & 0.043 & \textbf{0.028} \\
scan82 (9) & 21.22 & 8.00 & \textbf{27.37} & 24.48 & 27.06 & 26.65 & 0.949 & 0.820 & 0.972 & 0.970 & \textbf{0.975} & 0.969 & 0.040 & 0.206 & 0.040 & 0.028 & 0.031 & \textbf{0.022} \\
scan103 (3) & 19.37 & 19.08$^\dagger$ & 19.08 & 22.34 & 20.75 & \textbf{26.58} & 0.794 & 0.794$^\dagger$ & 0.836 & 0.899 & 0.897 & \textbf{0.928} & 0.196 & 0.196$^\dagger$ & 0.186 & 0.116 & 0.129 & \textbf{0.069} \\
scan103 (6) & 25.93 & 14.65 & 27.08 & 27.89 & 24.73 & \textbf{29.40} & 0.920 & 0.816 & 0.949 & 0.947 & 0.933 & \textbf{0.950} & 0.084 & 0.229 & 0.083 & 0.061 & 0.084 & \textbf{0.045} \\
scan103 (9) & 28.70 & 15.51 & 30.05 & 30.31 & \textbf{30.51} & 29.40 & 0.948 & 0.733 & 0.964 & 0.966 & \textbf{0.967} & 0.952 & 0.055 & 0.264 & 0.066 & \textbf{0.043} & 0.057 & \textbf{0.043} \\
scan110 (3) & \textbf{19.09}$^\dagger$ & \textbf{19.09}$^\dagger$ & \textbf{19.09}$^\dagger$ & \textbf{19.09}$^\dagger$ & \textbf{19.09}$^\dagger$ & \textbf{19.09} & \textbf{0.880}$^\dagger$ & \textbf{0.880}$^\dagger$ & \textbf{0.880}$^\dagger$ & \textbf{0.880}$^\dagger$ & \textbf{0.880}$^\dagger$ & \textbf{0.880} & \textbf{0.098}$^\dagger$ & \textbf{0.098}$^\dagger$ & \textbf{0.098}$^\dagger$ & \textbf{0.098}$^\dagger$ & \textbf{0.098}$^\dagger$ & \textbf{0.098} \\
scan110 (6) & 23.38 & 22.47$^\dagger$ & 23.90 & 23.77 & 22.47 & \textbf{24.22} & 0.929 & 0.929$^\dagger$ & 0.941 & 0.941 & 0.934 & \textbf{0.943} & 0.071 & 0.089$^\dagger$ & 0.089 & 0.070 & 0.089 & \textbf{0.051} \\
scan110 (9) & 25.46 & 16.87 & 26.44 & 26.68 & \textbf{26.97} & 26.19 & 0.953 & 0.847 & 0.964 & 0.966 & \textbf{0.968} & 0.959 & 0.051 & 0.164 & 0.066 & 0.043 & 0.055 & \textbf{0.037} \\
scan114 (3) & 20.87 & 14.59 & 19.26 & 22.91 & 23.22 & \textbf{24.04} & 0.854 & 0.720 & 0.847 & 0.918 & 0.913 & \textbf{0.921} & 0.123 & 0.228 & 0.134 & 0.072 & 0.084 & \textbf{0.066} \\
scan114 (6) & 26.06 & 20.37 & 26.06 & \textbf{26.68} & 26.55 & 26.15 & 0.931 & 0.837 & 0.943 & \textbf{0.946} & 0.944 & 0.940 & 0.060 & 0.155 & 0.063 & \textbf{0.047} & 0.055 & 0.053 \\
scan114 (9) & 25.93 & 22.86 & 27.16 & \textbf{27.87} & 26.77 & 26.87 & 0.942 & 0.894 & 0.957 & \textbf{0.959} & 0.957 & 0.946 & 0.051 & 0.111 & 0.053 & \textbf{0.035} & 0.045 & 0.049 \\
\textit{Mean} & 19.21 & 13.61 & 21.96 & 22.11 & 22.35 & 23.61 & 0.821 & 0.684 & 0.865 & 0.881 & 0.881 & 0.899 & 0.149 & 0.280 & 0.133 & 0.102 & 0.113 & 0.077 \\
\bottomrule
\end{tabular}%
}
\endgroup
\caption{\textbf{Per-scene results: sparse views (DTU).} Bold marks the best method in each row. $^\dagger$ marks failed runs, which take the worst score among the compared methods on that test case (Sec.~\ref{sec:exp:setup} in the main text); dataset means include these values and match Tab.~\ref{tab:main}. The number in parentheses after each scene is the number of training views. DNGS, CoR and Drop denote DNGaussian, CoR-GS and DropGaussian.}
\label{tab:supp:sparse-b}
\end{table*}

\begin{table*}[t]
\centering
\begingroup
\setlength{\tabcolsep}{2.25pt}
\renewcommand{\arraystretch}{0.96}
\resizebox{\textwidth}{!}{%
\begin{tabular}{@{}l@{\hspace{6pt}}cccccc@{\hspace{6pt}}cccccc@{\hspace{6pt}}cccccc@{}}
\toprule
 & \multicolumn{6}{c}{PSNR$\uparrow$} & \multicolumn{6}{c}{SSIM$\uparrow$} & \multicolumn{6}{c}{LPIPS$\downarrow$} \\
\cmidrule(lr){2-7}\cmidrule(lr){8-13}\cmidrule(lr){14-19}
Scene & \textcolor{gray}{COLMAP} & CF & HT & MASt3R & 3R-GS & Ours & \textcolor{gray}{COLMAP} & CF & HT & MASt3R & 3R-GS & Ours & \textcolor{gray}{COLMAP} & CF & HT & MASt3R & 3R-GS & Ours \\
\midrule
\multicolumn{19}{@{}l}{\textit{Tanks and Temples (8)}} \\
Ballroom & 32.16 & 17.30 & 17.33 & 20.81 & 23.28 & \textbf{32.76} & 0.956 & 0.433 & 0.434 & 0.622 & 0.734 & \textbf{0.962} & 0.038 & 0.214 & 0.212 & 0.118 & 0.078 & \textbf{0.034} \\
Barn & 29.44 & 18.84 & 18.47 & 22.86 & 24.49 & \textbf{30.46} & 0.925 & 0.510 & 0.501 & 0.704 & 0.732 & \textbf{0.940} & 0.076 & 0.248 & 0.225 & 0.119 & 0.097 & \textbf{0.054} \\
Church & 22.12 & 15.48 & 15.33 & 16.79 & 18.52 & \textbf{22.71} & 0.715 & 0.403 & 0.398 & 0.454 & 0.545 & \textbf{0.767} & 0.144 & 0.386 & 0.393 & 0.310 & 0.250 & \textbf{0.122} \\
Family & 28.45 & 15.40 & 16.19 & 19.77 & 20.14 & \textbf{27.42} & 0.925 & 0.451 & 0.444 & 0.584 & 0.590 & \textbf{0.888} & 0.113 & 0.335 & 0.316 & 0.198 & 0.178 & \textbf{0.091} \\
Francis & 31.51 & 16.44 & 16.91 & 17.43 & 23.14 & \textbf{32.00} & 0.914 & 0.492 & 0.505 & 0.516 & 0.679 & \textbf{0.921} & 0.151 & 0.442 & 0.425 & 0.402 & 0.227 & \textbf{0.134} \\
Horse & 24.81 & 16.41 & 16.26 & 15.78 & 19.21 & \textbf{23.59} & 0.881 & 0.564 & 0.560 & 0.555 & 0.640 & \textbf{0.869} & 0.153 & 0.270 & 0.282 & 0.311 & 0.197 & \textbf{0.127} \\
Ignatius & 30.00 & 17.54 & 17.49 & 20.34 & 20.57 & \textbf{25.87} & 0.928 & 0.365 & 0.362 & 0.546 & 0.545 & \textbf{0.900} & 0.079 & 0.292 & 0.288 & 0.190 & 0.187 & \textbf{0.098} \\
Museum & 30.49 & 13.94 & 13.92 & 27.15 & 27.62 & \textbf{28.26} & 0.918 & 0.322 & 0.322 & 0.848 & 0.841 & \textbf{0.873} & 0.054 & 0.472 & 0.474 & 0.064 & 0.068 & \textbf{0.058} \\
\textit{Mean} & 28.62 & 16.42 & 16.49 & 20.12 & 22.12 & 27.89 & 0.895 & 0.442 & 0.441 & 0.604 & 0.663 & 0.890 & 0.101 & 0.332 & 0.327 & 0.214 & 0.160 & 0.090 \\
\midrule
\multicolumn{19}{@{}l}{\textit{CO3D (5)}} \\
106\_12648\_23157 & 28.45 & 13.69 & 12.96 & 23.34 & 24.31 & \textbf{26.87} & 0.905 & 0.279 & 0.269 & 0.697 & 0.757 & \textbf{0.865} & 0.122 & 0.564 & 0.573 & 0.162 & 0.171 & \textbf{0.136} \\
110\_13051\_23361 & 33.14 & 13.91 & 15.58 & 26.09 & 26.52 & \textbf{31.40} & 0.945 & 0.424 & 0.450 & 0.744 & 0.719 & \textbf{0.942} & 0.105 & 0.597 & 0.517 & 0.157 & 0.160 & \textbf{0.105} \\
245\_26182\_52130 & 34.44 & 12.10 & 12.17 & 16.44 & 22.20 & \textbf{32.56} & 0.944 & 0.540 & 0.533 & 0.631 & 0.747 & \textbf{0.930} & 0.165 & 0.584 & 0.579 & 0.437 & 0.300 & \textbf{0.159} \\
34\_1403\_4393 & 33.33 & 15.99 & 17.56 & \textbf{31.01} & 27.12 & 31.00 & 0.944 & 0.586 & 0.594 & \textbf{0.897} & 0.738 & 0.896 & 0.133 & 0.450 & 0.406 & 0.141 & 0.181 & \textbf{0.139} \\
415\_57112\_110099 & 32.13 & 14.51 & 14.47 & 25.74 & \textbf{29.44} & 29.43 & 0.924 & 0.470 & 0.480 & 0.704 & 0.859 & \textbf{0.866} & 0.144 & 0.568 & 0.572 & 0.180 & \textbf{0.150} & 0.159 \\
\textit{Mean} & 32.30 & 14.04 & 14.55 & 24.52 & 25.92 & 30.25 & 0.932 & 0.460 & 0.465 & 0.734 & 0.764 & 0.900 & 0.134 & 0.553 & 0.530 & 0.215 & 0.192 & 0.140 \\
\midrule
\multicolumn{19}{@{}l}{\textit{Mip-NeRF 360 (4)}} \\
bicycle & 25.10 & 20.85$^\dagger$ & 20.85$^\dagger$ & 20.85 & 23.70 & \textbf{24.60} & 0.758 & 0.453$^\dagger$ & 0.453$^\dagger$ & 0.453 & 0.670 & \textbf{0.745} & 0.207 & 0.263$^\dagger$ & 0.263$^\dagger$ & 0.262 & 0.263 & \textbf{0.189} \\
counter & 29.13 & 9.75$^\dagger$ & 9.75 & 22.72 & 27.76 & \textbf{29.08} & 0.916 & 0.270$^\dagger$ & 0.270 & 0.703 & 0.876 & \textbf{0.914} & 0.179 & 0.682$^\dagger$ & 0.682 & 0.259 & 0.191 & \textbf{0.175} \\
garden & 27.57 & 18.70$^\dagger$ & 18.70$^\dagger$ & 18.70 & 19.86 & \textbf{27.32} & 0.866 & 0.299$^\dagger$ & 0.299$^\dagger$ & 0.299 & 0.367 & \textbf{0.863} & 0.106 & 0.292$^\dagger$ & 0.292$^\dagger$ & 0.292 & 0.267 & \textbf{0.105} \\
room & 31.48 & 8.62$^\dagger$ & 8.62 & 11.13 & 11.40 & \textbf{30.92} & 0.923 & 0.061$^\dagger$ & 0.061 & 0.463 & 0.508 & \textbf{0.915} & 0.196 & 0.638$^\dagger$ & 0.638 & 0.609 & 0.603 & \textbf{0.194} \\
\textit{Mean} & 28.32 & 14.48 & 14.48 & 18.35 & 20.68 & 27.98 & 0.866 & 0.271 & 0.271 & 0.480 & 0.605 & 0.859 & 0.172 & 0.469 & 0.469 & 0.355 & 0.331 & 0.166 \\
\midrule
\multicolumn{19}{@{}l}{\textit{DTU (4)}} \\
scan106 & 34.37 & 15.05 & 14.44 & 19.79 & 21.42 & \textbf{33.50} & 0.951 & 0.582 & 0.483 & 0.724 & 0.747 & \textbf{0.943} & 0.062 & 0.341 & 0.380 & 0.207 & 0.179 & \textbf{0.056} \\
scan110 & 31.93 & 17.71 & 17.06 & 28.59 & 30.81 & \textbf{33.02} & 0.948 & 0.689 & 0.613 & 0.916 & 0.942 & \textbf{0.961} & 0.077 & 0.276 & 0.319 & 0.068 & 0.055 & \textbf{0.054} \\
scan69 & 28.00 & 16.83 & 16.71 & 25.49 & 26.32 & \textbf{27.83} & 0.931 & 0.508 & 0.456 & 0.861 & 0.877 & \textbf{0.925} & 0.085 & 0.415 & 0.428 & 0.103 & 0.097 & \textbf{0.080} \\
scan83 & 32.66 & 17.70 & 19.25 & 26.59 & 25.01 & \textbf{34.22} & 0.942 & 0.733 & 0.754 & 0.890 & 0.881 & \textbf{0.950} & 0.046 & 0.227 & 0.224 & 0.091 & 0.113 & \textbf{0.041} \\
\textit{Mean} & 31.74 & 16.83 & 16.87 & 25.12 & 25.89 & 32.14 & 0.943 & 0.628 & 0.577 & 0.848 & 0.862 & 0.945 & 0.067 & 0.315 & 0.338 & 0.117 & 0.111 & 0.058 \\
\bottomrule
\end{tabular}%
}
\endgroup
\caption{\textbf{Per-scene results: unknown camera poses (pose-free).} Bold marks the best method in each row (excluding the gray reference). $^\dagger$ marks failed runs, which take the worst score among the compared methods on that test case (Sec.~\ref{sec:exp:setup} in the main text); dataset means include these values and match Tab.~\ref{tab:main}. CF, HT and MASt3R denote CF-3DGS, HT-3DGS and MASt3R+3DGS. COLMAP uses COLMAP poses and is a reference, not a competitor.}
\label{tab:supp:posefree}
\end{table*}

\begin{table*}[t]
\centering
\begingroup
\setlength{\tabcolsep}{2.25pt}
\renewcommand{\arraystretch}{0.96}
\resizebox{\textwidth}{!}{%
\begin{tabular}{@{}l@{\hspace{6pt}}cccccc@{\hspace{6pt}}cccccc@{\hspace{6pt}}cccccc@{}}
\toprule
 & \multicolumn{6}{c}{PSNR$\uparrow$} & \multicolumn{6}{c}{SSIM$\uparrow$} & \multicolumn{6}{c}{LPIPS$\downarrow$} \\
\cmidrule(lr){2-7}\cmidrule(lr){8-13}\cmidrule(lr){14-19}
Scene & 3DGS & SpotLess & WildGS & DeSplat & RobustS & Ours & 3DGS & SpotLess & WildGS & DeSplat & RobustS & Ours & 3DGS & SpotLess & WildGS & DeSplat & RobustS & Ours \\
\midrule
\multicolumn{19}{@{}l}{\textit{RobustNeRF (5)}} \\
android & 24.09 & 24.67 & 25.27 & 24.98 & \textbf{25.49} & 25.11 & 0.825 & 0.843 & 0.850 & 0.851 & \textbf{0.865} & 0.849 & 0.155 & 0.145 & 0.146 & 0.130 & \textbf{0.110} & 0.120 \\
crab1 & 30.95 & 33.10 & 31.56 & 36.28 & \textbf{36.81} & 34.99 & 0.954 & 0.958 & 0.950 & 0.968 & 0.969 & \textbf{0.970} & 0.115 & 0.122 & 0.133 & 0.079 & 0.104 & \textbf{0.062} \\
crab2 & 30.17 & 33.90 & 30.00 & \textbf{35.03} & 34.94 & 34.04 & 0.915 & 0.949 & 0.920 & 0.954 & \textbf{0.956} & 0.954 & 0.175 & 0.143 & 0.182 & 0.103 & 0.134 & \textbf{0.092} \\
statue & 20.88 & 22.83 & 22.86 & \textbf{23.51} & 23.44 & 22.46 & 0.820 & 0.846 & 0.862 & 0.875 & \textbf{0.880} & 0.858 & 0.187 & 0.175 & 0.149 & 0.123 & \textbf{0.117} & 0.133 \\
yoda & 29.77 & 34.01 & 31.31 & 35.38 & \textbf{35.58} & 34.09 & 0.912 & 0.944 & 0.932 & 0.957 & \textbf{0.960} & 0.955 & 0.173 & 0.152 & 0.168 & 0.095 & 0.121 & \textbf{0.080} \\
\textit{Mean} & 27.17 & 29.70 & 28.20 & 31.03 & 31.25 & 30.14 & 0.885 & 0.908 & 0.903 & 0.921 & 0.926 & 0.917 & 0.161 & 0.148 & 0.156 & 0.106 & 0.117 & 0.097 \\
\midrule
\multicolumn{19}{@{}l}{\textit{On-the-go (6)}} \\
corner & 21.05 & 24.29 & 24.65 & 25.44 & \textbf{25.69} & 24.95 & 0.739 & 0.831 & 0.843 & 0.858 & \textbf{0.871} & 0.861 & 0.273 & 0.214 & 0.190 & 0.151 & 0.158 & \textbf{0.136} \\
fountain & 20.36 & 21.71 & 21.27 & 21.36 & \textbf{21.83} & 21.46 & 0.681 & 0.683 & 0.693 & 0.713 & \textbf{0.729} & 0.693 & 0.234 & 0.269 & 0.247 & 0.211 & \textbf{0.200} & 0.218 \\
mountain & 18.81 & 21.30 & 21.06 & 20.94 & \textbf{21.35} & 21.33 & 0.660 & 0.691 & 0.707 & 0.714 & \textbf{0.723} & 0.703 & 0.270 & 0.284 & 0.278 & \textbf{0.216} & 0.226 & 0.220 \\
patio & 16.78 & \textbf{22.19} & 21.52 & 18.62 & 22.08 & 21.25 & 0.692 & 0.819 & 0.816 & 0.782 & \textbf{0.840} & 0.826 & 0.261 & 0.171 & 0.171 & 0.186 & 0.140 & \textbf{0.134} \\
patio\_high & 16.57 & 21.08 & 22.09 & 21.96 & 21.89 & \textbf{22.32} & 0.583 & 0.732 & 0.763 & 0.789 & 0.793 & \textbf{0.794} & 0.387 & 0.268 & 0.223 & 0.189 & 0.190 & \textbf{0.177} \\
spot & 16.52 & 21.57 & 23.97 & \textbf{25.60} & 25.19 & 24.91 & 0.608 & 0.695 & 0.805 & 0.844 & \textbf{0.853} & 0.846 & 0.472 & 0.398 & 0.218 & 0.165 & 0.161 & \textbf{0.140} \\
\textit{Mean} & 18.35 & 22.02 & 22.43 & 22.32 & 23.01 & 22.70 & 0.661 & 0.742 & 0.771 & 0.783 & 0.801 & 0.787 & 0.316 & 0.267 & 0.221 & 0.186 & 0.179 & 0.171 \\
\midrule
\multicolumn{19}{@{}l}{\textit{RealX3D (8)}} \\
Chocolate & 20.35 & 22.42 & 21.40 & \textbf{22.43} & 22.36 & 22.05 & 0.639 & 0.674 & 0.664 & 0.670 & \textbf{0.675} & 0.657 & 0.261 & 0.281 & 0.291 & 0.231 & 0.241 & \textbf{0.218} \\
Cupcake & 21.77 & 30.28 & 27.45 & \textbf{30.44} & 29.30 & 30.09 & 0.889 & 0.934 & 0.924 & 0.944 & 0.941 & \textbf{0.945} & 0.156 & 0.127 & 0.133 & 0.095 & 0.108 & \textbf{0.075} \\
GearWorks & 17.09 & 17.51 & 19.73 & \textbf{20.01} & 17.66 & 19.42 & 0.758 & 0.770 & 0.814 & \textbf{0.828} & 0.800 & 0.804 & 0.277 & 0.322 & 0.258 & 0.218 & 0.261 & \textbf{0.208} \\
Laboratory & 16.86 & 27.98 & 24.76 & 28.96 & \textbf{29.16} & 27.81 & 0.755 & 0.885 & 0.870 & \textbf{0.908} & 0.901 & 0.899 & 0.274 & 0.182 & 0.181 & 0.124 & 0.149 & \textbf{0.117} \\
Limon & 14.68 & 19.88 & 20.88 & \textbf{22.18} & 21.39 & 21.55 & 0.573 & 0.672 & 0.692 & \textbf{0.708} & 0.695 & 0.669 & 0.431 & 0.360 & 0.291 & \textbf{0.247} & 0.270 & 0.249 \\
MilkCookie & 24.71 & 24.32 & 22.16 & 23.17 & \textbf{26.91} & 23.35 & 0.903 & 0.894 & 0.885 & 0.907 & \textbf{0.933} & 0.904 & 0.147 & 0.210 & 0.184 & 0.141 & \textbf{0.124} & 0.133 \\
Popcorn & 25.48 & 27.91 & 26.54 & 28.00 & \textbf{28.52} & 27.68 & 0.870 & 0.879 & 0.866 & 0.893 & 0.896 & \textbf{0.897} & 0.144 & 0.168 & 0.181 & 0.128 & 0.142 & \textbf{0.095} \\
Ujikintoki & 23.50 & 31.72 & 26.81 & 31.28 & \textbf{31.89} & 30.48 & 0.850 & 0.893 & 0.876 & 0.908 & 0.904 & \textbf{0.911} & 0.259 & 0.267 & 0.282 & 0.195 & 0.235 & \textbf{0.129} \\
\textit{Mean} & 20.55 & 25.25 & 23.72 & 25.81 & 25.90 & 25.30 & 0.780 & 0.825 & 0.824 & 0.846 & 0.843 & 0.836 & 0.244 & 0.240 & 0.225 & 0.173 & 0.191 & 0.153 \\
\bottomrule
\end{tabular}%
}
\endgroup
\caption{\textbf{Per-scene results: temporary occlusions (distractors).} Bold marks the best method in each row. $^\dagger$ marks failed runs, which take the worst score among the compared methods on that test case (Sec.~\ref{sec:exp:setup} in the main text); dataset means include these values and match Tab.~\ref{tab:main}. WildGS and RobustS denote WildGaussians and RobustSplat.}
\label{tab:supp:transient}
\end{table*}

\begin{table*}[t]
\centering
\begingroup
\setlength{\tabcolsep}{2.25pt}
\renewcommand{\arraystretch}{0.96}
\resizebox{\textwidth}{!}{%
\begin{tabular}{@{}l@{\hspace{6pt}}cccccc@{\hspace{6pt}}cccccc@{\hspace{6pt}}cccccc@{}}
\toprule
 & \multicolumn{6}{c}{PSNR$\uparrow$} & \multicolumn{6}{c}{SSIM$\uparrow$} & \multicolumn{6}{c}{LPIPS$\downarrow$} \\
\cmidrule(lr){2-7}\cmidrule(lr){8-13}\cmidrule(lr){14-19}
Scene & 3DGS & Lum-GS & LITA & LL-GS & Aleth & Ours & 3DGS & Lum-GS & LITA & LL-GS & Aleth & Ours & 3DGS & Lum-GS & LITA & LL-GS & Aleth & Ours \\
\midrule
\multicolumn{19}{@{}l}{\textit{LOM low light (5)}} \\
bike & 6.39 & 17.70 & 10.96 & 19.45 & -- & \textbf{27.30} & 0.077 & 0.775 & 0.561 & 0.801 & -- & \textbf{0.861} & 0.533 & 0.365 & 0.629 & 0.330 & -- & \textbf{0.222} \\
buu & 7.51 & 21.96 & 20.61 & 18.04 & -- & \textbf{22.84} & 0.294 & \textbf{0.925} & 0.901 & 0.844 & -- & 0.913 & 0.406 & \textbf{0.160} & 0.203 & 0.188 & -- & 0.169 \\
chair & 6.06 & 18.52 & 20.95 & 18.40 & -- & \textbf{24.24} & 0.150 & 0.858 & 0.860 & 0.834 & -- & \textbf{0.906} & 0.624 & 0.319 & 0.295 & 0.325 & -- & \textbf{0.253} \\
shrub & 8.18 & 14.32 & 18.33 & 18.28 & -- & \textbf{20.77} & 0.046 & 0.671 & 0.650 & 0.675 & -- & \textbf{0.754} & 0.577 & 0.284 & 0.383 & 0.286 & -- & \textbf{0.233} \\
sofa & 6.29 & 23.22 & 18.39 & 17.34 & -- & \textbf{23.74} & 0.208 & 0.903 & 0.862 & 0.839 & -- & \textbf{0.923} & 0.566 & 0.292 & 0.318 & 0.357 & -- & \textbf{0.249} \\
\textit{Mean} & 6.88 & 19.14 & 17.85 & 18.30 & -- & 23.78 & 0.155 & 0.826 & 0.767 & 0.798 & -- & 0.871 & 0.541 & 0.284 & 0.366 & 0.297 & -- & 0.225 \\
\midrule
\multicolumn{19}{@{}l}{\textit{RealX3D low light (9)}} \\
BlueHawaii & 7.49 & 12.15 & -- & 17.35 & -- & \textbf{18.00} & 0.059 & 0.589 & -- & 0.554 & -- & \textbf{0.723} & 0.708 & 0.516 & -- & 0.544 & -- & \textbf{0.402} \\
Chocolate & 7.89 & 13.36 & -- & 16.52 & -- & \textbf{17.16} & 0.091 & 0.252 & -- & 0.405 & -- & \textbf{0.537} & 0.667 & 0.577 & -- & 0.602 & -- & \textbf{0.448} \\
Cupcake & 4.43 & \textbf{19.00} & -- & 11.62 & -- & 15.34 & 0.069 & \textbf{0.799} & -- & 0.563 & -- & 0.732 & 0.667 & 0.284 & -- & 0.327 & -- & \textbf{0.212} \\
GearWorks & 6.90 & 8.09 & -- & 10.15 & -- & \textbf{16.34} & 0.075 & 0.495 & -- & 0.279 & -- & \textbf{0.688} & 0.636 & 0.554 & -- & 0.539 & -- & \textbf{0.407} \\
Laboratory & 7.61 & 10.37 & -- & 15.73 & -- & \textbf{17.99} & 0.044 & 0.150 & -- & 0.430 & -- & \textbf{0.746} & 0.548 & 0.513 & -- & 0.507 & -- & \textbf{0.353} \\
MilkCookie & 6.20 & 16.45 & -- & 10.37 & -- & \textbf{16.62} & 0.056 & 0.619 & -- & 0.394 & -- & \textbf{0.718} & 0.606 & 0.501 & -- & 0.511 & -- & \textbf{0.352} \\
Popcorn & 7.47 & 13.73 & -- & 14.36 & -- & \textbf{18.05} & 0.064 & 0.230 & -- & 0.399 & -- & \textbf{0.646} & 0.715 & 0.440 & -- & 0.559 & -- & \textbf{0.345} \\
Sculpture & 6.06 & 12.86 & -- & 11.40 & -- & \textbf{16.69} & 0.027 & 0.337 & -- & 0.213 & -- & \textbf{0.540} & 0.722 & 0.528 & -- & 0.650 & -- & \textbf{0.414} \\
Ujikintoki & 6.24 & 10.77 & -- & 17.44 & -- & \textbf{19.78} & 0.065 & 0.615 & -- & 0.583 & -- & \textbf{0.711} & 0.810 & 0.564 & -- & 0.436 & -- & \textbf{0.389} \\
\textit{Mean} & 6.70 & 12.98 & -- & 13.88 & -- & 17.33 & 0.061 & 0.454 & -- & 0.424 & -- & 0.671 & 0.676 & 0.498 & -- & 0.519 & -- & 0.369 \\
\midrule
\multicolumn{19}{@{}l}{\textit{LOM over-exposure (3)}} \\
bike & 9.63 & 24.26 & -- & -- & 21.26 & \textbf{26.58} & 0.736 & 0.872 & -- & -- & 0.836 & \textbf{0.901} & 0.265 & 0.142 & -- & -- & 0.291 & \textbf{0.093} \\
chair & 11.12 & 21.90 & -- & -- & 20.05 & \textbf{23.35} & 0.803 & 0.851 & -- & -- & 0.884 & \textbf{0.897} & 0.256 & 0.180 & -- & -- & 0.304 & \textbf{0.146} \\
shrub & 10.34 & 16.37 & -- & -- & 17.06 & \textbf{21.99} & 0.660 & 0.806 & -- & -- & 0.644 & \textbf{0.867} & 0.205 & 0.151 & -- & -- & 0.338 & \textbf{0.125} \\
\textit{Mean} & 10.36 & 20.85 & -- & -- & 19.46 & 23.97 & 0.733 & 0.843 & -- & -- & 0.788 & 0.888 & 0.242 & 0.158 & -- & -- & 0.311 & 0.121 \\
\bottomrule
\end{tabular}%
}
\endgroup
\caption{\textbf{Per-scene results: poor exposure.} Bold marks the best method in each row. $^\dagger$ marks failed runs, which take the worst score among the compared methods on that test case (Sec.~\ref{sec:exp:setup} in the main text); dataset means include these values and match Tab.~\ref{tab:main}. Lum-GS, LITA and Aleth denote Luminance-GS, LITA-GS and Aleth-NeRF. LITA-GS is evaluated on LOM only; Aleth-NeRF on over-exposure only. Ours uses DarkIR for low light and LCDPNet for over-exposure.}
\label{tab:supp:exposure}
\end{table*}

\section{Matched-Input Comparison Details}
\label{sec:supp:matched}

\paragraph{Pair construction.}
A comparison pair is one specially-designed method on one test case.
A pair enters the comparison when all four runs can be evaluated: the specialized method and standard 3DGS, each on native and recovered inputs.
We focus on representative, competitive baselines covering different task-specific training designs.
Specifically, we evaluate FSGS, CoR-GS, and DropGaussian for sparse views, the strongest pose-free baseline 3R-GS, and the transient and exposure methods listed in \cref{tab:supp:matched}.
On recovered inputs, each method receives exactly the recovered quantity of its task and otherwise runs its official pipeline:
the VGGT-$\Omega$ point cloud replaces its native initialization (sparse views);
the BA-refined cameras and point cloud replace its native pose and point estimates (pose-free);
the recovered static masks are multiplied into its loss, while its own masking, uncertainty, or transient modules stay active (temporary occlusions);
and the restored, exposure-normalized images replace its training images, while its exposure modeling stays active (poor exposure).
For temporary occlusions, both sides keep the same training-only SfM initialization so that only the masks differ.
For pose-free pairs, the native comparison is 3R-GS versus standard 3DGS, both on MASt3R cameras and points. In the sparse-view benchmark, dense initialization also enabled 19 baseline runs that failed to initialize on native inputs. Those recoveries are reported separately from the 261 pairs used to compare PSNR gains.

\begin{table}[t]
    \centering
    \footnotesize
    \setlength{\tabcolsep}{4pt}
    \begin{tabular}{@{}llrrrr@{}}
        \toprule
        Setting & Method / subset & Pairs & Native & Recovered & Shrinks \\
        \midrule
        Sparse & FSGS & 87 & +2.27 & +0.33 & 71 \\
               & CoR-GS & 87 & +2.60 & +0.81 & 75 \\
               & DropGaussian & 87 & +2.84 & +1.19 & 68 \\
        \cmidrule(l){2-6}
               & LLFF & 72 & +3.42 & +2.41 & 48 \\
               & DTU & 129 & +3.06 & +0.35 & 119 \\
               & Mip-NeRF 360 & 42 & +1.41 & +0.62 & 37 \\
               & Blender & 18 & $-$1.71 & $-$2.29 & 10 \\
        \midrule
        Pose-free & 3R-GS & 21 & +1.69 & $-$0.89 & 18 \\
        \midrule
        Transient & SpotLess & 19 & +3.80 & +0.36 & 17 \\
                  & WildGaussians & 19 & +2.89 & +0.19 & 18 \\
                  & DeSplat & 19 & +4.48 & +0.79 & 18 \\
                  & RobustSplat & 19 & +4.79 & +1.12 & 17 \\
        \midrule
        Exposure & Luminance-GS & 14 & +8.86 & $-$2.20 & 14 \\
                 & LITA-GS (LOM) & 5 & +10.96 & $-$0.55 & 5 \\
                 & LL-Gaussian & 14 & +8.70 & $-$0.70 & 14 \\
                 & Aleth-NeRF & 3 & +9.10 & $-$2.11 & 3 \\
        \bottomrule
    \end{tabular}
    \caption{\textbf{Breakdown of Tab.~\ref{tab:shared}.} Mean PSNR gain (dB) of each specially-designed method over standard 3DGS on the same inputs, before (Native) and after (Recovered) input recovery. \emph{Shrinks} counts pairs with $\Delta g<0$.}
    \label{tab:supp:matched}
\end{table}

\input{sec/supp/supp-fig-stage2}
\paragraph{Breakdown.}
\cref{fig:supp:stage2} plots every pair, and \cref{tab:supp:matched} splits Tab.~\ref{tab:shared} by method and, for sparse views, by dataset.
The mean gain shrinks for every method listed.
For sparse views, the remaining gain is concentrated on forward-facing LLFF, while on DTU it falls from $+3.06$ to $+0.35$\,dB.
For pose-free reconstruction, the recovered-input gain of 3R-GS is negative on 13 of the 21 scenes.
For exposure, every method has a negative mean gain on restored images.

We retain the official implementations and hyperparameters when switching from native to recovered inputs, since our goal is to examine how existing training components behave as input quality improves.
Sensitivity to this change is itself part of the behavior under study.

\section{Controlled Component Probes}
\label{sec:supp:analysis}

In the shared gsplat trainer, each Stage~3 comparison in Sec.~\ref{sec:analysis} in the main text varies the designated component or input condition while holding the remaining settings fixed. Additional checks use the official 3R-GS, Luminance-GS, and RobustSplat implementations, as specified below.
Most probes train for 7k or 10k steps, since the effects they study appear early in training (Sec.~\ref{sec:analysis_degrade} in the main text); the step count is given for each probe below.
Seeds and scenes are fixed before any test metric is computed.

\subsection{Camera Pose Optimization}
\label{sec:supp:pose}

\paragraph{Setup.}
We use six pose-free scenes with BA-refined cameras (DTU scan69 and scan106, CO3D 110 and 245, Tanks and Temples Barn and Ignatius) and the recovered point cloud of Ours.
Each camera receives a 6-DoF pose offset applied on the left in the camera frame, optimized jointly with the Gaussians by Adam with learning rate $10^{-4}$ (the same order as 3R-GS), decayed exponentially to 1\% over 7k steps.
Test cameras are aligned with the training-only Sim(3) protocol of \cref{sec:supp:protocol}.

\paragraph{Controlled error injection (Fig.~\ref{fig:fig5_crossover}a in the main text).}
Starting from the BA cameras, we rotate every training camera about its center by a fixed angle in a random direction (0, 0.02, 0.05, 0.1, 0.2, or $0.5^\circ$; the same perturbation per scene for both settings).
Each angle is trained once with fixed cameras and twice with pose optimization.
\cref{tab:supp:dose_t2} lists the results.
The gain from pose optimization rises monotonically with the initial error and crosses zero between $0.05^\circ$ and $0.1^\circ$, just above the error left by BA ($0.052^\circ$).
After training, the rotation error ends near $0.065^\circ$ regardless of where it started, which explains the crossover: pose optimization corrects cameras whose error is above this level and adds error to cameras below it.

\begin{table}[t]
    \centering
    \footnotesize
    \setlength{\tabcolsep}{5pt}
    \begin{tabular}{@{}cccc@{}}
        \toprule
        Injected ($^\circ$) & Error before ($^\circ$) & Error after ($^\circ$) & $\Delta$PSNR (dB) \\
        \midrule
        0    & 0.052 & 0.067 & $-$0.74 \\
        0.02 & 0.059 & 0.067 & $-$0.63 \\
        0.05 & 0.075 & 0.063 & $-$0.01 \\
        0.1  & 0.117 & 0.065 & +1.23 \\
        0.2  & 0.207 & 0.062 & +3.15 \\
        0.5  & 0.497 & 0.074 & +5.87 \\
        \bottomrule
    \end{tabular}
    \caption{\textbf{Pose optimization versus initial camera error} (six scenes, 7k steps). $\Delta$PSNR is pose optimization minus fixed cameras. Errors are mean training-camera rotation errors before and after training.}
    \label{tab:supp:dose_t2}
\end{table}

\paragraph{Gradient direction and step size.}
In a diagnostic run that only adds logging, we record for every step in several windows (steps 0--99, 500--549, 1000--1049, 3000--3049, 6950--6999) the negative pose gradient of the sampled camera, the update actually applied by Adam to all cameras, and the per-camera error before and after the step.
Errors and correction directions are measured in a frame fixed at the start of training, so that the alignment does not change between steps.
On BA cameras, the negative gradient of the sampled camera has a mean cosine of $+0.28$ with the true correction direction in the first 100 steps, and about six in ten sampled updates reduce that camera's error.
At the same time, Adam momentum keeps moving the cameras that are not sampled in the current step; 81\% of these moves point away from the correction, and they dominate the net error change.
The mean BA error rises from $0.052^\circ$ to $0.081^\circ$ by step 500.
After about 3k steps the cosine is near zero for every error level and learning rate, and the fraction of improving steps is 43--56\%, close to chance.
\cref{fig:supp:cosine} shows these cosines for all injected error levels.
With learning rates $3\times10^{-5}$, $10^{-4}$, and $3\times10^{-4}$, the zero-gain crossover moves to $0.059^\circ$, $0.072^\circ$, and $0.101^\circ$ (Fig.~\ref{fig:fig5_crossover}c in the main text).

\paragraph{Delayed pose optimization.}
Starting pose optimization at step 3k instead of step 0 reduces the loss on BA cameras from $-0.74$ to $-0.17$\,dB, but also reduces the gain on raw VGGT-$\Omega$ cameras from $+2.38$ to $+0.66$\,dB, because the Gaussians have already adapted to the initial cameras.

\paragraph{Optimizer.}
The reversal is not specific to Adam.
With SGD at learning rates $10^{-3}$ and $3\times10^{-3}$, pose optimization gives $+2.29$ and $+1.26$\,dB on raw cameras, and $-2.62$ and $-4.56$\,dB on BA cameras (Adam: $+2.38$ and $-0.76$\,dB).
Updating only the sampled camera (selective Adam) reduces the loss on BA cameras but also removes most of the correction on raw cameras.

\paragraph{Official 3R-GS.}
Comparing the training poses written by the official 3R-GS with its BA input poses, 3R-GS increases the rotation error in 14 of 21 scenes (\eg, Ignatius $0.012^\circ\to0.157^\circ$, Horse $0.054^\circ\to0.452^\circ$).
On these 14 scenes, 3R-GS is on average 1.39\,dB below Ours; on the other 7 scenes, where its poses do not get worse, the two differ by only 0.11\,dB on average.

\input{sec/supp/supp-fig-cosine}
\subsection{Learnable Image Targets}
\label{sec:supp:target}

\paragraph{Setup.}
We use LOM \texttt{bike}, \texttt{chair}, and \texttt{sofa}.
To vary exposure quality without adding new artifacts, we multiply the restored training images by a scale $k$ and train for 10k steps.
The learnable target follows Luminance-GS: each training view has a monotone 16-knot tone curve $\Phi_\theta$ applied to the training image, and a brightness prior pulls the mean intensity of $\Phi_\theta(I)$ (averaged over pixels and RGB channels) towards 0.45.
The loss is $\mathcal{L}_{\mathrm{photo}}(\mathcal{R}(G),\Phi_\theta(I)) + w\,(\operatorname{mean}(\Phi_\theta(I))-0.45)^2$.
The fixed target uses $\Phi_\theta(I)=I$.
We report PSNR and color-calibrated PSNR (cc-PSNR), where each rendering is first mapped to the ground truth by the best per-image affine color transform. cc-PSNR thus discounts global color and exposure differences, but still counts all other appearance errors.

\begin{table}[t]
    \centering
    \footnotesize
    \setlength{\tabcolsep}{4.5pt}
    \begin{tabular}{@{}ccccc@{}}
        \toprule
        $k$ & Mean intensity & Fixed & Learnable & Gain \\
        \midrule
        0.1  & 0.049 & 6.19  & 20.02 & +13.84 \\
        0.2  & 0.097 & 7.15  & 18.53 & +11.37 \\
        0.35 & 0.170 & 8.83  & 18.05 & +9.21 \\
        0.5  & 0.243 & 10.89 & 17.75 & +6.86 \\
        0.7  & 0.341 & 14.65 & 17.54 & +2.89 \\
        0.85 & 0.414 & 18.97 & 17.50 & $-$1.46 \\
        1.0  & 0.487 & 25.04 & 17.50 & $-$7.53 \\
        1.15 & 0.558 & 23.66 & 17.50 & $-$6.16 \\
        1.3  & 0.628 & 17.69 & 17.45 & $-$0.24 \\
        \bottomrule
    \end{tabular}
    \caption{\textbf{Learnable target versus input exposure} (LOM \texttt{bike}, \texttt{chair}, \texttt{sofa}; 10k steps; PSNR in dB). $k=1$ is the restored input used by Ours.}
    \label{tab:supp:dose_t4}
\end{table}

\paragraph{Exposure sweep (Fig.~\ref{fig:fig5_crossover}b in the main text).}
\cref{tab:supp:dose_t4} shows that the learnable target gives nearly the same PSNR (17.5--20\,dB) at every exposure, while the fixed target follows the input quality.
The gain is large on dark inputs and negative near the restored exposure.

\paragraph{Contrast collapse (Fig.~\ref{fig:t4_moving_target} in the main text).}
To separate the roles of the photometric loss and the brightness prior, we run a $2\times2$ design: the photometric loss either updates $\theta$ or sees a detached target $\Phi_\theta(I)$, and the prior is either on or off.
The Gaussians always receive the full photometric gradient.
We measure target contrast as the standard deviation of the target luminance divided by that of the input, on eight fixed training views.
On restored inputs ($k=1$), whenever the photometric loss can update $\theta$, the target contrast decreases throughout training; on \texttt{bike} it falls to 14\% of the input after 10k steps with the full learnable target.
With the prior alone, the \texttt{bike} target keeps 92\% of its contrast at $k=1$.
On dark inputs the ratio is above 1 for every variant that uses the prior, because the prior brightens the target.
\cref{tab:supp:collapse} gives cc-PSNR for all four variants: cutting the photometric gradient to $\theta$ brings cc-PSNR close to the fixed-target level at all three exposures (see \cref{fig:supp:collapse} for the targets on all three scenes), while both variants in which the photometric loss updates $\theta$ lose 4--5\,dB on restored inputs.

\begin{table}[t]
    \centering
    \footnotesize
    \setlength{\tabcolsep}{4pt}
    \begin{tabular}{@{}ccccc@{}}
        \toprule
        & & Photo + prior & Photo only & Prior only \\
        $k$ & Fixed & (full) & (no prior) & (photo detached) \\
        \midrule
        0.1 & 29.48 & 29.99 & 20.96 & 30.46 \\
        0.5 & 30.52 & 25.81 & 26.74 & 30.47 \\
        1.0 & 30.51 & 25.93 & 26.59 & 30.40 \\
        \bottomrule
    \end{tabular}
    \caption{\textbf{cc-PSNR (dB) of the four target variants} (mean over LOM \texttt{bike}, \texttt{chair}, \texttt{sofa}; 10k steps). $k=0.1$ and $1.0$ average two seeds; $k=0.5$ uses one seed. These values are plotted in Fig.~\ref{fig:t4_moving_target}a in the main text.}
    \label{tab:supp:collapse}
\end{table}

\paragraph{Official Luminance-GS.}
We ran the official Luminance-GS on restored LOM images with its learnable target on (native) or replaced by the fixed restored image, three seeds each.
Across the runs that converge, enabling the learnable target loses 2.2--3.6\,dB PSNR: 2.2--2.5\,dB on \texttt{sofa} over three seeds and 3.6\,dB on \texttt{chair}.
On \texttt{sofa}, about 2.0\,dB of this loss remains after affine color calibration, so it is not only a global color shift.
The other two \texttt{chair} seeds diverge with the learnable target.
On the raw dark images, the same learnable target is 12--17\,dB better than a fixed target, of which about 3\,dB remains after color calibration.

\paragraph{Per-view affine color maps.}
A related appearance component applies a per-view $3\times3$ color matrix and bias to the rendering before the loss, as in exposure-compensation methods.
On restored LOM images (10k steps), free per-view maps lower PSNR by 0.8--1.9\,dB while cc-PSNR is unchanged, because the mean of the maps over views is not constrained and drifts during training.
Fixing this mean to the identity at every step removes the loss.

\begin{figure*}[t]
\centering
\iflatexml
\includegraphics[width=\linewidth,alt={Learned supervision targets for Bike, Chair, and Sofa using full updates or only the brightness prior.}]{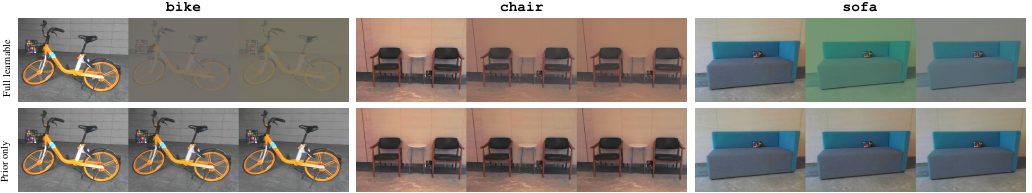}
\else
\setlength{\tabcolsep}{2pt}
\begin{tabular}{@{}r c c c@{}}
 & {\scriptsize\bfseries \texttt{bike}} & {\scriptsize\bfseries \texttt{chair}} & {\scriptsize\bfseries \texttt{sofa}} \\
 \rotatebox{90}{\tiny\hspace{2pt}Full learnable} &
 \includegraphics[width=0.32\linewidth]{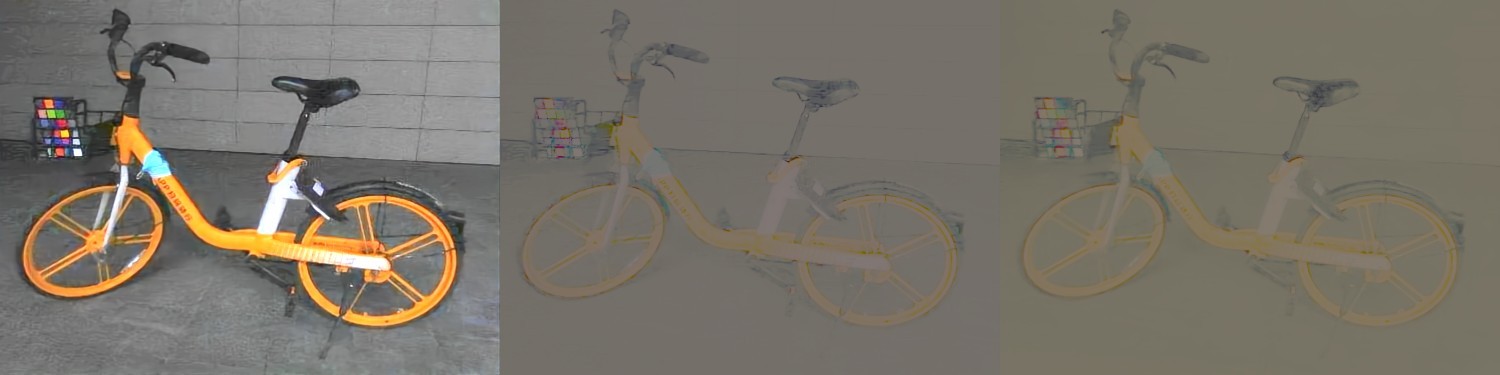} &
 \includegraphics[width=0.32\linewidth]{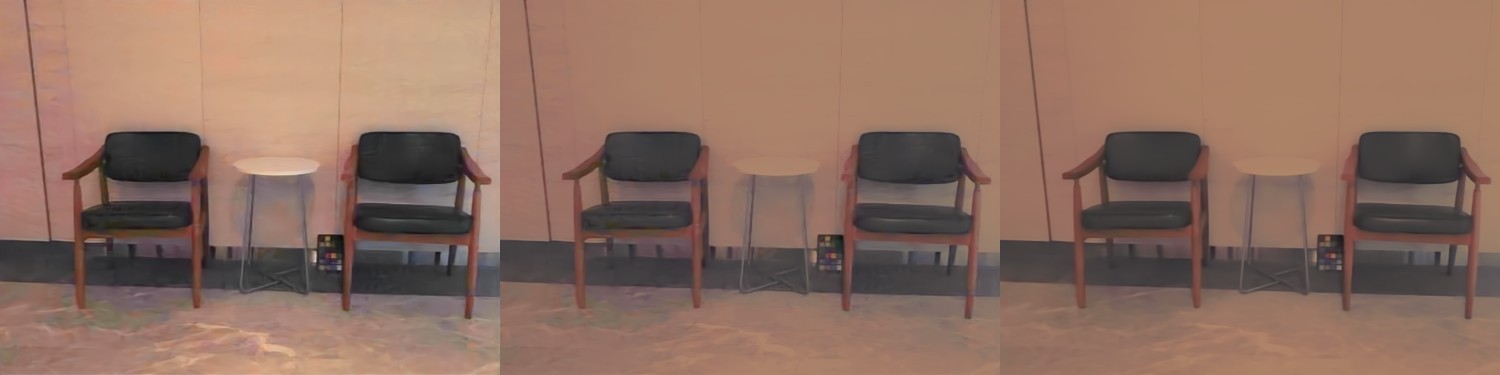} &
 \includegraphics[width=0.32\linewidth]{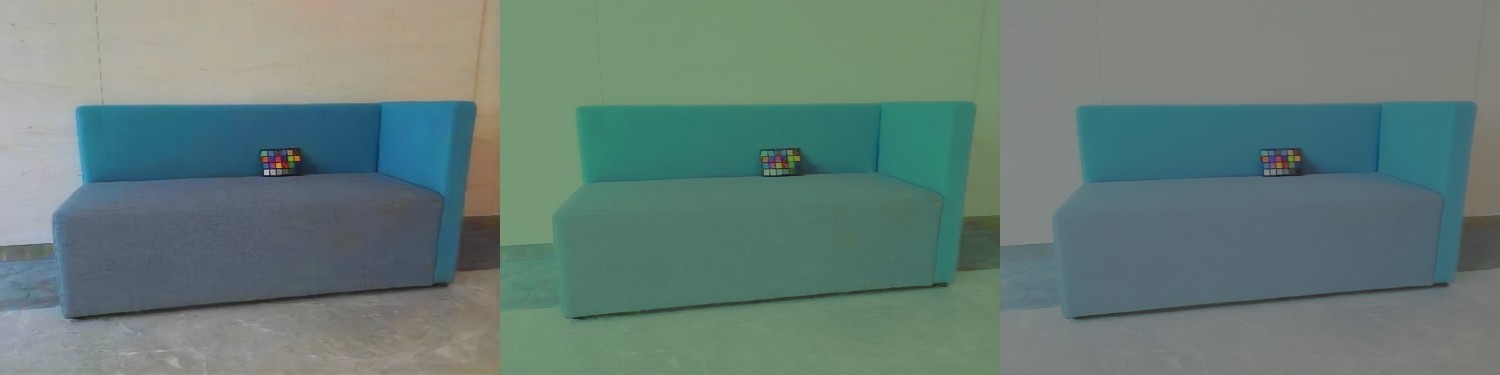} \\
 \rotatebox{90}{\tiny\hspace{4pt}Prior only} &
 \includegraphics[width=0.32\linewidth]{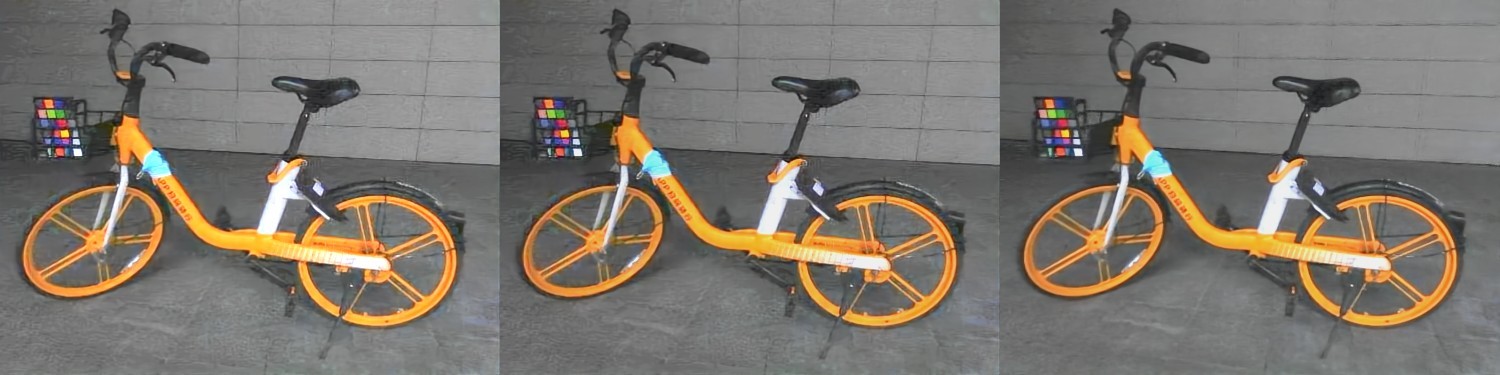} &
 \includegraphics[width=0.32\linewidth]{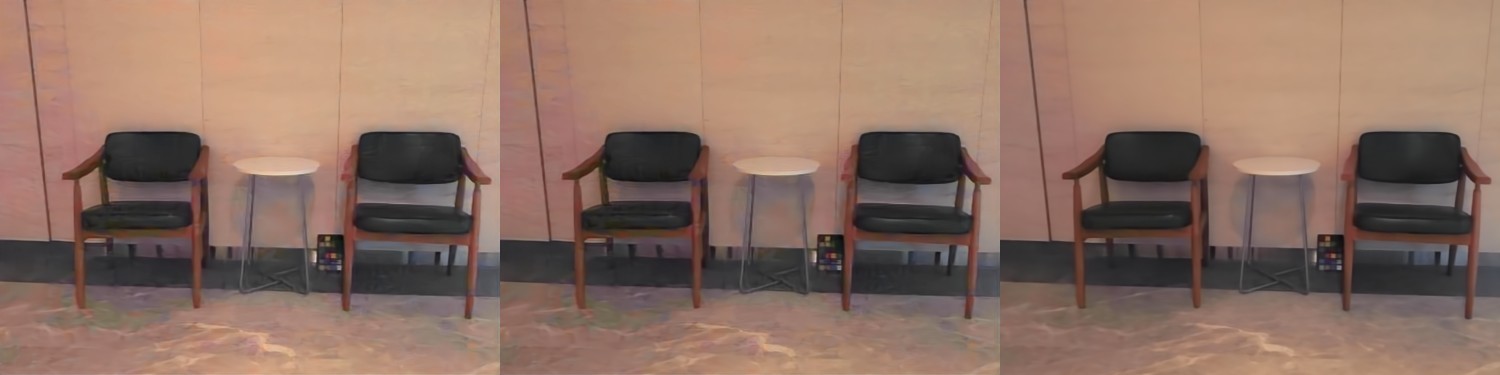} &
 \includegraphics[width=0.32\linewidth]{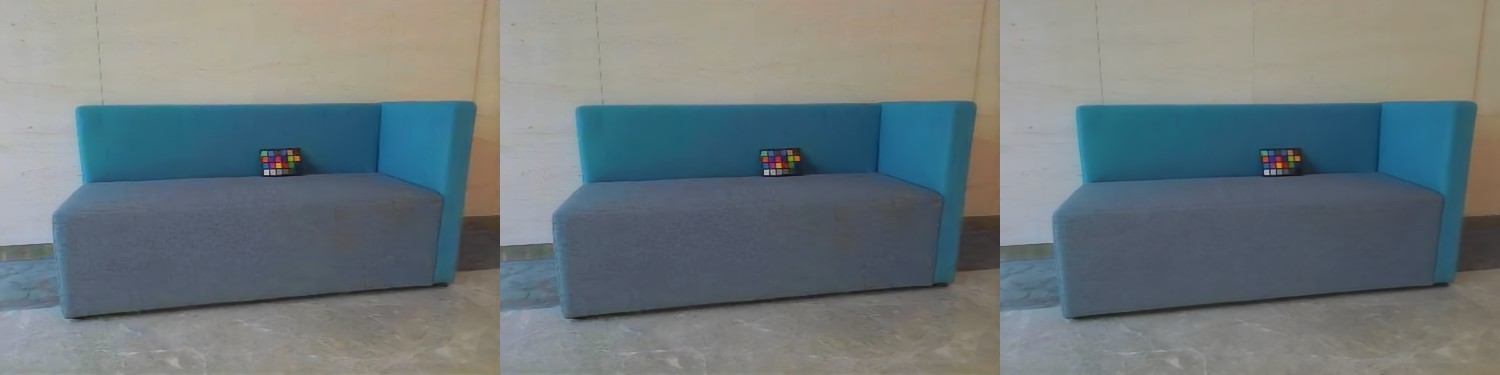} \\
\end{tabular}
\fi
\vspace{-4pt}
\caption{\textbf{Target contrast collapse on all three LOM scenes} (restored inputs, $k=1$, 10k steps; extends Fig.~\ref{fig:t4_moving_target}b--e in the main text). Each strip shows the training input, the learned target $\Phi_\theta(I)$, and the rendering. With the full learnable target (top), the target loses contrast and the rendering follows it; when the photometric gradient to $\theta$ is cut (bottom), the target keeps the contrast of the input.}
\label{fig:supp:collapse}
\vspace{-6pt}
\end{figure*}

\subsection{Depth Priors and Dropout in Sparse Views}
\label{sec:supp:sparse}

We use four LLFF scenes (\texttt{fern}, \texttt{fortress}, \texttt{room}, \texttt{trex}) and four DTU scans (8, 40, 63, 114) with three training views, and train for 10k steps from either a training-only MVS point cloud (similar to the native inputs of the specially-designed methods) or the VGGT-$\Omega$ point cloud of Ours.
Each variant adds one component to the base recipe: an FSGS-style Pearson depth-correlation loss against VGGT-$\Omega$ depth on the training views (weight 0.05), or DropGaussian-style opacity dropout (rate increasing linearly to 0.2 with $1/(1-p)$ compensation).

\begin{table}[t]
    \centering
    \footnotesize
    \setlength{\tabcolsep}{4pt}
    \begin{tabular}{@{}lcccc@{}}
        \toprule
        & \multicolumn{2}{c}{LLFF} & \multicolumn{2}{c}{DTU} \\
        \cmidrule(lr){2-3}\cmidrule(l){4-5}
        Component $-$ base & MVS & VGGT-$\Omega$ & MVS & VGGT-$\Omega$ \\
        \midrule
        Depth correlation loss & +0.70 & +0.30 & +2.16 & +0.63 \\
        \bottomrule
    \end{tabular}
    \caption{\textbf{Gain of the depth prior under two initializations} (PSNR change in dB, four scenes per dataset, 10k steps; plotted in Fig.~\ref{fig:fig5_survive}a in the main text). DTU is evaluated with object masks.}
    \label{tab:supp:sparse}
\end{table}

\cref{tab:supp:sparse} lists the results.
With the VGGT-$\Omega$ initialization, the depth of the initial points already matches the depth prior (Pearson $r$ between 0.58 and 0.998, at least 0.82 in 7 of 8 scenes, 100\% pixel coverage), and the depth-loss gain falls by more than half on both datasets.
With MVS initialization, the correlation ranges from $-0.17$ to 0.91, and initial points cover 23--100\% of the pixels.
On LLFF, opacity dropout changes sign, from $-1.41$\,dB with MVS points to $+0.51$\,dB with VGGT-$\Omega$ points: once the initialization is dense, the main risk shifts from missing geometry to overfitting the three training views.

\subsection{Remaining Gains under Temporary Occlusions}
\label{sec:supp:transient}

\paragraph{Setup.}
We use six scenes covering a range of mask quality and of the advantage of specially-designed methods: RobustNeRF \texttt{yoda} and \texttt{statue}, On-the-go \texttt{mountain} and \texttt{patio}, and RealX3D \texttt{Cupcake} and \texttt{MilkCookie}.
Standard 3DGS and the official RobustSplat receive the same recovered masks and the same training-only SfM initialization; the external mask is multiplied into RobustSplat's own mask, so its online masking and delayed densification stay active.
Both are trained for 30k steps with two seeds.

\paragraph{Repairing missed distractors.}
We estimate the distractors missed by the recovered masks without using any specially-designed method: the residual of standard 3DGS at 7k steps on the training views, thresholded at 0.1, cleaned by morphological opening, restricted to connected regions larger than 0.1\% of the image, and dilated by 2 pixels.
We then additionally exclude all of these regions from the loss, or, as a cross-check, all regions that RobustSplat itself cannot explain.
The 102 control runs in the main text cover these settings together with an intermediate-repair setting and an area-matched placement control on a subset of scenes, for both methods and two seeds (two seed-0 reference runs are reused from the main comparison).

\begin{table}[t]
    \centering
    \footnotesize
    \setlength{\tabcolsep}{4.5pt}
    \begin{tabular}{@{}lccc@{}}
        \toprule
        & Recovered & + Suspected & + RobustSplat \\
        Scene & masks & missed regions & residual regions \\
        \midrule
        \texttt{yoda} & +3.32 & +3.18 & +3.07 \\
        \texttt{statue} & +0.97 & +0.51 & +0.82 \\
        \texttt{mountain} & +1.37 & +1.20 & +1.13 \\
        \texttt{patio} & +0.81 & +0.76 & +0.33 \\
        \texttt{Cupcake} & +1.20 & +0.01 & +0.61 \\
        \texttt{MilkCookie} & +2.15 & +2.32 & +2.02 \\
        \midrule
        Mean & +1.64 & +1.33 & +1.33 \\
        \bottomrule
    \end{tabular}
    \caption{\textbf{RobustSplat gain over standard 3DGS (dB) after repairing missed distractors in the shared masks.} Mean of two seeds.}
    \label{tab:supp:maskdose}
\end{table}

\cref{tab:supp:maskdose} shows that excluding all suspected missed distractors lowers the mean gain of RobustSplat only from $+1.64$ to $+1.33$\,dB ($-0.31$\,dB, Sec.~\ref{sec:analysis_survive} in the main text), and excluding the RobustSplat residual regions gives the same result.
Five of the six scenes keep a gain of at least 0.5\,dB.

\begin{figure*}[t]
    \centering
\iflatexml
    \includegraphics[width=\linewidth,alt={Ground truth, renderings, and error maps on Yoda and MilkCookie with the same recovered masks.}]{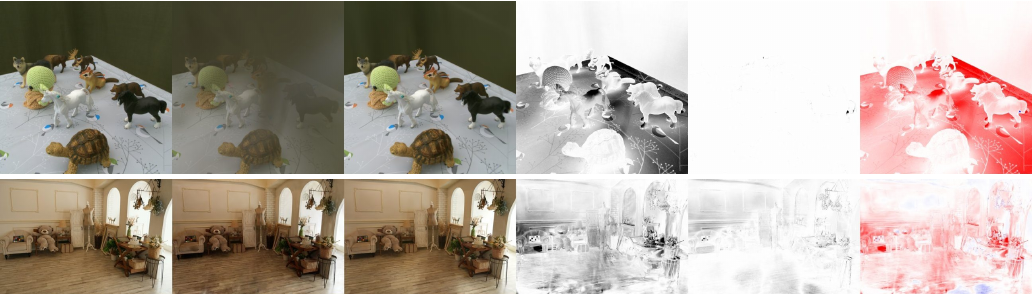}
\else
    \includegraphics[width=\linewidth]{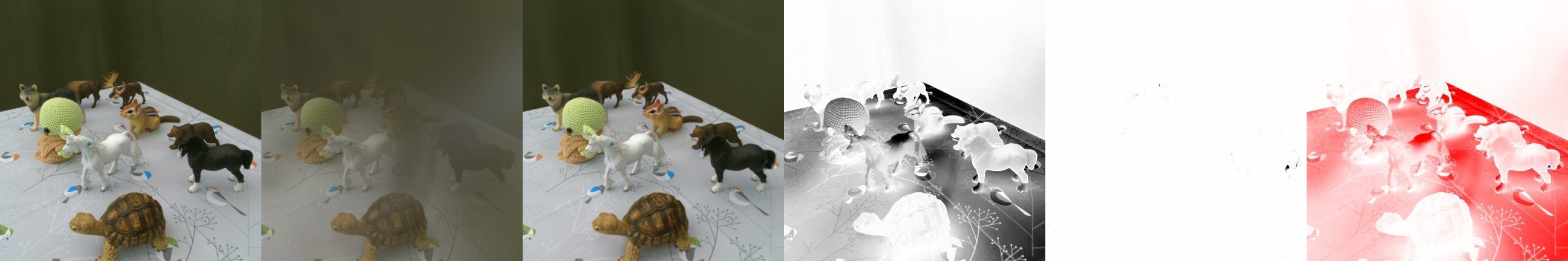}\\[2pt]
    \includegraphics[width=\linewidth]{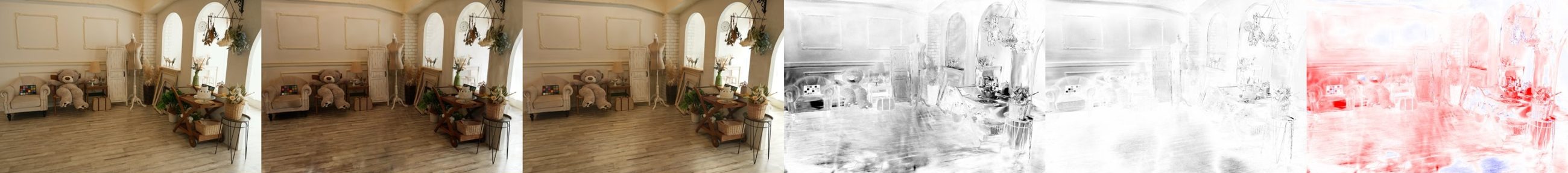}
\fi
    \caption{\textbf{Error maps on the test view with the largest gap.} Columns: GT, standard 3DGS, RobustSplat, error of 3DGS, error of RobustSplat, and their difference (red: 3DGS worse). Top: \texttt{yoda}, where 3DGS renders a near-camera haze. Bottom: \texttt{MilkCookie}, where 3DGS has a global color shift. Both methods use the same recovered masks and initialization.}
    \label{fig:supp:errmap}
\end{figure*}

\paragraph{Where the remaining gap comes from.}
We compare standard 3DGS and RobustSplat pixel by pixel on all test views.
On \texttt{yoda} and \texttt{Cupcake}, the worst 10\% of test views account for 91\% and 83\% of the MSE gap.
On \texttt{yoda}, about 10\% of the test views are covered by a gray haze in standard 3DGS (11--14\,dB versus 32--36\,dB for RobustSplat).
The haze is geometric: in these views the rendered depth of standard 3DGS is 1/5 to 1/7 of the scene depth, and it consists of many low-opacity Gaussians in front of the cameras that are nearly transparent in the training views.
On \texttt{MilkCookie}, the gap is spread over all views and half of it disappears after a per-image affine color correction; its training images also have the largest brightness variation of the six scenes.
\cref{fig:supp:errmap} shows examples.

\paragraph{Generic training changes.}
Keeping the masks and initialization fixed, we change one part of standard 3DGS at a time.
3DGS-MCMC~\cite{kheradmand20243d} replaces the densification strategy (refinement from 500 to 25k steps, a Gaussian budget equal to the final count of standard 3DGS on the same scene, and the default opacity and scale regularizers of 0.01).
A per-view $3\times4$ affine color transform is applied to the rendering in the training loss only, and not at test time.
RobustSplat's delayed densification window (10k--20k steps) is tested as a third change.

\begin{table}[t]
    \centering
    \footnotesize
    \setlength{\tabcolsep}{3.2pt}
    \begin{tabular}{@{}lccccc@{}}
        \toprule
        Scene & 3DGS & +Delay & MCMC & +Affine & RobustSplat \\
        \midrule
        \texttt{yoda} & 31.78 & 31.19 & \textbf{35.17} & 30.53 & 35.09 \\
        \texttt{statue} & 22.04 & 22.38 & \textbf{23.09} & 22.27 & 23.01 \\
        \texttt{mountain} & 20.04 & 16.79 & \textbf{22.16} & 20.24 & 21.41 \\
        \texttt{patio} & 21.15 & 20.66 & 21.43 & 21.46 & \textbf{21.96} \\
        \texttt{Cupcake} & 29.40 & 29.22 & 29.30 & 29.58 & \textbf{30.60} \\
        \texttt{MilkCookie} & 24.02 & 22.26 & 24.02 & \textbf{26.76} & 26.17 \\
        \midrule
        Mean PSNR & 24.74 & 23.75 & 25.86 & 25.14 & \textbf{26.38} \\
        Mean LPIPS & 0.151 & 0.175 & \textbf{0.118} & 0.146 & 0.138 \\
        \bottomrule
    \end{tabular}
    \caption{\textbf{Generic 3DGS training changes under the same recovered masks and initialization} (PSNR in dB). 3DGS and RobustSplat average two seeds on every scene; the three variants average two seeds on \texttt{yoda} and \texttt{MilkCookie} and use seed 0 on the other scenes. These are the values plotted in Fig.~\ref{fig:fig5_survive}b in the main text.}
    \label{tab:supp:mcmc}
\end{table}

\begin{figure*}[t]
\centering
\iflatexml
\includegraphics[width=\linewidth,alt={One held-out Yoda view rendered by standard 3DGS, MCMC and RobustSplat alongside the reference image.}]{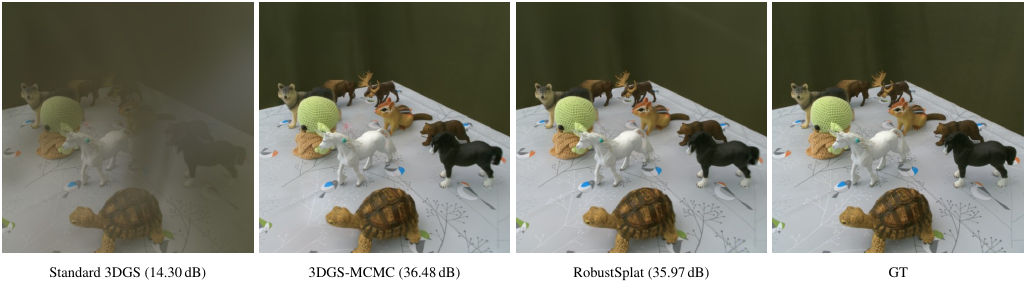}
\else
\setlength{\tabcolsep}{1.5pt}
\begin{tabular}{@{}cccc@{}}
\includegraphics[width=0.243\linewidth]{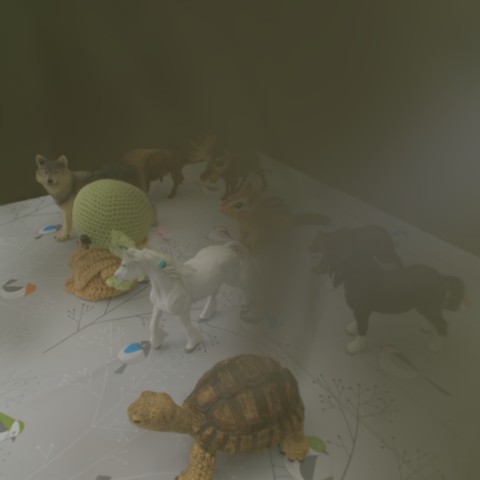} &
\includegraphics[width=0.243\linewidth]{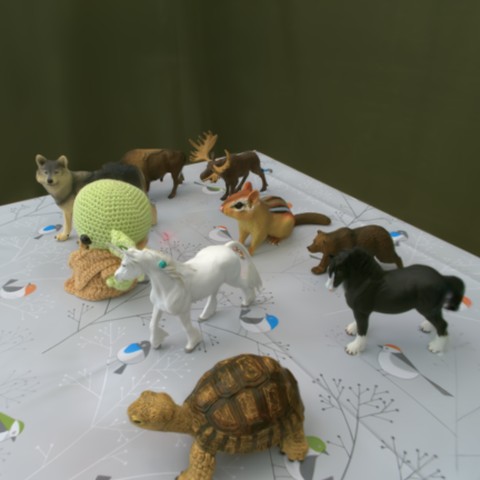} &
\includegraphics[width=0.243\linewidth]{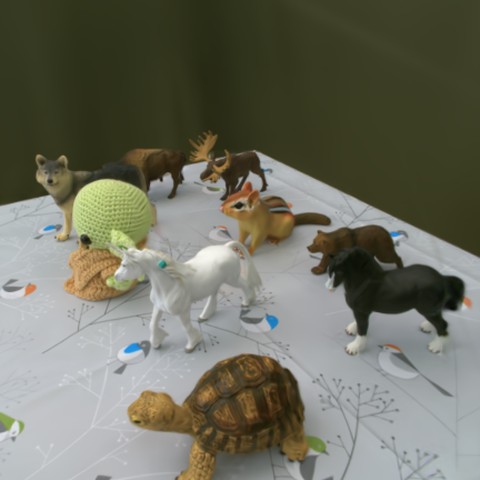} &
\includegraphics[width=0.243\linewidth]{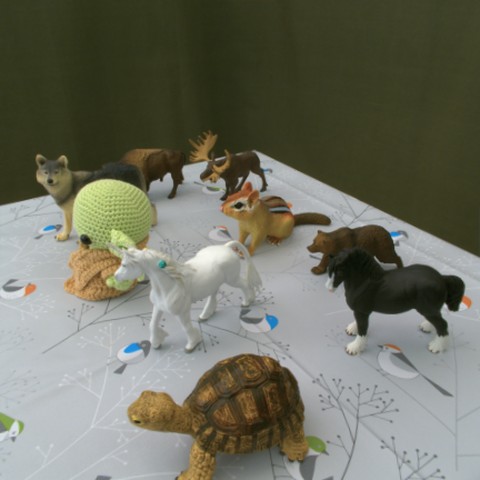} \\
{\scriptsize Standard 3DGS (14.30\,dB)} & {\scriptsize 3DGS-MCMC (36.48\,dB)} & {\scriptsize RobustSplat (35.97\,dB)} & {\scriptsize GT} \\
\end{tabular}
\fi
\vspace{-4pt}
\caption{\textbf{A generic densification change removes the near-camera floaters on \texttt{yoda}} (test view \texttt{1extra135}; extends Fig.~\ref{fig:fig5_survive}c in the main text). All three use the same recovered masks and initialization. Replacing only the densification strategy of standard 3DGS with 3DGS-MCMC removes the haze, without any occlusion-specific module.}
\label{fig:supp:floater}
\vspace{-6pt}
\end{figure*}

\cref{tab:supp:mcmc} shows that 3DGS-MCMC reduces the mean gap to RobustSplat from 1.64 to 0.52\,dB and gives a lower
LPIPS than RobustSplat on five of the six scenes (mean 0.118 vs.\ 0.138).
On the \texttt{yoda} view of \cref{fig:supp:floater}, it removes the haze and raises PSNR from 14.30 to 36.48\,dB.
The per-view color transform targets cross-view color inconsistency and helps most on \texttt{MilkCookie}, the scene with the largest brightness variation, where it exceeds RobustSplat.
In contrast, adopting RobustSplat's delayed densification window lowers the mean PSNR by 1.0\,dB, so its advantage does not come from this schedule.

\iflatexml\else
\begin{figure*}[!tp]
\noindent\begin{minipage}[t]{\textwidth}
\fi
\section{Additional Qualitative Results}
\label{sec:supp:qualitative}

\cref{fig:supp:qualitative} extends Fig.~\ref{fig:qualitative} in the main text (same scenes, views, and specially-designed methods) with a degraded training input or a reference image, and the vanilla 3DGS result, for each row.
For sparse views and pose-free reconstruction the training images themselves are not degraded, so the first column shows the held-out reference image of the displayed view; it is not a training input.
For the pose-free row, vanilla 3DGS uses MASt3R cameras and points, because standard 3DGS needs cameras.

\iflatexml
  \begin{figure*}
\fi
\input{sec/supp/supp-qualitative-figure}
\iflatexml
  \end{figure*}
\else
\end{minipage}\par
\end{figure*}
\fi

\end{document}